%% file: main.tex
\documentclass[journal]{new-aiaa}
\usepackage[utf8]{inputenc}
\usepackage{textcomp}

\usepackage{graphicx}
\usepackage{amsmath}
\usepackage[version=4]{mhchem}
\usepackage{siunitx}
\usepackage{longtable,tabularx}
\usepackage[export]{adjustbox}
\usepackage{bm}
\usepackage{multirow}
\usepackage{subcaption}
\usepackage{afterpage}
\usepackage{pdflscape}
\usepackage{accents}
\usepackage{rotating}

\usepackage{array}
\usepackage{colortbl} 
\usepackage[table]{xcolor}
\usepackage{booktabs} 
\newcommand{\ra}[1]{\renewcommand{\arraystretch}{#1}} 
\newcommand{\cc}[1]{\cellcolor{#1}} 
\newcolumntype{R}[1]{>{\raggedleft\arraybackslash}p{#1}}
\newcolumntype{L}[1]{>{\raggedright\arraybackslash}p{#1}}

\newcommand{\cbg}[2]{\setlength{\fboxsep}{0pt}\colorbox{#1}{\strut #2}}

\DeclareMathOperator*{\argmax}{arg\,max}
\DeclareMathOperator*{\argmin}{arg\,min}

\newcommand{\hvec}[1]{\underline{\bm{\mathrm{#1}}}}
\newcommand{\vvec}[1]{\bm{\mathrm{#1}}}
\newcommand{\uvec}[1]{\bm{\mathrm{#1}}}
\newcommand{\oB}{\mathrm{B}}
\newcommand{\oC}{\mathrm{C}}
\newcommand{\oS}{\mathrm{S}}
\newcommand{\oI}{\mathrm{I}}
\newcommand{\fB}{\mathcal{B}}
\newcommand{\fC}{\mathcal{C}}
\newcommand{\fS}{\mathcal{S}}

\newcommand{\given}[1][]{\:#1\vert\:}
\newcommand{\sk}[1]{  \left[#1\right]^\wedge }
\newcommand{\Exp}[1]{\texttt{exp}{\left( \sk{#1}\right)}}
\newcommand{\sang}{\iota}
\newcommand{\eang}{\varepsilon}

\usepackage{xcolor}
\usepackage{tikz}
\usetikzlibrary{shapes,arrows,calc,patterns,angles,quotes,fadings,decorations.pathreplacing}
\usepackage{tikz-3dplot}
\usepackage{tkz-graph}
\tikzstyle{vertex}=[circle, draw, inner sep=0pt, minimum size=8pt]
\newcommand{\vertex}{\node[vertex]}

\tikzset{%
  >=latex, 
  inner sep=0pt,%
  outer sep=1pt,%
  mark coordinate/.style={inner sep=0pt,outer sep=0pt,minimum size=3pt,
    fill=black,circle}%
}

\definecolor{desert}{rgb}{0.76, 0.6, 0.42}
\definecolor{darkpastelgreen}{rgb}{0.01, 0.75, 0.24}
\definecolor{darkspringgreen}{rgb}{0.09, 0.45, 0.27}
\definecolor{cobalt}{rgb}{0.0, 0.28, 0.67}
\definecolor{darkgreen}{rgb}{0.0, 0.5, 0.0}
\definecolor{coquelicot}{rgb}{1.0, 0.22, 0.0}
\definecolor{canaryyellow}{rgb}{1.0, 0.94, 0.0}
\definecolor{cadetgrey}{rgb}{0.57, 0.64, 0.69}
\definecolor{cadet}{rgb}{0.33, 0.41, 0.47}
\definecolor{mediumcandyapplered}{rgb}{0.89, 0.02, 0.17}
\definecolor{dimgray}{rgb}{0.41, 0.41, 0.41}

\title{Stereophotoclinometry Revisited}

\author{Travis Driver \footnote{PhD Candidate, Institute for Robotics and Intelligent Machines, Georgia Institute of Technology, Atlanta, GA}}
\affil{Georgia Institute of Technology, Atlanta, GA}
\author{Andrew Vaughan\footnote{Senior Engineer, Mission Design and Navigation, Jet Propulsion Laboratory, California Institute of Technology, Pasadena, CA}, 
Yang Cheng\footnote{Principal Robotics Technologist, Mobility and Robotic Systems, Jet Propulsion Laboratory, California Institute of Technology, Pasadena, CA}, and 
Adnan Ansar\footnote{Principal Robotics Technologist, Mobility and Robotic Systems, Jet Propulsion Laboratory, California Institute of Technology, Pasadena, CA}}
\affil{Jet Propulsion Laboratory, California Institute of Technology, Pasadena, CA}
\author{John Christian\footnote{Associate Professor, School of Aerospace Engineering, Georgia Institute of Technology, Atlanta, GA} and 
Panagiotis Tsiotras\footnote{David \& Andrew Lewis Chair, Professor, Institute for Robotics and Intelligent Machines, School of Aerospace Engineering, Georgia Institute of Technology, Atlanta, GA}}
\affil{Georgia Institute of Technology, Atlanta, GA}

\begin{document}

\maketitle

\begin{abstract}
Image-based surface reconstruction and characterization is crucial for missions to small celestial bodies, as it informs mission planning, navigation, and scientific analysis. 
However, current state-of-the-practice methods, such as stereophotoclinometry (SPC), rely heavily on human-in-the-loop verification and high-fidelity \textit{a priori} information.
This paper proposes \underline{Pho}toclinometry-from-\underline{Mo}tion (PhoMo), a novel framework that incorporates photoclinometry techniques into a keypoint-based structure-from-motion (SfM) system to estimate the surface normal and albedo at detected landmarks to improve \textit{autonomous} surface and shape characterization of small celestial bodies from \textit{in-situ} imagery. 
In contrast to SPC, we forego the expensive maplet estimation step and instead use dense keypoint measurements and correspondences from an \textit{autonomous} keypoint detection and matching method based on deep learning. 
Moreover, we develop a factor graph-based approach allowing for \textit{simultaneous} optimization of the spacecraft's pose, landmark positions, Sun-relative direction, and surface normals and albedos via fusion of Sun vector measurements and image keypoint measurements. 
The proposed framework is validated on \textit{real} imagery taken by the Dawn mission to the asteroid 4 Vesta and the minor planet 1 Ceres and compared against an SPC reconstruction, where we demonstrate superior rendering performance compared to an SPC solution and precise alignment to a stereophotogrammetry (SPG) solution without relying on any \textit{a priori} camera pose and topography information or humans-in-the-loop.
\end{abstract}

\newpage


\section{Introduction}
\input{text/introduction}


\input{text/related_work}


\section{Background}
\input{text/background}


\section{Methodology}
\input{text/methods}


\section{Experimental Setup}
\input{text/experiments}


\section{Results}
\input{text/results}


\section{Conclusion}

Image-based surface reconstruction and characterization play a critical role in enabling mission planning, navigation, and scientific analysis for small celestial bodies. 
However, traditional methods like stereophotoclinometry (SPC) rely heavily on human intervention and high-fidelity \textit{a priori} information, limiting their scalability and autonomy. 
This paper introduces PhoMo (Photoclinometry-from-Motion), a novel framework that integrates photoclinometry into a keypoint-based structure-from-motion (SfM) system to autonomously estimate surface normals and albedos for enhanced shape and surface characterization.

The PhoMo framework leverages dense, deep learning-based keypoint detection and matching to replace SPC's labor-intensive maplet estimation process. 
By employing a factor graph-based optimization approach, PhoMo enables the simultaneous estimation of spacecraft pose, landmark positions, Sun-relative direction, and photometric properties, achieving precise reconstructions without requiring any \textit{a priori} knowledge or human-in-the-loop verification.
Extensive validation on real imagery from NASA's Dawn mission demonstrates that PhoMo achieves superior rendering quality compared to SPC, as measured by PSNR, and aligns closely with stereophotogrammetry (SPG) reconstructions. 
Importantly, PhoMo’s physics-based photometric modeling allows for accurate representation of surface reflectance properties, outperforming implicit representations such as neural radiance fields and Gaussian splatting in terms of reliability and interpretability.

These results highlight PhoMo's potential to enable fully autonomous surface reconstruction and characterization for future missions to small celestial bodies. 
By eliminating the reliance on human intervention and high-fidelity \textit{a priori} information, PhoMo represents a significant step toward enhancing the autonomy and efficiency of planetary exploration missions. 
Future work may focus on extending the framework to incorporate additional sensors (e.g., pushbroom cameras), applying the proposed framework for global reconstruction of a small body, and continuing validation of its performance under diverse illumination and terrain conditions. 
The code and results will be made available at \href{https://github.com/travisdriver/phomo}{\texttt{https://github.com/travisdriver/phomo}} upon publication.


\section*{Appendix}
\vspace{5pt}
\input{text/appendix}


\section*{Acknowledgments}

This work was supported by NASA Space Technology Graduate Research Opportunity 80NSSC21K1265.
A portion of this research was carried out at the Jet Propulsion Laboratory, California Institute of Technology, under a contract with the National Aeronautics and Space Administration (80NM0018D0004). 
The authors would like to thank Kenneth Getzandanner and Andrew Liounis from NASA Goddard Space Flight Center for several helpful discussions and comments.
We also thank Stefanus E. Schr\"{o}der from Lule\r{a} University of Technology for providing support for the radiometric calibration process~\cite{schroder2013cal,schroder2014cal}.


\bibliography{sample}

\end{document}

%% file: text/introduction.tex
\lettrine{T}here has been an increasing interest in missions to small bodies (e.g., asteroids, comets) due to their great scientific value, with seven currently in operation (Hayabusa2, OSIRIS-APEX, Lucy, Psyche, Europa Clipper, Hera) and six scheduled to launch over the next five years (Odin, Tianwen-2, Mars Moon eXploration, MBR Explorer, DESTINY+, Comet Interceptor). 
In addition to planetary defense~\cite{cheng2018} and resource utilization~\cite{mazanek2015,rivkin2019}, small bodies are believed to be remnants of the solar system's formation, and studying their composition could provide insight into the evolution of the solar system and the origins of organic life on Earth~\cite{barucci2011}. 
These missions currently rely on an extended characterization phase, where a shape model is reconstructed from images acquired during a ground-controlled trajectory around the body, as shape models are essential for characterizing the body and estimating the spacecraft's relative pose in subsequent phases~\cite{bhaskaran2011}. 
However, current state-of-the-practice shape reconstruction methods rely on humans-in-the-loop and accurate \textit{a priori} information to ensure accurate results. 

\begin{figure}[t!]
    \centering
    \includegraphics[width=0.8\linewidth]{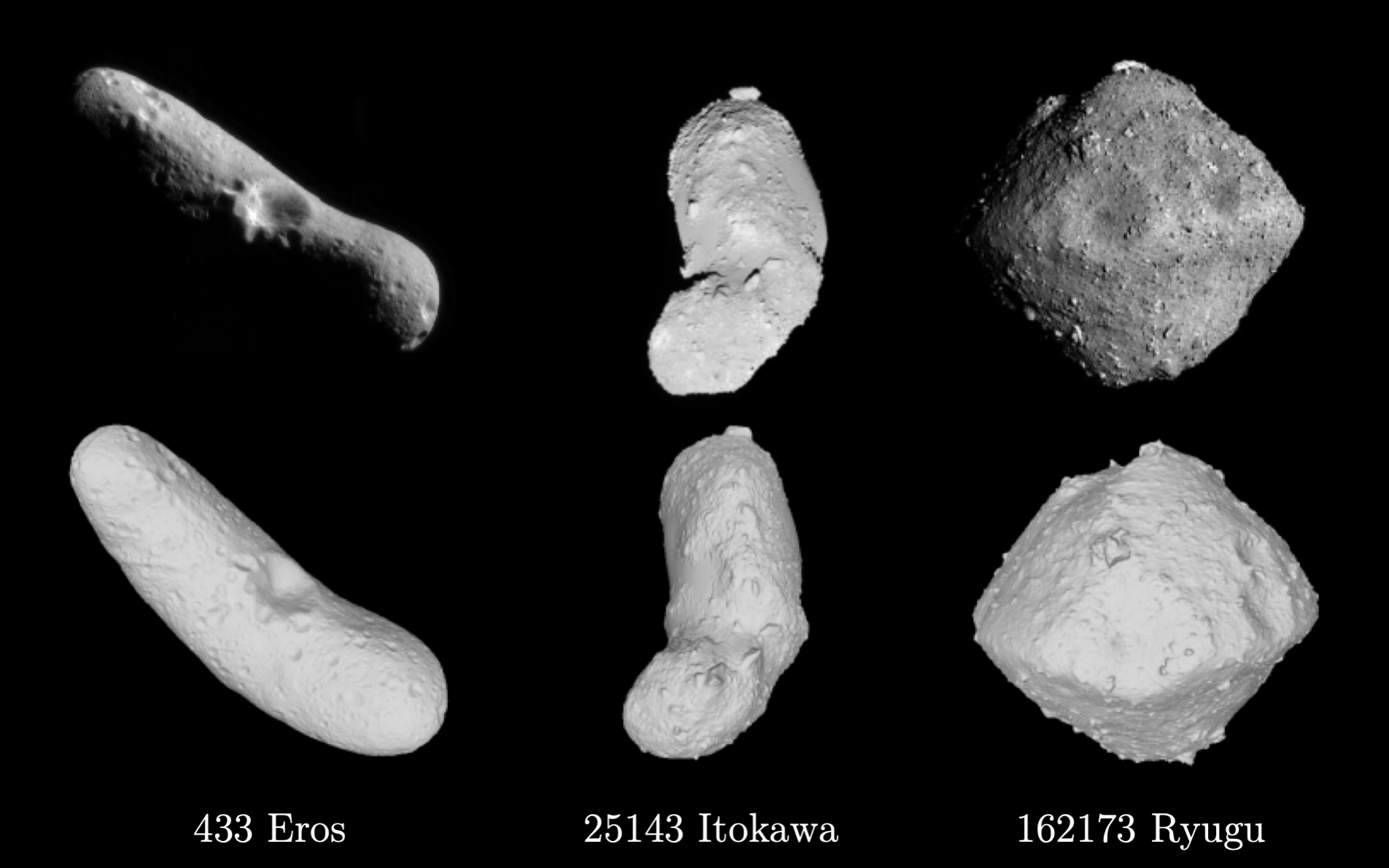}
    \caption{Examples of shape models produced by SPC.}
    \label{fig:spc-examples}
\end{figure}

Stereophotoclinometry~\cite{gaskell2008,gaskell2023psj} (SPC) is the current method of choice for 3D reconstruction of small bodies, and has been used to model a broad suite of celestial bodies, including the Moon, 433 Eros, 25143 Itokawa~\cite{gaskell2008}, 4 Vesta~\cite{raymond2011dawn, mastrodemos2012}, and 101955 Bennu~\cite{lorenz2017, barnouin2020} (see Fig. \ref{fig:spc-examples}).
While SPC has proven effective, the process requires extensive human-in-the-loop verification and high-fidelity \textit{a priori} information to achieve accurate results. 
Specifically, SPC attempts to estimate a collection of digital terrain maps (DTMs), high-resolution local topography and albedo maps, through direct alignment of ortho-rectified projections, or \textit{orthoimages}, of a given surface patch from multiple images. 
This alignment process depends on an initial shape model and precise \textit{a priori} estimates of the spacecraft’s pose (position and orientation). 
Photoclinometry is then applied to derive surface gradients and albedo values for the imaged surface patches at each pixel of the DTM.
%
The local topography solutions are fixed upon convergence, typically requiring human input to achieve precise alignment to the images, and used to refine pose and landmark position estimates through a multistep iterative process by rendering the DTM and aligning it across multiple views~\cite{gaskell2023psj}. 
Finally, the local DTMs can be collated into a global shape model by exploiting overlap and limb constraints within a separate iterative processing loop. 
While this approach has achieved much success, its reliance on extensive human involvement for extended durations and a complex multistep optimization process limits mission capabilities and increases operational costs~\cite{quadrelli2015,nesnas2021,getzandanner2022scitech}. 

In contrast to the maplet-based approach of SPC, this work proposes \underline{Pho}toclinometry-from-\underline{Mo}tion (PhoMo), a keypoint-based framework that fuses dense image keypoint measurements and correspondences from a deep learning algorithm with Sun vector measurements within a structure-from-motion (SfM) system to estimate the topography of small body surfaces. 
Indeed, SfM and simultaneous localization and mapping (SLAM), which leverage autonomous keypoint detection and matching methods to estimate correspondences between images, have been shown to be promising technologies for \textit{autonomous} optical navigation and mapping for missions to small bodies~\cite{dor2021cvpr,dor2022astroslam}. 
Consequently, we propose the incorporation of photoclinometry constraints and Sun vector measurements into a feature-based SfM system to estimate surface normals and albedos at estimated landmarks, providing detailed information for surface characterization and shape reconstruction. 
The proposed framework, which leverages \textit{factor graphs}~\cite{dellaert2017} to model and solve the complex photoclinometry and SfM process, forgoes the expensive interative local maplet alignment step, streamlines the optimization process, and renders SPC more amenable to recent and future advances in computer vision, namely feature detection, description, and matching methods based on \textit{deep learning}~\cite{driver2022astrovision}.  

The key contributions of this work are summarized below.
\begin{enumerate}
    \item We propose the fusion of dense image keypoint measurements and correspondences, derived using a data-driven keypoint detection and matching approach~\cite{edstedt2024roma}, and Sun vector measurements in a SfM system using photoclinometry to reconstruct a landmark map of the body and the relative pose of the spacecraft while \textit{simultaneously} estimating the surface normal and albedo at landmarks to provide a more efficient and autonomous alternative to SPC.
    \item We model the photoclinometry constraints using the formalism of \textit{factor graphs} and analyze its performance with respect to five different reflectance models.
    \item We apply the proposed framework to \textit{real} imagery from the Dawn mission to asteroid 4 Vesta and minor planet 1 Ceres and demonstrate superior rendering performance to an SPC-derived map and precise alignment to a stereophotogrammetry (SPG) solution.
\end{enumerate}

%% file: text/related_work.tex
\section{Related Work}

In this section, we review previous work related to sparse and dense 3D reconstruction of small celestial bodies, as well as recently proposed rendering methods based on deep learning. 


\subsection{State-of-the-Practice for 3D Reconstruction of Airless Bodies}

SPC is the \textit{de facto} technique for 3D reconstruction of small celestial bodies. 
It works by densely aligning an image of a surface patch to a rendering under identical illumination conditions generated using \textit{a priori} estimates of the camera pose, illumination direction, surface topography, albedo, and a reflectance model through correlation-based matching.
This process is repeated across multiple images to generate multiple correspondences for each cell in the initial maplet, followed by photoclinometry to refine the surface topography and albedo of the resulting DTM.
The proposed approach takes inspiration from the success of SPC, but introduces several key differences. 
First, and most notably, we forego the dense alignment step used to generate local DTMs and, instead, rely on keypoint measurements, which can be computed by autonomous feature detection and matching methods, which have been shown to be robust to the significant illumination and perspective changes inherent to small body imagery~\cite{driver2022astrovision}. 
Thus, our framework treats keypoints (referred to as \textit{landmark image locations} in the SPC text~\cite{gaskell2023psj}) as measurements rather than another variable of interest that must be estimated independently. 
Second, SPC implicitly represents surface normals by estimating $x$- and $y$-surface slopes at each pixel defined in a local frame of each DTM, which are subsequently integrated to generate a dense topography solution. 
Conversely, we leverage a minimal representation of the surface normals based on retractions of the tangent space of the $\mathbb{S}^2$ manifold, as detailed in Section \ref{sec:unit-vectors}, allowing for global and simultaneous optimization of all observed landmarks and their associated surface normals. 
Lastly, we exploit the formalism of factor graphs (see Section \ref{sec:factor-graphs}) to model the complex estimation problem, as opposed to the iterative multistep process employed by the traditional SPC implementation~\cite{gaskell2023psj}, allowing for the simultaneous optimization of camera poses, landmark positions, and surface normals and albedos. 

Stereophotogrammetry (SPG) is another popular method for dense reconstruction of planetary bodies~\cite{gwinner2009derivation,preusker2011stereo}.
Similar to SPC, SPG orthorectifies images based on \textit{a priori} knowledge of the topography and spacecraft state, followed by correlation-based matching and subpixel refinement using least-squares matching~\cite{gwinner2009derivation}. 
However, SPG has strict illumination requirements for accurate matching because the correlation-based matching works directly on the original (orthorectified) images, unlike SPC, which renders images to the same illumination conditions to facilitate matching.
For example, the SPG pipeline employed by the Dawn mission required that the stereo pairs have $<$$10^\circ$ difference in illumination direction. 
Despite these constraints, previous studies have shown that SPG can achieve more accurate reconstructions than SPC when illumination variations are limited~\cite{raymond2011dawn}. 
In Section~\ref{sec:spc-spg-sfm-comparison}, we compare PhoMo with the reconstructions generated by both SPC and SPG. 


\subsection{Reconstruction using Active Sensing}

Vision-based methods, i.e., SPC, have traditionally been applied to 3D reconstruction for small celestial bodies. 
However, approaches based on active sensors such as Flash-LiDARs have also been proposed. 
Bercovici et al.~\cite{bercovici2019} proposed a pose estimation and shape reconstruction approach based on Flash-LiDAR measurements by solving a maximum likelihood estimation problem via particle-swarm optimization to refine an initial Bezier surface mesh, followed by a least-squares filter providing measurements for the position and orientation of the spacecraft. 
Other works in the field have established proofs-of-concept for batch optimization and graph-based approaches for near-asteroid navigation and shape reconstruction. 

Notably, Nakath et al.~\cite{nakath2020} present an active SLAM framework that also employs Flash-LiDAR as the base measurement, with sensor fusion of data from an inertial measurement unit and star tracker, tested with simulated data. 
However, the limited range of Flash-LiDAR instruments restricts the spacecraft's orbit to unrealistically small radii, reducing the feasible scenarios to either navigation near very small asteroids or the touchdown phase. 
For example, the OSIRIS-REx Guidance, Navigation, and Control (GNC) Flash-LiDAR, which is mentioned in both \cite{nakath2020} and \cite{bercovici2019}, has only a maximum operational range of approximately 1 km and a relatively small $128\times 128$ detector array~\cite{church2020lidar,leonard2022lidar} 

The recent OSIRIS-REx mission to asteroid 101955 Bennu was also equipped with the OSIRIS-REx Laser Altimeter (OLA)~\cite{daly2017} to provide an alternative means of shape reconstruction to SPC. 
The OLA reconstruction process begins by generating local digital elevation maps (DEMs) from range measurements, which are then merged through an iterative closest-point algorithm. 
However, the OSIRIS-REx camera suite (OCAMS)~\cite{rizk2018ocams} features a long-range camera that provides higher spatial resolution than OLA at the same distance. 
As a result, although the SPC process is more time-consuming than the OLA-based approach, SPC products can be available before OLA models, and with a higher resolution. 
Moreover, testing of OLA-generated DEMs showed that uncertainty in OLA measurements created unacceptable errors in elevations of smaller features, and albedo is not automatically included as part of the solution~\cite{lorenz2017}. 
Ultimately, the SPC data products were used in the final touch-and-go phase of the mission. 

The Lunar Orbiter Laser Altimeter (LOLA) onboard the Lunar Reconnaissance Orbiter (LRO) has also been used to generate global DEMs of the lunar surface. 
However, the spatial resolution and accuracy of LOLA DEMs is relatively low compared to what can be achieved by processing images from the Lunar Reconnaissance Orbiter Camera (LROC) suite.
For example, Boatwright et al.~\cite{boatwright2024lola} leveraged 5 meters/pixel LOLA DEMs~\cite{barker2021lola} to initialize an SPC process with LROC Narrow Angle Camera (NAC) images to generate maps at 1 meter/pixel for potential Artemis landing sites.


\subsection{Featured-Based SfM and SLAM}

Feature-based methods have been shown to be promising technologies for \textit{autonomous} optical navigation and mapping for missions to small bodies. 
Most notably, Dor et al.~\cite{dor2021cvpr} demonstrated precise visual localization and mapping on \textit{real} images of Asteroid 4 Vesta through a feature-based SLAM system based on Oriented FAST and Rotated BRIEF (ORB) features~\cite{rublee2011iccv}. 
ORB is a handcrafted method based on Features from Accelerated Segment Test (FAST) keypoints~\cite{rosten2006eccv} and Binary Robust Independent Elementary Features (BRIEF) descriptors~\cite{calonder2010eccv} and outputs binary descriptor vectors, enabling more efficient matching. 
This work was extended in \cite{dor2022astroslam} to include known dynamical motion constraints between the small body and the spacecraft to further improve mapping and localization performance. 
Furthermore, Driver et al.~\cite{driver2022astrovision} proposed the use of deep learning-based feature detection and description methods, which were shown to significantly outperform traditional handcrafted methods (e.g., ORB), especially in scenarios involving considerable changes in illumination and perspective. 
Our work capitalizes on this recent success in feature-based SLAM and SfM for autonomous optical navigation by imbuing the traditional SfM framework with added characterization power by incorporating photoclinometry constraints for concurrent estimation of surface normals and albedos. 


\subsection{Photometric Stereo and Implicit Scene Representations} \label{sec:related-dl}

SPC borrows many techniques from the process of \textit{photometric stereo}~\cite{woodham1980}, which has been used extensively in terrestrial applications.
This is not to be confused with \textit{shape-from-shading} (SfS), whereby the shape of a 3D object may be recovered from shading in a \textit{single image}.
However, terrestrial photometric stereo formulations have relied on a number of simplifying assumptions, including Lambertian reflectance~\cite{hayakawa1994photometric,logothetis2019differential} and specialized lighting or image capture setups~\cite{shi2013bi,ikehata2014photometric,cho2018semi}.  
Methods based on deep learning have also been proposed but, as before, require specially designed image acquisition setups~\cite{santo2017dpsn,chen2018psfcn,bi2020deep3d, kaya2023multi}, and thus cannot be leveraged in a general multi-view reconstruction scenario. 
We refer the reader to \cite{ackermann2015survey} and \cite{ju2022deep} for more information about physics-based and data-driven approaches, respectively, to photometric stereo for terrestrial applications. 

Neural Radiance Fields (NeRFs) and 3D Gaussian Splatting (3DGS) are also notable photometric reconstruction methods that leverage implicit scene representations to model surface structure and reflectance. 
NeRFs~\cite{mildenhall2020nerf} capture the reflectance and material properties of a target object or scene through the learned weights of multilayer perceptrons (MLPs), which can be sampled at discrete points along a ray to render the imaged scene. 
NeRFs have demonstrated impressive 3D reconstruction and rendering capabilities on asteroid imagery~\cite{chen2024asteroidnerf,givens2024nerf}. 
In contrast, 3DGS~\cite{kerbl20233dgs} models the environment as a collection of 3D Gaussians, which are projected--or ``splatted''--onto the image plane to reconstruct the scene. 
Our approach diverges from these methods by leveraging semi-empirical photometric models of airless bodies with a small number of free parameters, allowing us to explicitly estimate the topography and material properties of the surface.
We will compare against these implicit surface representations and demonstrate superior rendering quality on images of airless bodies.

%% file: text/background.tex
In this section, we summarize the representation of 3D poses as elements of the Special Euclidean group $\mathrm{SE}(3)$ (Section \ref{sec:rigid-transformations}), and introduce a minimal representation of unit three-vectors (Section \ref{sec:unit-vectors}).


\subsection{3D Rigid Body Transformations} \label{sec:rigid-transformations}

We represent the relative position and orientation of the spacecraft---its relative \textit{pose}---as an element of the Special Euclidean group $\mathrm{SE}(3)$ by a matrix
\begin{equation}
    T_{\fB\fS} \triangleq
    \begin{bmatrix}
    R_{\fB\fS} & \vvec{r}^\fB_{\oS\oB} \\
    \bm{0}_{1 \times 3}      & 1
    \end{bmatrix}, 
\end{equation}
where $R_{\fB\fS} \in \mathrm{SO}(3)$ is the orientation of some body-fixed frame of the small body $\fB$ with respect to a spacecraft body-fixed frame $\fS$, and $\vvec{r}^\fB_{\oS\oB} \in \mathbb{R}^3$ is the position of the spacecraft's origin with respect to the origin of $\fB$, expressed in $\fB$. 
Moreover, the \textit{fixed} pose $T_{\fS\fC}$ of the onboard camera relative to the body-fixed frame of the spacecraft is precisely known \textit{a priori}, which can be used to derive \textit{the camera's relative pose} $T_{\fB \fC} = T_{\fB \fS}T_{\fS \fC}$.

Optimization of poses $T \in \mathrm{SE}(3)$ can be parameterized in terms of the \textit{local coordinates} $\vvec{\zeta} = [\vvec{\gamma}^\top\, \vvec{\tau}^\top]^\top \in \mathbb{R}^6$~\cite{forster2016manifold,absil2007trust} by defining a \textit{retraction} $\mathfrak{R}_T: \mathrm{SE}(3)\times \mathbb{R}^6 \rightarrow\mathrm{SE}(3)$ using the exponential map (at the identity) of the $\mathrm{SO}(3)$ group of rotations and the \textit{hat} operator (see~\cite{chirikjian2011}):
\begin{align} \label{eq:ret-pose}
	\mathfrak{R}_T(\vvec{\gamma}, \vvec{\tau}) \triangleq 
	T
	\begin{bmatrix}
		\Exp{\vvec{\gamma}} & \vvec{\tau} \\ 
		\vvec{0}_{1 \times 3}             & 1
	\end{bmatrix} =
        \begin{bmatrix}
		R\Exp{\vvec{\gamma}} & \vvec{r} + R\vvec{\tau} \\ 
		\vvec{0}_{1 \times 3}             & 1
	\end{bmatrix} \in \mathrm{SE}(3).
\end{align}
This reparameterization is referred to as \textit{lifting}~\cite{absil2007trust}. 
A useful first-order approximation of the exponential map is $\Exp{\vvec{\gamma}} \approx I_3 + \sk{\vvec{\gamma}}$. 
The uncertainty of the camera's pose can be defined in a similar way as
\begin{align}
	T_{\fB\fC} \triangleq 
	\overline{T}_{\fB\fC} 
	\begin{bmatrix}
		\Exp{\vvec{\omega}} & \vvec{\nu} \\ 
		\vvec{0}_{1 \times 3}             & 1
	\end{bmatrix},
\end{align}
where $\vvec{\omega} \sim \mathcal{N}\left( \vvec{0}_{3 \times 1}, \Sigma_R\right)$, $\vvec{\nu} \sim \mathcal{N}\left( \vvec{0}_{3 \times 1}, \Sigma_r\right)$, and $\overline{T}_{\fB\fS}$ is the \textit{actual} pose~\cite{dellaert2017,forster2016manifold}. 
Therefore, the estimated pose $T_{\fB\fC}$ is represented by the uncertain orientation $R_{\fB\fC} \triangleq \overline{R}_{\fB\fC}\Exp{\bm{\omega}}$ and the uncertain position $\vvec{r}^{\fB}_{\oC\oB} \triangleq \overline{\vvec{r}}_{\oC\oB}^{\fB} + \overline{R}_{\fB\fC}\vvec{\nu}$. 


\subsection{The Unit 2-Sphere} \label{sec:unit-vectors}

An important two-dimensional manifold is the unit 2-sphere $\mathbb{S}^2 \triangleq \left\{ \uvec{x} \in \mathbb{R}^3 \given \|\uvec{x}\| = 1\right\}$, i.e., the topological space composed of all unit vectors in $\mathbb{R}^3$.
The tangent space $\mathfrak{T}_{\uvec{x}}\left(\mathbb{S}^2\right)$ at a point $\uvec{x} \in \mathbb{S}^2$ is defined as the set of all three-vectors tangent to $\mathbb{S}^2$ at $\uvec{x}$: 
\begin{equation}
    \mathfrak{T}_{\uvec{x}}\left(\mathbb{S}^2\right) \triangleq \left\{\vvec{y} \in \mathbb{R}^3 \given \uvec{x}^\top\vvec{y} = 0, \uvec{x} \in \mathbb{S}^2\right\}.
\end{equation}
For any $\vvec{y} \in \mathfrak{T}_{\uvec{x}}\left(\mathbb{S}^2\right)$, we can write $\vvec{y} = B_{\uvec{x}}\vvec{\xi}$ where $\vvec{\xi} \in \mathbb{R}^2$ lies in the plane tangent to $\mathbb{S}^2$ at $\uvec{x}$ defined by the basis vectors defined in the columns of the matrix $B_{\uvec{x}} \in \mathbb{R}^{3 \times 2}$. 
With these definitions, we can define another useful retraction $\mathfrak{R}_{\uvec{x}}(\vvec{\xi})$ as follows~\cite{dellaert2017}:
\begin{equation} \label{eq:ret-unit}
    \mathfrak{R}_{\uvec{x}}(\vvec{\xi}) \triangleq \cos(\|B_{\uvec{x}}\vvec{\xi}\|)\uvec{x} + \sin(\|B_{\uvec{x}}\vvec{\xi}\|)\frac{B_{\uvec{x}}\vvec{\xi}}{\|B_{\uvec{x}}\vvec{\xi}\|} \in \mathbb{S}^2.
\end{equation}
A useful first-order approximation to the above retraction is $\mathfrak{R}_{\uvec{x}}(\vvec{\xi}) \approx \vvec{x} + B_{\uvec{x}}\vvec{\xi}$. 
This minimal representation allows for the optimization of the unit vector $\uvec{x} \in \mathbb{S}^2$ with respect to the local coordinates $\vvec{\xi} \in \mathbb{R}^2$ according to the basis $B_{\uvec{x}}$. 
Uncertainty in the unit vector can also be defined in the local coordinate system defined by $B_{\overline{\vvec{x}}}$ at the true value $\overline{\uvec{x}}$, i.e., $\uvec{x} = \mathfrak{R}_{\overline{\uvec{x}}}\left(\vvec{\varepsilon}\right)$, where $\vvec{\varepsilon} \sim \mathcal{N}\left(\vvec{0}_{2\times 1}, \Sigma_\xi\right)$.
We will use this formulation, implemented in the Georgia Tech Smoothing and Mapping (GTSAM) library~\cite{dellaert2012}, to represent and optimize surface normal estimates (detailed in Section \ref{sec:photoclinometry}). 

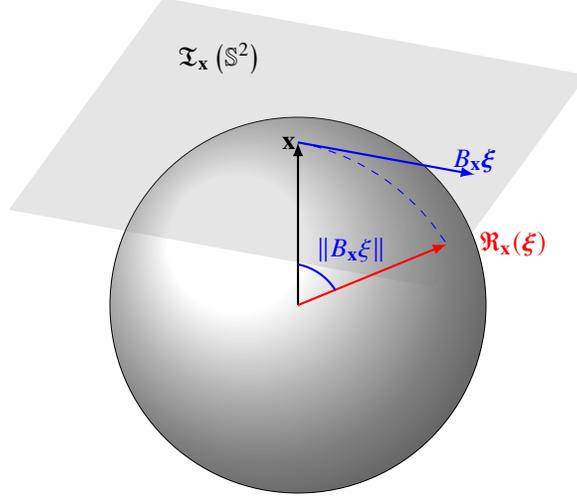
\begin{figure}
    \centering
    \input{figures/unit-sphere}
    \caption{Unit vector retraction.}
    \label{fig:unit-sphere}
\end{figure}

%% file: figures/unit-sphere.tex
%
\begin{center}
\tdplotsetmaincoords{60}{110}
\begin{tikzpicture}[scale=2.5,tdplot_main_coords]
    \tdplotsetrotatedcoords{20}{60}{0}
    \draw [ball color=white,very thin,tdplot_rotated_coords] (0,0,0) circle (1) ;
    \fill[black!50,opacity=0.2] (-1.2,-1.2,1) -- (-1.2,1.2,1) -- (1.2,1.2,1) -- (1.2,-1.2,1) -- (-1.2,-1.2,1);
    \node at (-.7, -.7, 1) {$\mathfrak{T}_{\uvec{x}}\left(\mathbb{S}^2\right)$};
    
    \node at (0, .3, .4) [blue] {$\|B_{\uvec{x}}\xi\|$};
    

    \draw[thick,->] (0,0,0) -- (0,0,1) node [black,left] {$\uvec{x}$};
    \draw[thick,blue,->] (0,0,1) -- (0,1,1) node [blue,above] {$B_{\uvec{x}}\vvec{\xi}$};
    \draw[thick,red,->] (0,0,0) -- (0,.8415,.5403) node [red,right] {$\quad\, \mathfrak{R}_{\uvec{x}}(\vvec{\xi})$};

    \tdplotdefinepoints(0,0,0)(0,0,1)(0,.8415,.5403)
    \tdplotdrawpolytopearc[dashed, blue]{1}{}{}
    \tdplotdefinepoints(0,0,0)(0,0,1/3)(0,.8415/3,.5403/3)
    \tdplotdrawpolytopearc[thick, blue]{1/4}{}{}
    
\end{tikzpicture}
\end{center}

%% file: text/methods.tex
\begin{figure}[htb]
    \centering
    \includegraphics[width=\linewidth]{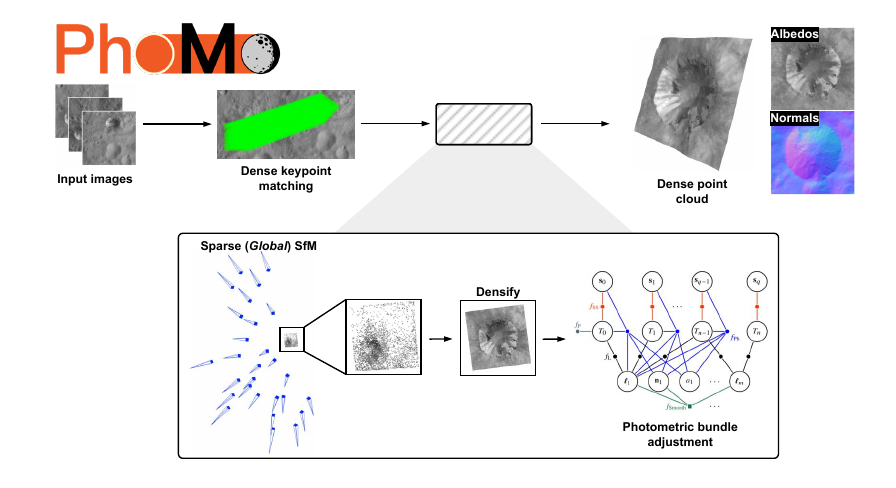}
    \caption{PhoMo Overview.}
    \label{fig:phomo-flowchart}
\end{figure}

First, we introduce the SfM problem (Section \ref{sec:vslam}). 
Next, we formulate SfM as a maximum \textit{a posteriori} (MAP) inference problem using the formalism of \textit{factor graphs} (Section \ref{sec:factor-graphs}). 
Finally, we present the photoclinometry framework and detail its integration into the proposed system (Sections~\ref{sec:photoclinometry}, \ref{sec:sun-sensor}, and \ref{sec:smoothness}).
An overview of the proposed approach is shown in Fig. \ref{fig:phomo-flowchart}.


\subsection{Structure-from-Motion} \label{sec:vslam}

The proposed formulation will leverage SfM to estimate the spacecraft's relative pose and a landmark map of the small body's surface.
Feature-based SfM and SLAM~\cite{cadena2016tr} leverage monocular images taken from multiple viewpoints along a trajectory to jointly estimate the robot's pose with respect to its environment and construct a 3D model of the scene.
The SfM architecture is typically comprised of two main components: the \textit{front-end} and the \textit{back-end}.
The front-end extracts 2D interest points (\textit{keypoints}) from images, represents each keypoint with a local feature \textit{descriptor}, and matches keypoints between images by comparing their associated descriptors~\cite{driver2022astrovision}.
The front end also performs \textit{data association} by associating the 2D keypoint measurements with specific points in 3D space (the \textit{landmarks}). 
Finally, the associations from the front-end are used to \emph{simultaneously} reconstruct a map of the environment and resolve the pose of the camera through inference in the \textit{back-end} via maximum \textit{a posteriori} (MAP) estimation. 


Formally, let $\fB$ denote some body-fixed frame of the small body with origin $\oB$, and let $\fC_k$ denote the camera frame at time index $k$ with origin $\oC_k$.
Moreover, let $\vvec{\ell}^\fB_j = \left[\ell_{x,j}^\fB\ \ell_{y,j}^\fB\ \ell_{z,j}^\fB\right]^\top \in \mathbb{R}^3$ denote the vector from $\oB$ to the $j$th surface landmark expressed in $\mathcal{B}$,
let $\vvec{q}^{\fC_k}_j = \left[q_{x,j}^{\fC_k}\ q_{y,j}^{\fC_k}\ q_{z,j}^{\fC_k}\right]^\top \in \mathbb{R}^3$ denote the vector from $\oC_k$ to the $j$th landmark expressed in $\fC_k$, 
and let $\vvec{p}_{j,k} = \left[u_{j,k}\ v_{j,k}\right]^\top \in \mathbb{R}^2$ denote the 2D image coordinates of the $j$th landmark observed by camera $\mathcal{C}_k$, i.e., the keypoint.
SfM seeks the MAP estimate of the camera poses $\mathcal{T} := \left\{T_{\fB\fC_k}\in \mathrm{SE}(3) \mid k = 0,\ldots,n\right\}$ and the collection of landmarks (the \textit{map}) $\mathcal{L} := \{\vvec{\ell}^\fB_j \in \mathbb{R}^3 \mid j = 1,\ldots,m\}$ given the (independent) keypoint \textit{measurements} $\mathcal{P} := \{\Hat{\vvec{p}}_{j,k} \in \mathbb{R}^2 \mid k=0,\ldots,n, j=1,\ldots,m\}$:
\begin{align}
    \mathcal{T}^*, \mathcal{L}^* &= \argmax_{\mathcal{T}, \mathcal{L}} p\left(\mathcal{T}, \mathcal{L}\mid \mathcal{P}\right) \\
    &\propto \argmax_{\mathcal{T}, \mathcal{L}} p\left(\mathcal{T}, \mathcal{L}\right) p\left(\mathcal{P} \mid \mathcal{T}, \mathcal{L}\right) \\
    &= \argmax_{\mathcal{T}, \mathcal{L}} p\left(\mathcal{T}, \mathcal{L}\right) \prod_{k} \prod_{j} p\left(\Hat{\vvec{p}}_{j,k} \mid T_{\fB\fC_k}, \vvec{\ell}^\fB_j\right). \label{eq:map}
\end{align}
Note that the SfM solution is innately expressed in some arbitrary body-fixed frame since most SfM techniques assume operation in a static scene, typically referred to as the ``world'' frame~\cite{cadena2016tr}. 
By assuming that the measurements $\Hat{\vvec{p}}_{j,k}$ are corrupted by zero-mean Gaussian noise, i.e., $\Hat{\vvec{p}}_{j,k} = \overline{\vvec{p}}_{j,k} + \vvec{\eta}_{j,k}$ where $\vvec{\eta}_{j,k} \sim \mathcal{N}(\vvec{0},\Sigma_{j,k})$, we get 
\begin{equation}
    p\left(\Hat{\vvec{p}}_{j,k} \mid T_{\fB\fC_k}, \vvec{\ell}^\fB_j\right) \propto \exp\left\{-\frac{1}{2}\|\Pi\left(\vvec{\ell}^\mathcal{B}_{j}, T_{\fB\fC_k}; K\right) - \Hat{\hvec{p}}_{j,k}\|_{\Sigma_{j,k}}^2\right\},
\end{equation}
where $\|\vvec{e}\|^2_{\Sigma} := \vvec{e}^\top\Sigma^{-1}\vvec{e}$, and the \textit{forward-projection} function $\Pi$ relates landmarks $\vvec{\ell}^\mathcal{B}_{j}$ to their (homogenous) coordinates $\hvec{p}_{j,k}$ in the $k$th image, i.e., 
\begin{align} \label{eq:fproj}
    \hvec{p}_{j,k} = \Pi\left(\vvec{\ell}^\fB_{j}, T_{\fB\fC_k}; K\right) &= \frac{1}{d_{j}^{\fC_k}}
    \left[K\,|\,\bm{0}^{3\times1}\right] T_{\fB\fC_k}^{-1} \hvec{\ell}^{\mathcal{B}}_j = \frac{1}{d_{j}^{\fC^k}} K \vvec{q}^{\fC_k}_j,
\end{align}
where $d_{j}^{\fC_k} = q_{z,j}^{\fC_k}$ is the landmark depth in $\fC_k$, $\hvec{\ell}^{\mathcal{B}}_{j} = \left[\left(\vvec{\ell}^\fB_j\right)^\top\ 1\right]^\top \in \mathbb{P}^3$ and $\hvec{p}_{j,k} = \left[\left(\vvec{p}_{j,k}\right)^\top\ 1\right]^\top \in \mathbb{P}^2$ denote the homogeneous coordinates of $\vvec{\ell}^{\fB}_j$ and $\vvec{p}_{j,k}$, respectively, and $K$ is the camera calibration matrix:
\begin{equation}
K = 
    \begin{bmatrix}
    f_x & 0 & c_x \\
    0 & f_y & c_y \\
    0 & 0 & 1
    \end{bmatrix},
\end{equation}
where $f_x$ and $f_y$ are the \textit{focal lengths} in the $x$- and $y$-directions of the camera frame, and $(c_x, c_y)$ is the \textit{principal point} of the camera.

Finally, the MAP estimate can be formulated as the solution to a nonlinear least-squares problem by taking the negative logarithm of \eqref{eq:map}:
\begin{equation} \label{eq:ba}
    \mathcal{T}^*, \mathcal{L}^* = \argmin_{\mathcal{T}, \mathcal{L}} \sum_{k,j} \|\Pi\left(\vvec{\ell}^\fB_j, T_{\fB\fC_k}; K\right) - \Hat{\hvec{p}}_{j,k}\|_{\Sigma_{j,k}}^2,
\end{equation}
where we have omitted the priors $p\left(\mathcal{T}, \mathcal{L}\right)$ for conciseness and generality, which can be ignored if no prior information is assumed (i.e., $p\left(\mathcal{T}, \mathcal{L}\right) = const.$) or can encode relative pose constraints via known dynamical models~\cite{dor2022astroslam}.
This process is commonly referred to as \textit{Bundle Adjustment} (BA)~\cite{schonberger2016structure}. 
Note that the SPC optimization process decouples estimation of the poses and the landmarks, i.e., landmark position and camera pose estimates are passed back-and-forth between the pose determination and DTM construction steps, respectively, until convergence~\cite{gaskell2008}. 

In this work, we use Georgia Tech's Structure-from-Motion (GTSfM) library~\cite{gtsfmcvpr} to generate a sparse SfM solution, followed by a densification step to construct an initial dense map of the imaged surface. 
The reasoning behind the initial sparse solution is two-fold: (1) leveraging \textit{sparse} correspondences, as opposed to the per-pixel matches, significantly reduces the number of computations in the redundant two-view estimation step; (2) by leveraging only the most confident matches, we reduce the risk of incorporating any outlier matches into the estimation scheme as dense matches can be more reliably verified during the densification step before being added to the map. 
The keypoint measurements and correspondences, $\Hat{\vvec{p}}_{j,k}$, are computed by a state-of-the-art, \textit{autonomous} keypoint detection and matching method based on deep learning, i.e., RoMa~\cite{edstedt2024roma}. 
Next, densification of the sparse GTSfM solution is conducted by computing the squared Sampson error~\cite{sampson1982fitting} of each of the putative correspondences and adding the match if its error is below 1. 
The dense map is then triangulated from the 2D keypoint measurements using the Direct Linear Transform (DLT)~\cite[Chapter~4.1]{hartley2003multiple}. 

This dense map, and the associated camera poses from the sparse solution, are used to initialize the graph-based photoclinometry step described in Section \ref{sec:photoclinometry}.
Image brightness measurements are extracted at observed keypoints and combined with a prior reflectance model to determine the surface normal and albedo at the estimated landmarks (see Section \ref{sec:implementation-details} for more details). 
The back-end leverages factor graphs to represent the MAP estimation problem, which will be discussed in Section \ref{sec:factor-graphs}. 


\subsection{Factor Graphs} \label{sec:factor-graphs}

\begin{figure}[tb!]
\centering
\begin{subfigure}[t]{\linewidth}
  \centering
  \input{figures/factor-graph-sfm.tex}
  \caption{Typical factor graph formulation of the SfM problem.}
  \label{fig:factor-graph-sfm}
\end{subfigure}\\
\vspace{10pt}
\begin{subfigure}[t]{\linewidth}
  \centering
  \input{figures/factor-graph-spc.tex}
  \caption{With proposed factors \textcolor{blue}{$f_{\mathrm{Ph}}$} relating to photoclinometry constraints, \textcolor{coquelicot}{$f_{\mathrm{SS}}$} relating to Sun sensor measurements, and \textcolor{darkspringgreen}{$f_{\mathrm{Smooth}}$} relating to local smoothness constraints.}
  \label{fig:factor-graph-ps}
\end{subfigure}%
\caption{Variable nodes are camera poses $T_k \in \mathrm{SE}(3)$, landmarks $\bm{\ell}_j \in \mathbb{R}^3$, sun vectors $\mathbf{s}_l \in \mathbb{S}^2$, surface normals $\mathbf{n}_i \in \mathbb{S}^2$, and surface albedos $a_i \in [0, 1]$. Factor nodes $f_\mathrm{L}$ and \textcolor{cadet}{$f_\mathrm{P}$} relate to keypoint-based landmark measurements and possibly a prior factor, respectively.}
\label{fig:factor-graph}
\end{figure}
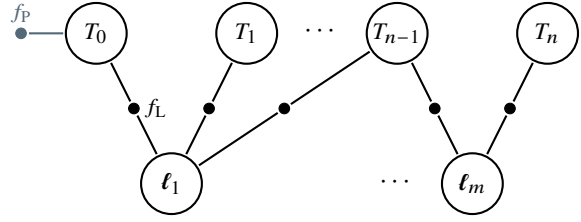
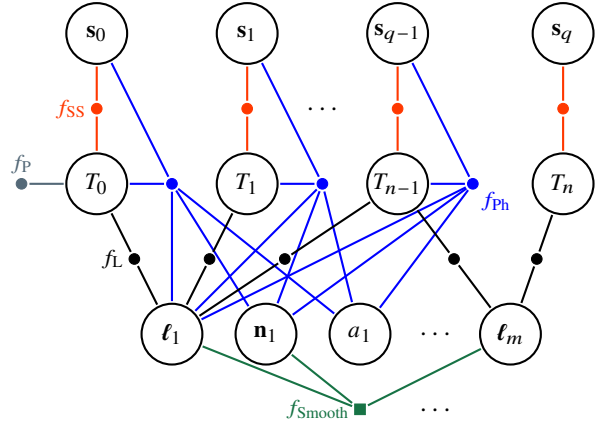

We leverage the formalism of factor graphs to facilitate the fusion of keypoint and Sun vector measurements through small body reflectance models to \textit{simultaneously} estimate the camera poses, landmark positions, and surface normals and albedos at each landmark. 
Formally, a factor graph is a bipartite graph $G = \left(\mathcal{F}, \Theta, \mathcal{E}\right)$ with \textit{factor nodes} $f_i \in \mathcal{F}$ that abstract the measurements and prior knowledge $z_i \in \mathcal{Z}$ as generalized probabilistic constraints between \textit{variable nodes} $\theta_j \in \Theta$, the unknown random variables, where \textit{edges} $e_{i,j} \in \mathcal{E}$ define the interdependence relationships between a factor $f_i$ and a variable $\theta_j$.
With these definitions, a factor graph $G$ defines a factorized function
\begin{equation} \label{eq:f}
    f(\Theta) = \prod_i f_i\left(\Theta_i\right),
\end{equation}
where each measurement factor $f_i(\Theta_i) = l(\Theta_i; z_i)$ of the variables $\Theta_i = \{\theta_j \in \Theta \given e_{i,j} \in \mathcal{E}\}$, with likelihood $l\left(\Theta_i; z_i\right) \propto p\left(z_i \given \Theta_i\right)$, and each prior factor $f(\Theta_i) = p(\Theta_i)$ represents a term in the joint probability density function (PDF), i.e., $f(\Theta) \propto p\left(\Theta \given \mathcal{Z}\right)$. 
We seek the variable assignment $\Theta^*$ through maximum \textit{a-posteriori} (MAP) inference over the joint probability distribution encoded by the factors in the factor graph: 
\begin{equation} \label{eq:Theta_opt}
    \Theta^* = \argmax_\Theta \prod_i f_i\left(\Theta_i\right).
\end{equation}
Assuming a zero-mean Gaussian noise model with measurement covariance $\Sigma_i$,
yields factors of the form 
\begin{equation} \label{eq:fi}
    f_i(\Theta_i) \propto \exp\left\{-\frac{1}{2} \|h_i(\Theta_i) - z_i\|^2_{\Sigma_i}\right\},
\end{equation}
where $h_i(\cdot)$ is a measurement prediction function. 
Moreover, we assume that the priors $p(\Theta_i)$ take the form $p(\Theta_i) \propto \exp\left\{-\frac{1}{2} \|h_i(\Theta_i) - z_i\|^2_{\Sigma_i}\right\}$ with prior mean and covariance $z_i$ and $\Sigma_i$, respectively. 
Therefore, solving \eqref{eq:Theta_opt} is equivalent to minimizing the sum of nonlinear least-squares via
\begin{equation} \label{eq:logf}
    \argmin_\Theta(-\log f(\Theta)) = \argmin_\Theta \frac{1}{2}\sum_i \|h_i(\Theta_i) - z_i\|^2_{\Sigma_i}.
\end{equation}
This formulation allows factor graphs to support PDFs or cost functions of any number of variables~\cite{kaess2012}, allowing for the inclusion of multiple sensor modalities, as well as prior knowledge and constraints to uniquely determine the MAP solution for the unknown variables $\Theta^*$. 
The typical factor graph formulation of SfM is shown in Fig.~\ref{fig:factor-graph-sfm}, where the factors $f_\mathrm{L}\left(\vvec{\ell}^\fB_k, T_{\fB\fC_k}; K\right)$ relate to the forward-projection error function defined in Equation \eqref{eq:ba}. 

Solving the nonlinear least-squares problem in \eqref{eq:logf} typically involves repeated linearization. 
For nonlinear measurement functions $h_i(\cdot)$, nonlinear optimization approaches such as the Levenberg-Marquardt algorithm (LMA) leverage repeated first-order linear approximations to \eqref{eq:logf} to approach the minimum. 
In addition, the interdependence relationships encoded by the edges of the factor graph capture the factored nature of the PDF and sparsity of the underlying information matrix, allowing for \textit{exact} nonlinear optimization in an \textit{incremental} setting by exploiting the sparse edge connections to identify the variables to be optimized when a new measurement becomes available~\cite{kaess2012}. 
The factor graph formulation of the proposed keypoint-based SPC problem is implemented using the Georgia Tech Smoothing and Mapping (GTSAM) library~\cite{dellaert2012}, an estimation toolbox based on factor graphs pioneered at Georgia Tech. 
The toolbox provides a fully customizable framework for factor graph construction and a suite of nonlinear optimization methods. 

\subsection{Small Body Photometry} \label{sec:photometry}

\begin{figure}[t]
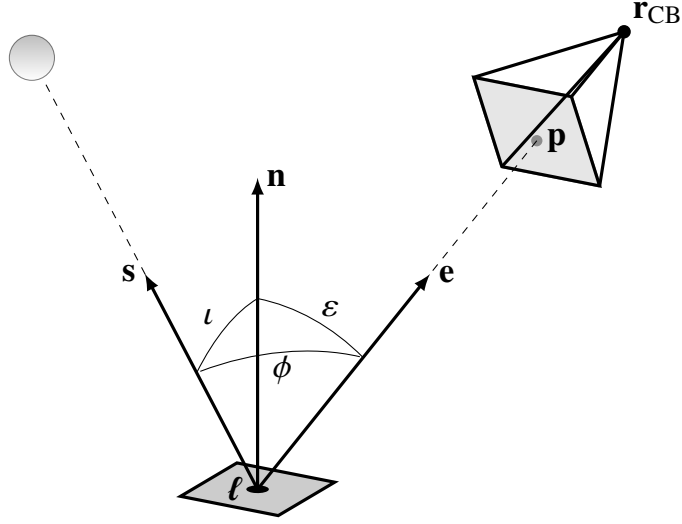

\centering
\include{figures/photometry-conventions}
\vspace{-20pt}
\caption{Photometry conventions.}
\label{fig:angles}
\end{figure}

\textit{Photogrammetry}, serving as the theoretical foundation for contemporary techniques such as SfM, primarily concentrates on establishing geometric relationships between points in an image (the \textit{keypoints}) and the corresponding points in the scene (the \textit{landmarks}). 
In contrast, \textit{photometry} seeks to model the observed ``brightness'' of a scene point in an image as influenced by its surface topography and material properties.
Photogrammetry has historically garnered more attention in the computer vision community, due, in part, to the relatively low complexity as compared to photometry.
Indeed, photometric modeling is inherently complex. 
This is especially true for terrestrial applications, which typically require the use of deep neural networks to achieve accurate photometric reconstructions~\cite{mildenhall2020nerf}. 
However, operation in space presents us with a number of advantages that simplify the photometric modeling process. 
First, we may treat the Sun as a point source delivering collimated light to the surface owing to the large distance to the Sun (i.e., the Sun subtends an angle of $0.5^\circ$ at Earth~\cite{shepard2017}) and the lack of atmosphere of most small bodies to scatter the incoming light. 
Second, the direction of the incoming light can be precisely measured using typical onboard instrumentation (e.g., Sun sensors, star trackers). 
Third, previous photometric modeling of small bodies has demonstrated that \textit{global} reflectance functions, as opposed to \textit{spatially-varying} reflectance functions, can precisely estimate the observed brightness across the surface of a target small body.
Moreover, these global reflectance properties are similar across asteroids of the same taxonomic class~\cite{li2015}.

More formally, an image may be considered as a mapping $I \colon \Omega \rightarrow \mathbb{R}_+$ over the pixel domain $\Omega \subset \mathbb{R}^2$ that maps a point in the image, $\vvec{p} \in \Omega$, to its corresponding ``brightness" value $I(\vvec{p}) \in \mathbb{R}_+$. 
Here, ``brightness'' refers to the fact that the image values correspond to the amount of light falling on the photodetector inside the camera, referred to as the image \textit{irradiance} with units of power per unit area (W$\cdot$m$^{-2}$). 
Specifically, each brightness value $I(\vvec{p})$ is initially represented by a digital number (DN), which is computed by converting the charge accumulated by the photodetector over some exposure time $\Delta t$ to an integer DN using an analog-to-digital converter. 
For example, the Dawn mission to Asteroid 4 Vesta and Minor Planet 1 Ceres digitized the signal from the framing camera to a 14-bit integer DN~\cite{sierks2011dawn}. 
An important requirement for deep space imagers is response linearity, i.e., an almost perfectly linear relationship between the incident irradiance on the detector and the quantized charge rate $\mathrm{DN} / \Delta t$~\cite[Chapter 7]{christian2025}. 
This linear relationship is characterized by a rigorous \textit{radiometric calibration} process conducted both on the ground and during flight~\cite{sierks2011dawn,schroder2013cal,schroder2014cal}. 
It can be shown~\cite[Chapter~10.3]{horn1986} that the image irradiance is proportional to the \textit{radiance}, with units of power per unit solid angle per unit area (W$\cdot$sr$^{-1}$$\cdot$m$^{-2}$), reflected towards the camera from the surface. 
Thus, each point $\vvec{p} \in \Omega$ corresponds to the radiance emitted from a point (or, more precisely, an infinitesimal patch) on the surface of the body. 

The emitted radiance from the surface, $L(\sang, \eang, \phi, a)$ and the incident (collimated) irradiance from the Sun, $F$, which is inversely related to the square of the distance to the Sun~\cite{shepard2017}, are related by the \textit{bidirectional reflectance function}, $r$ (in units of $\text{sr}^{-1}$):
\begin{equation} \label{eq:brf}
    \frac{L(\sang, \eang, \phi, a)}{F} = r(\sang, \eang, \phi, a),
\end{equation}
where $a$ is the surface albedo, $\sang$ is the angle between the incoming light and the surface normal, or the \textit{incidence angle}, $\eang$ is the angle between the emitted light and the surface normal, or the \textit{emission angle}, and $\phi$ is the angle between the emitted light and the incoming light, or the \textit{phase angle} (see Fig. \ref{fig:angles}). 
A similar measure of reflectance, which is very popular in the context of planetary photometry~\cite{shepard2017, li2015}, is the \textit{bidirectional radiance factor}, $r_\mathrm{F}$, which is the ratio of the bidirectional reflectance of a surface to that of a perfect Lambertian surface illuminated and viewed from overhead (i.e., $\sang = \eang = \phi = 0$):
\begin{equation} \label{eq:rad-factor}
    r_\mathrm{F}(\sang, \eang, \phi, a) = \pi r(\sang, \eang, \phi, a) = \frac{\pi L(\sang, \eang, \phi, a)}{F}. 
\end{equation}
While the radiance factor is dimensionless, it is common to refer to its value as being in units of reflectance or $L/F$ (or, more commonly, $I/F$ when $I$ is used to denote the radiance). 
Henceforth, we will refer to the radiance factor when mentioning the reflectance function. 

If we normalize the radiance factor by the radiance factor when observed and illuminated from overhead ($\iota = \varepsilon = \phi = 0^\circ$), we get the \textit{photometric function}, $\rho(\sang, \eang, \phi)$:
\begin{equation}
    \rho(\sang, \eang, \phi) = \frac{r_\mathrm{F}(\sang, \eang, \phi)}{r_\mathrm{F}(0, 0, 0)} \Leftrightarrow r_\mathrm{F}(\sang, \eang, \phi) = a_\mathrm{n} \rho(\sang, \eang, \phi), 
\end{equation}
where $a_\mathrm{n} := r_\mathrm{F}(0, 0, 0)$ is the \textit{normal albedo}. 
Finally, the photometric function may be factorized into the \textit{phase function}, $\Lambda(\phi)$, and the disk function, $d(\sang, \eang, \phi)$~\cite{shkuratov2011optical}:
\begin{equation}
    r_\mathrm{F}(\sang, \eang, \phi) = a_\mathrm{n} \Lambda(\phi) d(\sang, \eang, \phi). 
\end{equation}
The phase function, normalized to unity at $\phi = 0^\circ$, models the phase-dependent brightness variations that are independent of the incident and emission angles, referred to as the \textit{opposition effect}~\cite[Chapter 9]{hapke2012theory}, where $\Lambda(\phi)$ is often represented as an $n$th-order polynomial with parameters fit to imagery of the specific target body~\cite{schroder2013resolved,schroder2014cal}.
The disk function models brightness variations due to the underlying topography (which may also be a function of the phase angle). 
Henceforth, the term albedo will refer to the normal albedo and be denoted by $a$. 

We will also consider the case where the image values have not been radiometrically calibrated to units of reflectance. 
Recalling the response linearity of the photodetector, we may rewrite Equation \eqref{eq:rad-factor} as
\begin{equation}
    \frac{\pi L(\sang, \eang, \phi)}{F} \propto \frac{\mathrm{DN}(\sang, \eang, \phi)}{\Delta t} = \Tilde{a}_\mathrm{n} \lambda d(\sang, \eang, \phi) + \xi. 
\end{equation}
The scale term $\lambda$ accounts for the phase dependent brightness in the absence of an explicit phase function, the incident solar flux, and the lack of radiometric calibration to radiance. 
The bias term $\xi$ may be included to account for possible background noise~\cite{gaskell2008,park2019spc}. 
The \textit{relative} albedo $\Tilde{a}_\mathrm{n}$ is only proportional to the absolute albedo since compensation for the lack of a radiometric calibration may also be expressed through scaling of the albedos.
We refer to this case as \textit{uncalibrated}.

\begin{table}[]
    \centering
    \ra{2.0}
    \caption{Investigated reflectance functions. \normalfont{The coefficients for $g(\phi)$ and $\Lambda(\phi)$ model are listed in Table \ref{tab:sch-coeff}.}}
    \begin{adjustbox}{width=\linewidth}
    \input{figures/rad-factors} 
    \end{adjustbox}
    \label{tab:rad-factors}
\end{table}

\begin{table}[]
    \centering
    \ra{1.5}
    \caption{Coefficients for each reflectance model proposed by Schr\"{o}der et al.~\cite{schroder2013resolved,schroder2017resolved} \normalfont{We have normalized $c_i$, $i=0,\ldots,m$, such that $c_0 = 1$, or, equivalently, $\Lambda(0) = 1$.}}
    \begin{adjustbox}{width=\linewidth}
    \input{figures/rad-factors-coeffs}
    \end{adjustbox}
    \label{tab:sch-coeff}
\end{table}

We consider five photometric functions in this investigation, defined in Table \ref{tab:rad-factors}.
The \textit{McEwen} model (Equation \eqref{eq:ref-mcewen}) features a combination of Lambert and Lommel-Seeliger photometric functions weighted according to an exponential function of the phase angle, the \textit{phase weighting function} $g(\phi)$, which was fit to lunar imagery captured by the Galileo spacecraft~\cite{mcewen1991, mcewen1996precise}. 
This model was chosen because it is nominally used by SPC and has been shown to be well-suited for photometry on a wide range of small bodies~\cite{gaskell2008, raymond2011dawn, barnouin2020, gaskell2023psj}. 
Note that we leverage the exponential approximation to McEwen's original polynomial weighting function proposed in \cite{gaskell2008}. 
Similarily, the \textit{Lunar-Lambert} model (Equation \eqref{eq:ref-llambert}), a generalization of the McEwen model, again features a combination of Lambert and Lommel-Seeliger photometric functions, but is weighted according to an \textit{affine} function of the phase angle. 

The \textit{Akimov} model (Equation \eqref{eq:ref-akimov}) is a parameter-free model derived from the formal condition that an extremely rough surface subjected to small random undulations, i.e., random deviations or fluctuations in its geometry, maintains the same reflectance as before the undulations~\cite{shkuratov1994principle}. 
Although such perfectly rough surfaces are rare in nature, they serve as a useful abstraction that has been shown to accurately approximate lunar reflectance. 
The Akimov model depends on the photometric latitude $\beta$, the angle between the surface normal and the \textit{scattering plane} (i.e., the plane containing the light source, the landmark, and the observer), and longitude $\gamma$, the angle in the scattering plane between projection of the normal and the vector from the landmark to the observer. 
These values are related to the incidence and emission angle as follows~\cite{kreslavsky2000photometric}:
\begin{align}
    \tan\gamma &= \frac{\cos\iota / \cos\varepsilon - \cos\phi}{\sin\phi}, \\
    \cos\beta &= \frac{\cos\varepsilon}{\cos\gamma}.
\end{align}
The Akimov model was also extended to include a phase weighting term, which we refer to as the Akimov+ model~\cite{shkuratov2011optical,schroder2013resolved}. 
Finally, the \textit{Minnaert} model (Equation \eqref{eq:ref-minnaert}) is a generalization of Lambertian reflectance that includes dependence on the emission angle according to the phase weighting function~\cite{minnaert1941reciprocity,hapke2012theory}. 
The coefficients for the Akimov+, Lunar-Lambert, and Minnaert models were independently fit to approach imagery of both Asteroid 4 Vesta and Minor Planet 1 Ceres captured during the Dawn mission by Schr\"{o}der et al.~\cite{schroder2013resolved,schroder2017resolved} and are provided in Table \ref{tab:sch-coeff}. 

We  discuss how these photometric principles can be incorporated into a feature-based SfM system in the following section.
We consider both calibrated and uncalibrated imagery in this investigation. 
For the uncalibrated case, we use corrected DN values where various error sources such as dark current and readout smear have been removed~\cite{schroder2013cal,schroder2014cal}. 
For the calibrated case, the images have been converted to units of $L/F$ according to the radiometric calibration process detailed in \cite{schroder2014cal}.


\subsection{Photoclinometry Constraints} \label{sec:photoclinometry}

Photoclinometry techniques are integrated into the feature-based SfM system to estimate surface normals and albedos at estimated landmarks. 
Photoclinometry~\cite{woodham1980} is the process of determining surface gradients of an object by observing it from different viewpoints and lighting conditions and is leveraged by SPC to facilitate dense surface reconstruction. 
As before, let an image taken at time index $k$ be denoted by $I_k \colon \Omega \rightarrow \mathbb{R}_+$ over the pixel domain $\Omega \subset \mathbb{R}^2$.
The measured image brightness $\Hat{I}_k(\vvec{p}_{j,k})$ (calibrated to units of $L/F$) at a keypoint $\vvec{p}_{j,k} \in \Omega$ in image $I_k$ associated with a landmark $\bm{\ell}_j \in \mathbb{R}^3$ can be modeled by an appropriate reflectance function, as detailed in Section \ref{sec:photometry}:
\begin{equation} \label{eq:Ik_angles}
    I(\sang_{j,k}, \eang_{j,k}, \phi_{j,k}, a_j) = a_j \Lambda(\phi_{j,k}) d(\sang_{j,k}, \eang_{j,k}, \phi_{j,k}),
\end{equation}
where $a_j$ is the albedo at landmark $\vvec{\ell}_j$ and $\sang_{j,k}$, $\eang_{j,k}$, and $\phi_{j,k}$  are the incidence, emission, and phase angles, respectively, at landmark $\vvec{\ell}_j$ in the $k^{th}$ image.
When considering \textit{uncalibrated} imagery, a scale, $\lambda_k$, and bias, $\xi_k$, term are typically included in Equation \eqref{eq:rad-factor} to account for factors such as distance to the Sun and background noise for each image~\cite{gaskell2008,gaskell2023psj,alexandrov2018multiview}, as discussed in the previous section:
\begin{equation} \label{eq:Ik_angles_uncal}
    I(\sang_{j,k}, \eang_{j,k}, \phi_{j,k}, a_j) = a_j \lambda_k d(\sang_{j,k}, \eang_{j,k}, \phi_{j,k}) + \xi_k. 
\end{equation}
In this case, $a_j$ is the \textit{relative} surface albedo, which refers to the fact that, unless considering radiometrically calibrated imagery~\cite{schroder2013cal,schroder2014cal}, the albedo $a_j$ is only proportional to the absolute albedo as discussed in the previous section.

The Sun-relative direction $\mathbf{s}^\fB_{k} \in \mathbb{S}^2$ in $I_k$, expressed in the body-fixed frame of the small body $\mathcal{B}$, can be estimated using measurements from typical onboard instrumentation (e.g., Sun sensors, star trackers), detailed in Section \ref{sec:sun-sensor}.  
The emitted light vector $\mathbf{e}^\fB_{k,j} = (\vvec{r}^\fB_{\oC_k\oB} - \vvec{\ell}^\fB_j) / \|\vvec{r}^\fB_{\oC_k\oB} - \vvec{\ell}^\fB_j\|$ can be determined from the estimates of $T_{\fB\fC_k}$ and $\vvec{\ell}^\fB_j$ provided by a typical SfM system.  
Finally, dropping the superscripts and letting $T_k$ denote $T_{\fB\fC_k}$ for conciseness, Equation \eqref{eq:Ik_angles} can be written in terms of $\vvec{s}_k$, $\vvec{e}_{k,j}$, and the surface normal $\vvec{n}_j \in \mathbb{S}^2$ at $\vvec{\ell}_j$ (see Fig. \ref{fig:angles}) by noticing that  $\cos\iota_{j,k} = \mathbf{s}_k^\top\mathbf{n}_j$, $\cos\varepsilon_{j,k} = \mathbf{e}_{k,j}^\top\mathbf{n}_j$, and $\phi_{j,k} = \cos^{-1}\left(\mathbf{s}_k^\top\mathbf{e}_{k,j}/\|\mathbf{e}_{k,j}\|\right)$. 
For example, the Lunar-Lambert model (Equation \eqref{eq:ref-llambert}) becomes
\begin{equation} \label{eq:Ik_vecs}
    I(T_k, \mathbf{s}_k, \bm{\ell}_j, \mathbf{n}_j, a_j) = a_j\Lambda(\mathbf{s}_k, \mathbf{e}_{k,j})\left((1 - g(\mathbf{s}_k, \mathbf{e}_{k,j}))\mathbf{s}_k^\top\mathbf{n}_j + g(\mathbf{s}_k, \mathbf{e}_{k,j})\frac{2\mathbf{s}_k^\top\mathbf{n}_j}{\mathbf{s}_k^\top\mathbf{n}_j + \mathbf{e}_{k,j}^\top\mathbf{n}_j}\right).
\end{equation}
We can now define a factor $f_{\mathrm{Ph}}$ corresponding to the presented photoclinometry constraints (assuming zero-mean Gaussian noise) as follows:
\begin{equation} \label{eq:fSPC}
    f_{\mathrm{Ph}}(T_k, \mathbf{s}_k, \vvec{\ell}_j, \vvec{n}_j, a_j; \Sigma_k) \propto \exp\left\{-\frac{1}{2} |I(T_k, \mathbf{s}_k, \vvec{\ell}_j, \vvec{n}_j, a_j) - \Hat{I}_k(\Hat{\vvec{p}}_{j,k})|^2_{\Sigma_k}\right\}.
\end{equation}
This allows for the estimation of $\vvec{n}_j$ and $a_j$ using the measurements $\Hat{I}_k(\Hat{\vvec{p}}_{j,k})$, while also further constraining the landmark's position $\vvec{\ell}_j$, Sun-relative direction $\vvec{s}_k$, and the position of the spacecraft $\vvec{r}_{\oC_k\oB}$. 
The corresponding factor graph diagram is shown in Fig. \ref{fig:factor-graph-ps}. 


\subsection{Sun Vector Measurements} \label{sec:sun-sensor}

Our framework assumes knowledge of the Sun vector $\vvec{s}_k$. 
This direction is usually determined from the target body's ephemeris and onboard attitude estimates (e.g., from a star tracker), which allow this direction to be expressed in the camera frame. 
The Sun vector may also be measured directly by a Sun sensor. 
For example, the OSIRIS-REx spacecraft featured multiple coarse Sun sensors, each with an accuracy of $\pm 1^{\circ}\ (3\sigma)$~\cite{bierhaus2018}. 
Moreover, fine Sun sensors have accuracy on the order of $\pm 0.01^{\circ}\ (3\sigma)$. 

Regardless of the source---either from ephemerides or from a Sun sensor---we assume that the measurements $\Hat{\vvec{s}}^\fC_k  \in \mathbb{S}^2$ are available at each time index $k$ and are expressed in the camera frame $\fC$.
Recalling that $T_k$ denotes $T_{\fB \fC_k}$ and $\vvec{s}_k$ is expressed in the $\fB$ frame, a measurement prediction function $\vvec{s}^C(T_k, \vvec{s}_k)$ can be defined to predict the measured incident light direction $\hat{\vvec{s}}^\fC$ in the $\fC$ frame from the current estimates of $T_k$ and $\vvec{s}_k$:
\begin{equation}
    \mathbf{s}^\mathcal{C}(T_{k}, \mathbf{s}_k) \triangleq R_{\fB\fC_k}^{-1}\mathbf{s}_k.
\end{equation}
We define a factor $f_{SS}$ to incorporate Sun vector measurements into the estimation problem as follows:
\begin{equation} \label{eq:fSS}
    f_{\mathrm{SS}}\left(T_k, \mathbf{s}_k; \Sigma_{j,k}\right) \propto \exp\left\{-\frac{1}{2}\|\mathbf{s}^\mathcal{C}(T_{k}, \mathbf{s}_k) - \Hat{\mathbf{s}}^\mathcal{C}_k\|_{\Sigma_{j,k}}^2\right\}.
\end{equation}
This further constrains the orientation of the camera $R_{\fB\fC_k}$ and the Sun vector $\vvec{s}_k$.


\subsection{Local Smoothness Constraints} \label{sec:smoothness}

Although the photometric minimization and Sun vector terms modeled by $f_\mathrm{Ph}$ and $f_\mathrm{SS}$, respectively, are sufficient to estimate the surface normal and albedo, Horn~\cite{horn1990} indicates that the solution tends to be unstable and gets stuck in local minima, especially if starting far from the solution. 
This has also been demonstrated in other works on small body shape reconstruction~\cite{capanna2013spc,alexandrov2018multiview}.
Thus, Horn proposed the use of local smoothness constraints which minimize the ``departure from smoothness.'' 
Our smoothness constraint \textit{factors} are defined as follows:
\begin{equation} \label{eq:fsmooth}
    f_\mathrm{Smooth}\left(\vvec{\ell}_j, \mathbf{n}_j, \vvec{\ell}_{j'}; \eta\right) \propto \exp\left\{-\frac{1}{2}\eta\left|\cos^{-1}\left(\mathbf{d}_{j',j}^\top\mathbf{n}_j\right) - 90^\circ\right|^2\right\},
\end{equation}
where $\eta$ weights the local smoothness penalty and $\vvec{d}_{j',j} = (\vvec{\ell}_{j'} - \vvec{\ell}_j) / \|\vvec{\ell}_{j'} - \vvec{\ell}_j\|$. 
In words, $f_\mathrm{Smooth}$ encourages landmarks to be locally smooth with respect to the reference landmark's surface normal ($\vvec{n}_j$) by enforcing that $\vvec{d}_{j',j}$, i.e., the vector pointing from the reference landmark ($\vvec{\ell}_j$) towards a neighboring landmark ($\vvec{\ell}_{j'}$), be perpendicular to $\vvec{n}_j$. 
Previous work~\cite{driver2024spc} has demonstrated that adding these smoothness factors results in more feasible surface normal estimates and lower photometric errors.


%% file: figures/factor-graph-sfm.tex
\begin{tikzpicture}[scale=1.0]

    \vertex[label=\color{cadet}\footnotesize$f_{\mathrm{P}}$, draw=cadet, fill=cadet, minimum size=4pt](p) at (-3,0) {};
    
    \vertex[label=center:\small$T_0$, thick, draw=black, minimum size=23pt](T0) at (-2,0) {};
    \vertex[label=center:\small$T_1$, thick, draw=black, minimum size=23pt](T1) at (0,0) {};
    \vertex[label=center:\small$T_{n-1}$, thick, draw=black, minimum size=23pt](Tn1) at (2,0) {};
    \vertex[label=center:\small$T_n$, thick, draw=black, minimum size=23pt](Tn) at (4,0) {};
    
    \vertex[label=center:\small$\bm{\ell}_1$, thick, draw=black, minimum size=23pt](l1) at (-1,-2) {};
    \vertex[label=center:\small$\bm{\ell}_m$, thick, draw=black, minimum size=23pt](lm) at (3,-2) {};
    
    \vertex[label=right:\footnotesize$f_{\mathrm{L}}$,fill=black, minimum size=4pt](m1) at (-1.5,-1) {};
    \vertex[fill=black, minimum size=4pt](m2) at (-.5,-1) {};
    \vertex[fill=black, minimum size=4pt](m3) at (.5,-1) {};
    \vertex[fill=black, minimum size=4pt](m4) at (2.5,-1) {};
    \vertex[fill=black, minimum size=4pt](m5) at (3.5,-1) {};
    
    
    
    % ...
    \vertex[label=center:$\cdots$,draw=none](ee1) at (1,0) {};
    \vertex[label=center:$\cdots$,draw=none](ee2) at (2,-2) {};
    \vertex[draw=none](d2) at (1.2,0) {};
    
    
    \path[draw=cadet, thick, -] (p) edge (T0);
    
    \Edge(T0)(m1)  \Edge(m1)(l1) 
    \Edge(T1)(m2)  \Edge(m2)(l1)
    \Edge(Tn1)(m3) \Edge(m3)(l1)
    \Edge(Tn1)(m4) \Edge(m4)(lm) 
    \Edge(Tn)(m5)  \Edge(m5)(lm)
    

\end{tikzpicture}

%% file: figures/factor-graph-spc.tex
\begin{tikzpicture}[scale=1.0]

    \vertex[label=\color{cadet}\footnotesize$f_{\mathrm{P}}$, draw=cadet, fill=cadet, minimum size=4pt](p) at (-3,0) {};
    
    \vertex[label=center:\small$T_0$, thick, draw=black, minimum size=23pt](T0) at (-2,0) {};
    \vertex[label=center:\small$T_1$, thick, draw=black, minimum size=23pt](T1) at (0,0) {};
    \vertex[label=center:\small$T_{n-1}$, thick, draw=black, minimum size=23pt](Tn1) at (2,0) {};
    \vertex[label=center:\small$T_n$, thick, draw=black, minimum size=23pt](Tn) at (4.2,0) {};
    
    \vertex[label=center:\small$\bm{\ell}_1$, thick, draw=black, minimum size=23pt](l1) at (-1,-2) {};
    \vertex[label=center:\small$\bm{\ell}_m$, thick, draw=black, minimum size=23pt](lm) at (3.5,-2) {};
    
    \vertex[label=center:\small$\mathbf{n}_1$, thick, draw=black, minimum size=23pt](n1) at (.25,-2) {};
    
    \vertex[label=center:\small$a_1$, thick, draw=black, minimum size=23pt](a1) at (1.5,-2) {};
    
    \vertex[label=center:\small$\mathbf{s}_0$, thick, draw=black, minimum size=23pt](s0) at (-2,2) {};
    \vertex[label=center:\small$\mathbf{s}_1$, thick, draw=black, minimum size=23pt](s1) at (0,2) {};
    \vertex[label=center:\small$\mathbf{s}_{q-1}$, thick, draw=black, minimum size=23pt](sk1) at (2,2) {};
    \vertex[label=center:\small$\mathbf{s}_q$, thick, draw=black, minimum size=23pt](sk) at (4.2,2) {};
    
    \vertex[draw = blue, fill=blue, minimum size=4pt](fps0) at (-1,0) {};
    \vertex[draw = blue, fill=blue, minimum size=4pt](fps1) at ( 1,0) {};
    \vertex[label=south east:\color{blue}\footnotesize$f_{\mathrm{Ph}}$, draw = blue, fill=blue, minimum size=4pt](fps2) at ( 3,0) {};
    
    \path[draw=blue, thick, -] (s0) edge (fps0);
    \path[draw=blue, thick, -] (T0) edge (fps0);
    \path[draw=blue, thick, -] (l1) edge (fps0);
    \path[draw=blue, thick, -] (n1) edge (fps0);
    \path[draw=blue, thick, -] (a1) edge (fps0);
    
    \path[draw=blue, thick, -] (s1) edge (fps1);
    \path[draw=blue, thick, -] (T1) edge (fps1);
    \path[draw=blue, thick, -] (l1) edge (fps1);
    \path[draw=blue, thick, -] (n1) edge (fps1);
    \path[draw=blue, thick, -] (a1) edge (fps1);
    
    \path[draw=blue, thick, -] (sk1) edge (fps2);
    \path[draw=blue, thick, -] (Tn1) edge (fps2);
    \path[draw=blue, thick, -] (l1) edge (fps2);
    \path[draw=blue, thick, -] (n1) edge (fps2);
    \path[draw=blue, thick, -] (a1) edge (fps2);
    
    \vertex[label=left:\footnotesize$f_{\mathrm{L}}$,fill=black, minimum size=4pt](m1) at (-1.5,-1) {};
    \vertex[fill=black, minimum size=4pt](m2) at (-.5,-1) {};
    \vertex[fill=black, minimum size=4pt](m3) at (.5,-1) {};
    \vertex[fill=black, minimum size=4pt](m4) at (2.75,-1) {};
    \vertex[fill=black, minimum size=4pt](m5) at (3.85,-1) {};
    
    \vertex[label=left:\color{coquelicot}\footnotesize$f_{\mathrm{SS}}$, draw=coquelicot, fill=coquelicot, minimum size=4pt](fss0)  at (-2,1) {};
    \vertex[draw=coquelicot, fill=coquelicot, minimum size=4pt](fss1)  at ( 0,1) {};
    \vertex[draw=coquelicot, fill=coquelicot, minimum size=4pt](fssk1) at ( 2,1) {};
    \vertex[draw=coquelicot, fill=coquelicot, minimum size=4pt](fssk)  at ( 4.2,1) {};

    \node[rectangle, label=left:\color{darkspringgreen}\footnotesize$f_{\mathrm{Smooth}}$, draw=darkspringgreen, fill=darkspringgreen, minimum size=4pt](fsm0) at (1.5,-3) {};
    
    
    
    % ...
    \vertex[label=center:$\ldots$,draw=none](ee1) at (1,1) {};
    \vertex[label=center:$\ldots$,draw=none](ee2) at (2.5,-2) {};
    \vertex[label=center:$\ldots$,draw=none](ee2) at (2.5,-3) {};
    \vertex[draw=none](d2) at (1.2,0) {};
    
    \path[draw=cadet, thick, -] (p) edge (T0);
    
    \Edge(T0)(m1) \Edge(m1)(l1) 
    \Edge(T1)(m2) \Edge(m2)(l1)
    \Edge(Tn1)(m3) \Edge(m3)(l1)
    \Edge(Tn1)(m4) \Edge(m4)(lm) 
    \Edge(Tn)(m5)  \Edge(m5)(lm)
    
    \path[draw=coquelicot, thick, -] (s0) edge (fss0);
    \path[draw=coquelicot, thick, -] (T0) edge (fss0);
    \path[draw=coquelicot, thick, -] (s1) edge (fss1);
    \path[draw=coquelicot, thick, -] (T1) edge (fss1);
    \path[draw=coquelicot, thick, -] (sk1) edge (fssk1);
    \path[draw=coquelicot, thick, -] (Tn1) edge (fssk1);
    \path[draw=coquelicot, thick, -] (sk) edge (fssk);
    \path[draw=coquelicot, thick, -] (Tn) edge (fssk);

    \path[draw=darkspringgreen, thick, -] (l1) edge (fsm0);
    \path[draw=darkspringgreen, thick, -] (lm) edge (fsm0);
    \path[draw=darkspringgreen, thick, -] (n1) edge (fsm0);

\end{tikzpicture}

%% file: figures/photometry-conventions.tex
\tdplotsetmaincoords{70}{120}
\begin{tikzpicture}[tdplot_main_coords, scale=1.5, every node/.style={scale=1.3}]

    \coordinate (oo) at (0,0,0);
    \coordinate (bl) at (-1/2,-1/2,0);
    \coordinate (br) at (1/2,-1/2,0);
    \coordinate (tl) at (-1/2,1/2,0);
    \coordinate (tr) at (1/2,1/2,0);
    \filldraw[very thick, fill=gray!40] (tl) -- (tr) -- (br) -- (bl) -- cycle;
    \node[label=west:{$\bm{\ell}\,$}] (a) at (0,0,0) {};

    \coordinate (ss) at (3/0.75, 0, 4/0.75);
    \coordinate (ss2) at (3/0.8, 0, 4/0.8);
    
    \coordinate (cc) at (0, 3/0.80, 4/0.80);
    \node[label=north east:{$\vvec{r}_{\oC\oB}$}] () at (cc) {};
    
    \coordinate (cp) at (0, 3/1.05, 4/1.05);
    \coordinate (ctl) at (0.5, 2.5, 4.5);
    \coordinate (ctr) at (0.5, 3.5, 4.5);
    \coordinate (cbl) at (0, 2.5, 3.5);
    \coordinate (cbr) at (0, 3.5, 3.5);
    \draw[very thick,gray] (ctl) -- (ctr);
    \draw[very thick,gray] (ctl) -- (cbl);
    \draw[very thick,gray] (cbl) -- (cbr);
    \draw[very thick,gray] (ctr) -- (cbr);
    
    \filldraw[very thick, fill=gray!20] (ctl) -- (ctr) -- (cbr) -- (cbl) -- cycle;
    \draw[very thick] (cc) -- (ctl);
    \draw[very thick] (cc) -- (ctr);
    \draw[very thick] (cc) -- (cbr);
    \draw[very thick] (cc) -- (cbl);


    \shadedraw[opacity = .5] (3/0.75, 0, 4/0.75) circle (.2cm);
    \fill[black] (oo) circle (0.10);
    \filldraw[draw=none, fill=gray!90] (cp) circle (.05cm);
    \filldraw[thick, fill=black] (cc) circle (.05cm);
    
    \node[inner sep=0pt] (camera) at (0, 3/1.1, 4/1.1) {};
    \node[inner sep=0pt] (tmp) at (cp) {$\,\quad\vvec{p}$};
    

    \draw[dashed] (ss2) -- (oo);
    \draw[dashed] (cp) -- (oo);
    \draw[very thick,->] (oo) -- (0,0,5/1.7) coordinate (nhat) node [black,right] {$\,\mathbf{n}$};
    \draw[very thick,->] (oo) -- (3/1.5,0,4/1.5) coordinate (ihat) node [black,left] {$\mathbf{s}\,$};
    \draw[very thick,->] (oo) -- (0,3/1.7,4/1.7) coordinate (ehat) node [black,right] {$\,\mathbf{e}$};
    
    \tdplotsetrotatedcoords{0}{-90}{0}
    \tdplotdrawarc[tdplot_rotated_coords]{(0,0,0)}{1.8}{0}{90-53.13}{anchor=south}{$\quad\varepsilon$}
    \tdplotsetrotatedcoords{90}{-90}{0}
    \tdplotdrawarc[tdplot_rotated_coords]{(0,0,0)}{1.8}{0}{-36.87}{anchor=south}{$\iota\quad$}
    \tdplotsetrotatedcoords{45}{-62.0616}{0}
    \tdplotdrawarc[tdplot_rotated_coords]{(0,0,0)}{1.8}{24}{-24}{anchor=north}{$\phi$}
\end{tikzpicture}

%% file: figures/rad-factors.tex
\begin{tabular}{p{3cm}ccccc}
        \toprule
        & $d(\sang, \eang, \phi)$ & $g(\phi)$ & $\Lambda(\phi)$ \\
        \midrule
        Akimov~\cite{shkuratov1994principle,shkuratov2011optical}
        & $\cos\left(\frac{\phi}{2}\right) \cos\left[\frac{\pi}{\pi - \phi}\left(\gamma - \frac{\phi}{2}\right)\right] \frac{(\cos\eta)^{\phi / (\pi - \phi)}}{\cos\gamma}$ & --- & --- & \parbox{1cm}{\begin{equation}\label{eq:ref-akimov}\end{equation}} \\
        \midrule
        \rowcolor[gray]{0.9}
        McEwen~\cite{mcewen1991,mcewen1996precise}
        & $(1 - g(\phi))\cos\sang + g(\phi)\frac{2\cos\sang}{\cos\sang + \cos\eang}$ & $\exp(-\phi/60)$ & --- & \parbox{1cm}{\begin{equation}\label{eq:ref-mcewen}\end{equation}} \\
        \midrule
        Akimov+
        & $\cos\left(\frac{\phi}{2}\right) \cos\left[\frac{\pi}{\pi - \phi}\left(\gamma - \frac{\phi}{2}\right)\right] \frac{(\cos\eta)^{g(\phi) \phi / (\pi - \phi)}}{\cos\gamma}$ & $w_0 + w_1\phi$ & $\sum_{i=0}^m c_i\phi^i$ & \parbox{1cm}{\begin{equation}\label{eq:ref-akimov5}\end{equation}} \\
        \midrule
        \rowcolor[gray]{0.9}
        Lunar-Lambert
        & $(1 - g(\phi))\cos\sang + g(\phi)\frac{2\cos\sang}{\cos\sang + \cos\eang}$ & $w_0 + w_1\phi$ & $\sum_{i=0}^m c_i\phi^i$ & \parbox{1cm}{\begin{equation}\label{eq:ref-llambert}\end{equation}} \\
        \midrule
        Minnaert~\cite{minnaert1941reciprocity}
        & $(\cos(\sang))^{g(\phi)}(\cos(\eang))^{g(\phi) - 1}$ & $w_0 + w_1\phi$ & $\sum_{i=0}^m c_i\phi^i$ & \parbox{1cm}{\begin{equation}\label{eq:ref-minnaert}\end{equation}} \\
        \bottomrule
    \end{tabular}

%% file: figures/rad-factors-coeffs.tex
\begin{tabular}{p{1mm}lcccccccc}
    \toprule
    & Model & $w_0$ & $w_1$ && $c_1$ & $c_2$ & $c_3$ & $c_4$ \\
    \midrule
    \rowcolor[gray]{0.9}
    & Akimov+       & $1.57 $ & $-9.88\times 10^{-3}$ && $-1.9219\times 10^{-2}$ & $2.2193\times 10^{-4}$ & $-1.6245\times 10^{-6}$ & $4.6468\times 10^{-9}$ \\
    \rowcolor[gray]{0.9}
    & Lunar-Lambert & $0.830$ & $-7.22\times 10^{-3}$ && $-1.7160\times 10^{-2}$ & $1.8306\times 10^{-4}$ & $-1.0399\times 10^{-6}$ & $2.3223\times 10^{-9}$ \\
    \rowcolor[gray]{0.9}
    \multirow{-3}{*}{\parbox[t]{2mm}{\rotatebox[origin=c]{90}{Vesta}}}
    & Minnaert      & $0.554$ & $ 4.35\times 10^{-3}$ && $-1.6910\times 10^{-2}$ & $1.7807\times 10^{-4}$ & $-9.7674\times 10^{-7}$ & $2.1063\times 10^{-9}$ \\
    \midrule
    \multirow{3}{*}{\parbox[t]{2mm}{\rotatebox[origin=c]{90}{Ceres}}}
    & Akimov+       & $1.109$ & $-2.85\times 10^{-3}$ && $-2.2435\times 10^{-2}$ & $2.1477\times 10^{-4}$ & $-7.5103\times 10^{-7}$ & --- \\
    & Lunar-Lambert & $0.896$ & $-8.87\times 10^{-3}$ && $-2.2118\times 10^{-2}$ & $2.0912\times 10^{-4}$ & $-6.4209\times 10^{-7}$ & --- \\
    & Minnaert      & $0.514$ & $ 5.09\times 10^{-3}$ && $-2.2568\times 10^{-2}$ & $2.2297\times 10^{-4}$ & $-7.3108\times 10^{-7}$ & --- \\
    \bottomrule
\end{tabular}

%% file: text/experiments.tex
\begin{figure}
    \centering
    \begin{subfigure}[T]{.32\linewidth}
        \includegraphics[width=\linewidth]{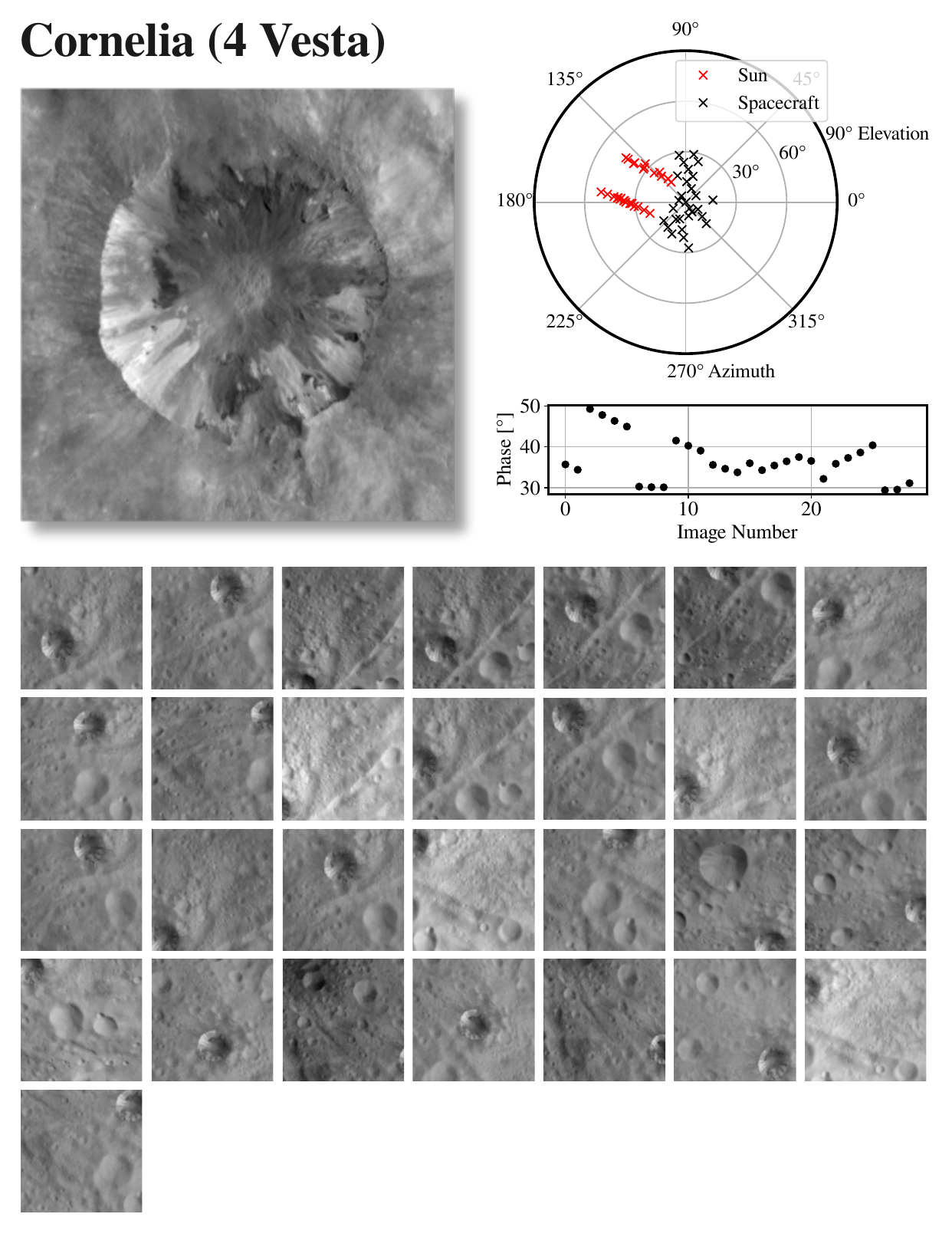}
    \end{subfigure}%
    \hfill
    \begin{subfigure}[T]{.32\linewidth}
        \includegraphics[width=\linewidth]{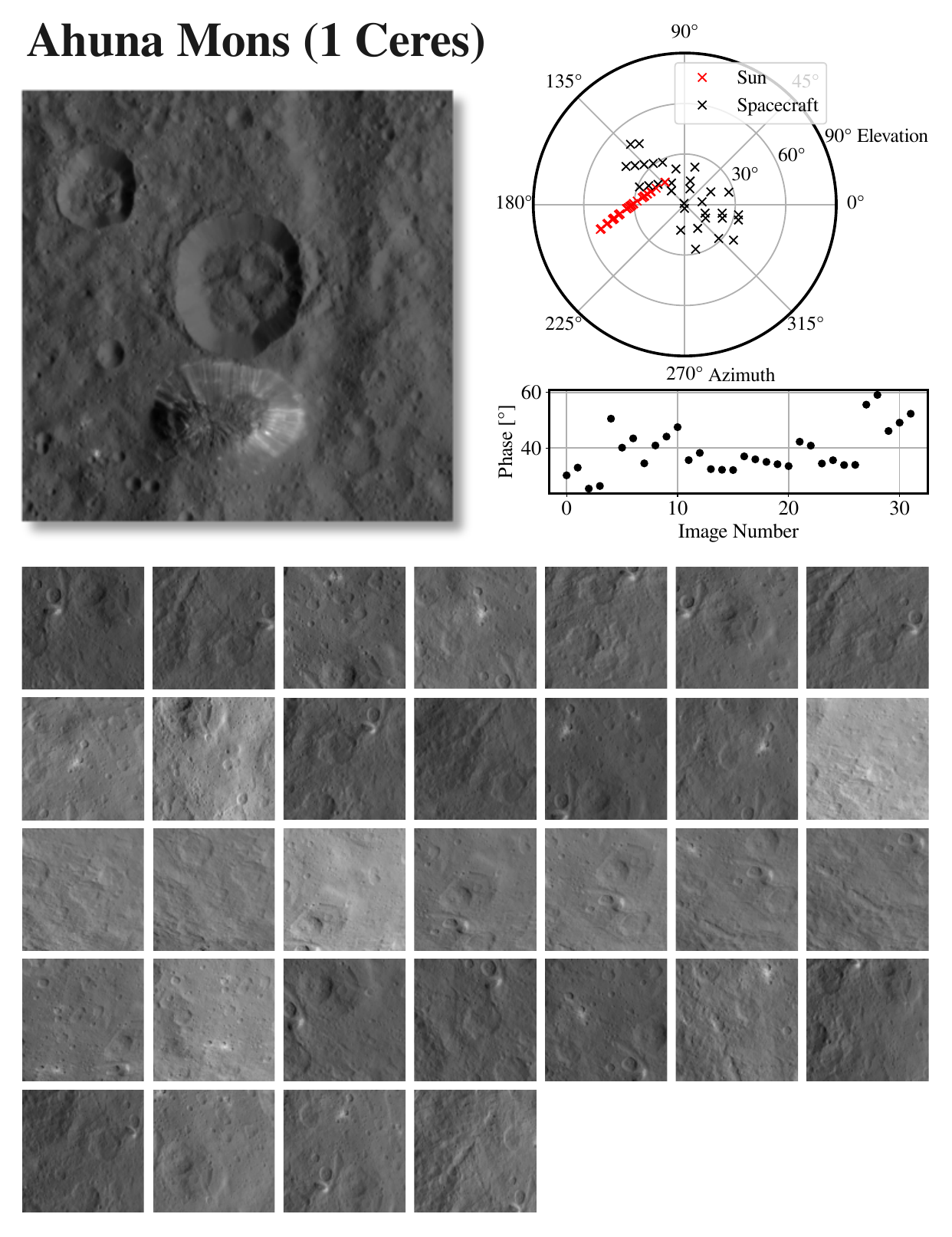}
    \end{subfigure}%
    \hfill
    \begin{subfigure}[T]{.32\linewidth}
        \includegraphics[width=\linewidth]{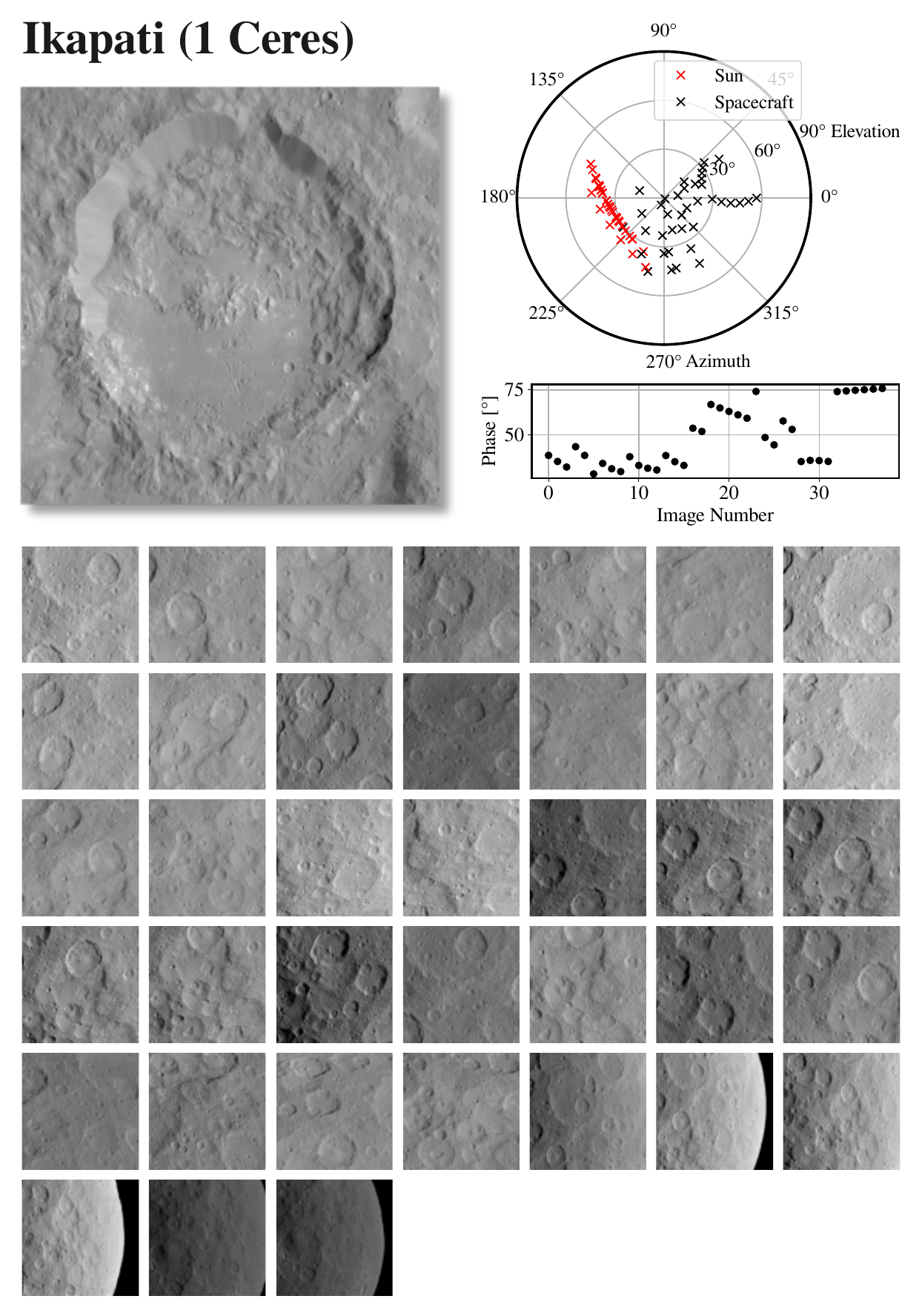}
    \end{subfigure}%
    \caption{Images and baselines for each experiment. \normalfont{The Sun and spacecraft azimuths and elevations are relative to the first image.}}
    \label{fig:image-sequences}
\end{figure}

Our experiments leverage images of Asteroid 4 Vesta and Minor Planet 1 Ceres captured by NASA's Dawn mission~\cite{russell2012dawn,sierks2011dawn} to evaluate the proposed approach. 
Ceres is the largest object in the asteroid belt, while Vesta is the second-largest object and the brightest asteroid visible from Earth.
The images used in this investigation are publicly available on NASA's Planetary Data System (PDS)~\cite{pds} and maintained by NASA's Navigation and Ancillary Information Facility (NAIF).
Each image has a resolution of $1024 \times 1024$ pixels. 
We will focus on one site on 4 Vesta, Cornelia, and two sites on 1 Ceres, Ahuna Mons and Ikapati. 
The full image sequences used in this investigation for each site are given in Figure~\ref{fig:image-sequences}. 

\paragraph{Cornelia (4 Vesta)} 
Cornelia (centered at 15.6$^\circ$E and 9.4$^\circ$S) is an approximately 15-km diameter crater on the surface of Vesta that has been the subject of numerous studies~\cite{denevi2012pitted,williams2014lobate,schroder2013resolved}.  
This site was chosen in part because of its interesting albedo distribution with both bright and dark features, which represents a challenging scenario for photometric reconstruction methods such as SPC and the proposed approach. 
The 29 images used in our reconstruction of Cornelia were captured during the High Altitude Mapping Orbit (HAMO) with a ground sample distance (GSD) of $\sim$60 meters. 

\paragraph{Ahuna Mons (1 Ceres)} 

Ahuna Mons (centered at 315.8$^\circ$E and 10.5$^\circ$S) is the tallest mountain on the surface of Ceres with an average height of approximately 4 km (13,000 ft) from its base, believed to have formed due to cryovolcanic activity. 
The bright streaks on the side of the dome are attributed to salt and water ice deposits from ancient cryovolcanic eruptions, which offset the generally darker surface of Ceres~\cite{russell2016ceres,ruesch2016cryovolcanism}. 
Ahuna Mons is also accompanied by a large 17-km diameter impact crater just northwest of its base, offering a wide range of challenging topography to validate the proposed approach. 
The 32 images used in our reconstruction of Ahuna Mons were captured during the HAMO with a GSD of $\sim$140 meters. 

\paragraph{Ikapati (1 Ceres)} 

Ikapati (centered at 45.9$^\circ$E and 33.4$^\circ$N) is an approximately 50-km diameter impact crater on the surface of Ceres that includes bright deposits believed to be salt-rich material exposed by the impact~\cite{krohn2016cryogenic}. 
The crater interior features challenging ridges and pitted terrain. 
The 38 images used in our reconstruction of Ikapati were captured during the HAMO with a GSD of $\sim$140 meters and images captured during the Extended Mission Orbit 7 (XMO7) with a GSD of $\sim$300 meters. 


\subsection{Experimental Baselines} \label{sec:baselines}

We compared our pipeline against three different baselines generated using three different approaches: SPC, SPG, and dense SfM.

\paragraph{Stereophotoclinometry (SPC)} 
A key comparison point involves the traditional SPC pipeline~\cite{gaskell2008, gaskell2023psj}, which serves as a baseline to evaluate our surface normal and albedo estimates.
The SPC reconstruction of Cornelia utilizes the Lunar-Lambert reflectance model (Equation \eqref{eq:ref-llambert}), with parameters detailed in Table \ref{tab:sch-coeff}. 
A regional bigmap was generated by merging 289 overlapping sub-maps, or maplets, each with a spatial resolution of 30 meters. 
These maplets were constructed from images acquired during both the HAMO and LAMO mission phases. 
The \textit{a priori} topography and pose information for the maplets were derived from previously converged maps at a 50-meter spatial resolution, themselves based on earlier 100-meter resolution maps. 
Additional information on the SPC reconstruction process for Vesta is available in \cite{mastrodemos2012}.

For the SPC reconstructions of Ahuna Mons and Ikapati, a variant of the McEwen reflectance model (Equation \eqref{eq:ref-mcewen}) with a constant phase weighting function was employed. 
These reconstructions achieved a spatial resolution of 100 meters, utilizing images captured during both the HAMO and LAMO mission phases. 
The \textit{a priori} topography and pose information was sourced from a previously converged global shape model of Ceres, derived from imagery collected during the Approach and Survey phases. More details on the SPC reconstruction process for Ceres can be found in \cite{park2019spc}.

\paragraph{Stereophotogrammetry (SPG)} 

SPG reconstructions for each site developed by the German Aerospace Center (DLR) provide another point of comparison. 
Each site was reconstructed using the SPG pipeline described in \cite{raymond2011dawn}. 
The Cornelia map was sampled from a global SPG DTM of 4 Vesta with a spatial resolution of $\sim$70 meters reconstructed using HAMO imagery~\cite{vesta_dtm}.
The Ahuna Mons and Ikapati maps, generated using LAMO imagery, have a spatial resolution of $\sim$30 meters~\cite{ceres_dtm}.

\paragraph{Dense Structure-from-Motion (SfM)} 
This baseline corresponds to the proposed PhoMo pipeline without the $f_\mathrm{Ph}$, $f_\mathrm{SS}$, and $f_\mathrm{Smooth}$ factors. 
This baseline is used to investigate the effect of the photoclinometry constraints on the estimated topography. 

\paragraph{} 
Since our method does not assume any priors on the camera poses or landmark positions, resulting in a scale ambiguity, we must align our solution to the baselines before comparison. 
To do this, we estimated a $\mathrm{Sim}(3)$ transformation between the estimated and ground truth camera poses using Karcher mean and \cite{Zinsser2005icprip} (implemented in GTSAM's \texttt{Similarity3.align} function), followed by iterative closest point alignment between the PhoMo and baselines maps. 


\subsection{Implementation Details} \label{sec:implementation-details}

Keypoint measurements and matches were computed using RoMa~\cite{edstedt2024roma}, which provides dense, per-pixel correspondences. 
Since these ``detector-free'' methods do not provide a discrete set of keypoints per image, and instead compute a dense mapping between the pixel coordinates of each image pair, we choose a reference image and match all images with respect to the pixel cetners of this image. 
The matching is constrained to a $400 \times 400$ pixel region centered around each site in the reference image, as this approximately contains the extent of the baseline SPC and SPG maps. 
Thus, each of the maps estimated by our approach contains 160,000 points. 
The keypoint measurements are assigned a covariance of $\Sigma_{j,k} = \mathrm{I}_2$.
Image brightness values are (bilinearly) interpolated at the keypoint measurements $\Hat{\vvec{p}}_{j,k}$ to derive the measurements $\Hat{I}_k(\Hat{\vvec{p}}_{j,k})$ used in the proposed Photoclinometry factors $f_\mathrm{Ph}$ defined in Equation \eqref{eq:fSPC}. 
The image brightness measurements are assigned a standard deviation of $\sigma_I = 0.5$ for the uncalibrated case and $\sigma_I = 0.01$ for the calibrated case, which we found to work well empirically. 
Only landmarks with $\geq 6$ keypoint measurements are inserted into the graph. 

The (simulated) Sun sensor measurements $\hat{\vvec{s}}^\fC_k \in \mathbb{S}^2$ were derived from the normalized ground truth values from SPC of the Sun's relative position to the origin of the camera frame $\oC_k$ expressed in the camera frame $\fC_k$, i.e., $\vvec{r}_{\oI\oC_k}^{\fC_k}$, and assigned an uncertainty of $\Sigma_\xi =  \sigma_\xi^2\mathrm{I}_2$ where $\sigma_\xi = 1 \times 10^{-3}$. 
Next, the surface normals were initialized by finding the $32$ closest neighbors to each point in the point cloud and fitting a plane to this local terrain, and the normal to the plane is taken as the initial surface normal. 
These initial surface normals were then used to initialize the albedo by independently computing the albedo in each image using the initial camera poses and landmark positions, where the initial albedo of each landmark was taken to be the average albedo computed over all views from which it was seen.
The smoothness factors $f_\mathrm{Smooth}$ (Equation \eqref{eq:fsmooth}) were inserted into the graph between each landmark and its four closest neighbors. 
We found a very small value for the local smoothness weight to work well for our experiment, where we used a value of $\eta = 10^{-4}$.
We leverage the GTSAM library~\cite{dellaert2012} to model the proposed keypoint-based SPC problem using factor graphs and optimize the resulting nonlinear least-squares using the Levenberg-Marquardt algorithm and the analytical partial derivatives of the measurement functions for the respective factors (see Appendix \hyperref[sec:partials]{A}). 


\subsection{Performance metrics} \label{sec:metrics}

We define the following metrics to measure the performance of the proposed approach:
\begin{equation}
    \delta \ell_j \triangleq \| \bm{\ell}_j - \overline{\bm{\ell}}_j\|_2,
\end{equation}
\begin{equation}
    \delta\epsilon_j \triangleq \cos^{-1}\left(\mathbf{n}_j^\top\overline{\mathbf{n}}_j\right),
\end{equation}
and
\begin{equation}
    \delta a_j \triangleq |a_j - \overline{a}_j| / \overline{a}_j.
\end{equation}
As before, $\overline{\bm{\ell}}$, $\overline{\bm{n}}$, $\overline{a}$ denote the ground truth values of the landmark position, surface normal, and albedo, respectively. 
These ground truth values are assigned by finding the closest point in our reconstructed map to that of the baseline (after the alignment step), and taking the position of that landmark, as well as the associated normal and albedo for the SPC baseline, as the ground truth. 
Next, the root mean squared error between the measured, $\Hat{I}_k(\Hat{\vvec{p}}_{j,k})$, and estimated, $I(T_k, \mathbf{s}_k, \vvec{\ell}_j, \vvec{n}_j, a_j)$, image brightness values, normalized by the average measured brightness, is used as a photometric error metric for each landmark, as in \cite{schroder2013resolved}:
\begin{equation} \label{eq:photo-error}
    \delta I_j \triangleq \left(\frac{1}{|\mathcal{K}_j|}\sum_{k\in \mathcal{K}_j} \Hat{I}_k(\Hat{\vvec{p}}_{j,k})\right)^{-1}\sqrt{\frac{1}{|\mathcal{K}_j|}\sum_{k\in \mathcal{K}_j}\left(I(T_k, \mathbf{s}_k, \vvec{\ell}_j, \vvec{n}_j, a_j) - \Hat{I}_k(\Hat{\vvec{p}}_{j,k})\right)^2},
\end{equation}
where $\mathcal{K}_j$ denotes the set of indices of the images from which the $j$th landmark was viewed.
Finally, we evaluated the rendering performance of PhoMo and other approaches using the peak-signal-to-noise ratio (PSNR). 
For images normalized to the range $[0, 1]$, the PSNR is defined as
\begin{equation} \label{eq:psnr}
    \mathrm{PSNR} = 10\log_{10}\left(\frac{1}{\mathrm{MSE}}\right),
\end{equation}
where $\mathrm{MSE} = \frac{1}{H\cdot W}\sum_{i=0}^H\sum_{j=0}^W I_{i, j} - \Hat{I}_{i, j}$ is the mean squared error between the actual $I \in [0, 1]^{H\times W}$ and rendered $\hat{I} \in [0, 1]^{H\times W}$ and $I_{i,j}$ represents the image value in the $i$th row and $j$th column.

%% file: text/results.tex

\subsection{Rendering Comparisons between PhoMo and SPC}

\begin{table}[tb!]
    \centering
    \ra{1.5}
    \caption{PSNR comparison (higher is better) between PhoMo and SPC. \normalfont{Values in each column are color-coded using a linear gradient from 
    \cbg{gray!00}{ }\cbg{gray!04}{w}\cbg{gray!08}{o}\cbg{gray!012}{r}\cbg{gray!16}{s}\cbg{gray!20}{t}\cbg{gray!23}{ }\cbg{gray!27}{t}\cbg{gray!31}{o}\cbg{gray!34}{ }\cbg{gray!38}{b}\cbg{gray!42}{e}\cbg{gray!45}{s}\cbg{gray!48}{t}\cbg{gray!50}{ }.
    }}
    \begin{adjustbox}{width=0.8\linewidth}
    \input{figures/psnr-results-spc}
    \end{adjustbox}
    \label{tab:psnr-results-spc}
\end{table}

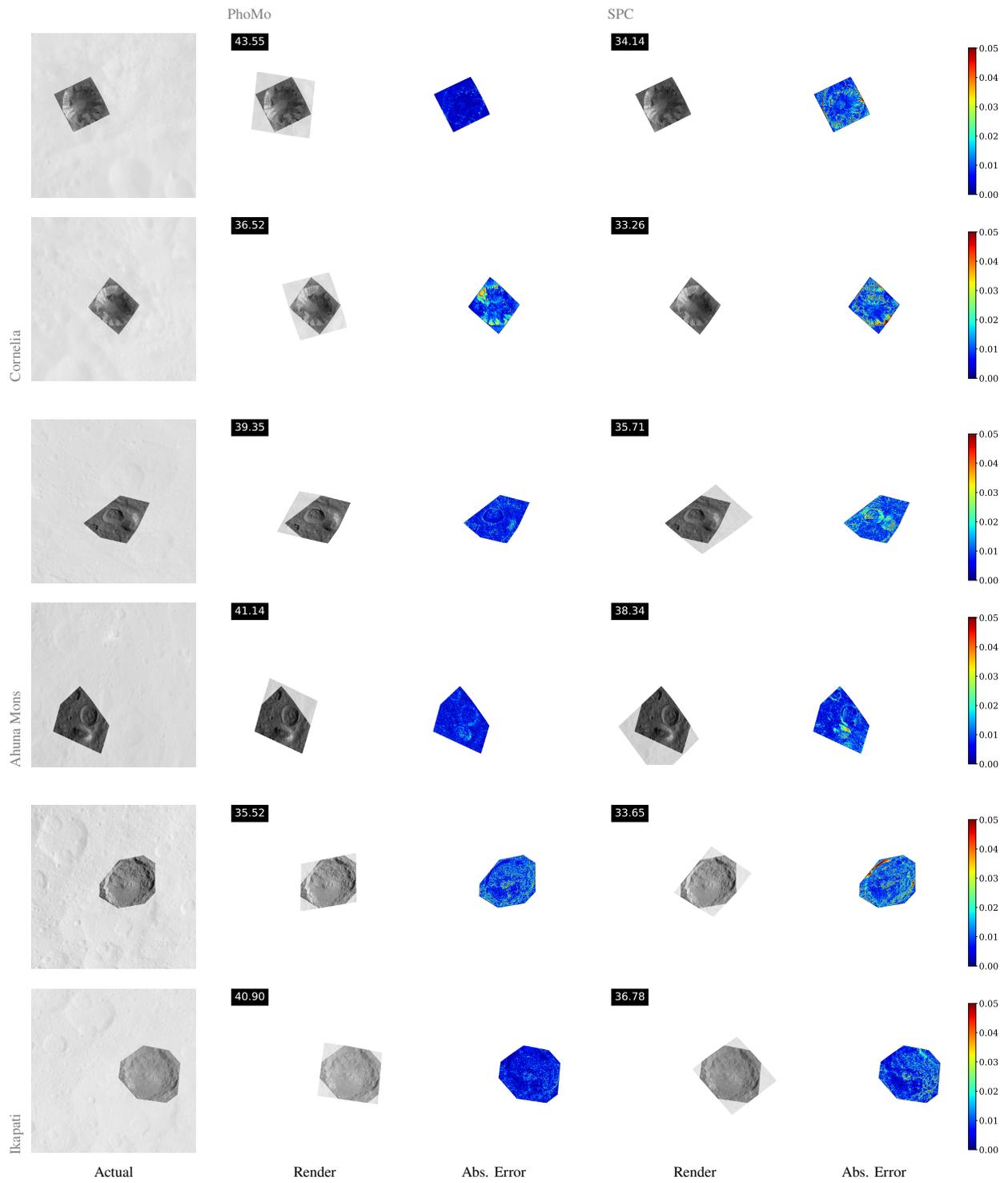
\begin{figure}[p]
    \centering
    \input{figures/qual-compare-phomo-spc-cornelia}
    \caption{PhoMo and SPC renderings. \normalfont{The PSNR for each rendering is highlighted in the top-left corner of the image. The PhoMo renderings use the L-Lambert model.}}
    \label{fig:qual-compare-phomo-spc-cornelia}
\end{figure}

The outputs of both PhoMo and SPC were used to generate renderings of the imaged scene. 
Specifically, the estimated landmark positions, albedos, and surface normals from PhoMo and SPC were combined with the estimated camera position and Sun vector for a particular image within a given reflectance model to generate brightness values for each landmark as discussed in Section \ref{sec:photoclinometry}. 
We then projected these landmarks with their associated brightness values into each image using the estimated camera relative pose and (bilinearly) interpolated the image brightness at the pixel centers contained within the convex hull of the 2D coordinates of the projected landmarks. 
The resulting renderings were then compared with the actual images to assess the quality of the reconstructions. 

We compared the renderings from PhoMo for each of the five reflectance models and the renderings from SPC with the actual images of each site using the PSNR (Equation \eqref{eq:psnr}) in Table \ref{tab:psnr-results-spc}. 
We also provide example renderings for each site in Fig. \ref{fig:qual-compare-phomo-spc-cornelia}. 
On average, all reflectance models used in PhoMo achieve a PSNR exceeding 38, while SPC achieved a lower, but still impressive, average PSNR of 33.49.
Among the investigated reflectance models, the McEwen and Lunar-Lambert models performed the best on average, albeit marginally, while the Akimov+ model performed slightly worse. 
Indeed, the range between the lowest and highest PSNR for each site is 0.60, 0.32, and 0.26 for Cornelia, Ahuna Mons, and Ikapati, respectively, indicating very little difference in rendering quality between the best and worst performing reflectance models for each site. 
The uncalibrated models (Akimov and McEwen) marginally outperform their calibrated counterparts (Akimov+ and Lunar-Lambert), likely due to the additional degrees of freedom provided by the scale and bias factors. 
We compare the values of the per-image scale factor estimated by the uncalibrated models to the explicit phase function leveraged by the calibrated models in Appendix \ref{sec:reflect-analysis}. 
Nevertheless, both uncalibrated and calibrated reflectance models demonstrated high rendering quality. 


\subsection{Reconstruction Comparisons between PhoMo, SPC, SPG, and SfM} \label{sec:spc-spg-sfm-comparison}

\begin{table}[tb!]
    \centering
    \ra{1.5}
    \caption{Comparison between the PhoMo reconstructions and each experimental baseline. \normalfont{Values in each column are color-coded using a linear gradient from 
    \cbg{gray!00}{ }\cbg{gray!04}{l}\cbg{gray!08}{e}\cbg{gray!012}{a}\cbg{gray!16}{s}\cbg{gray!20}{t}\cbg{gray!23}{ }\cbg{gray!27}{t}\cbg{gray!31}{o}\cbg{gray!34}{ }\cbg{gray!38}{gr}\cbg{gray!42}{ea}\cbg{gray!45}{te}\cbg{gray!48}{st}\cbg{gray!50}{ }.
    }}
    \begin{adjustbox}{width=0.6\linewidth}
    \input{figures/spc-comparison}
    \end{adjustbox}
    \label{tab:spc-comparison}
\end{table}

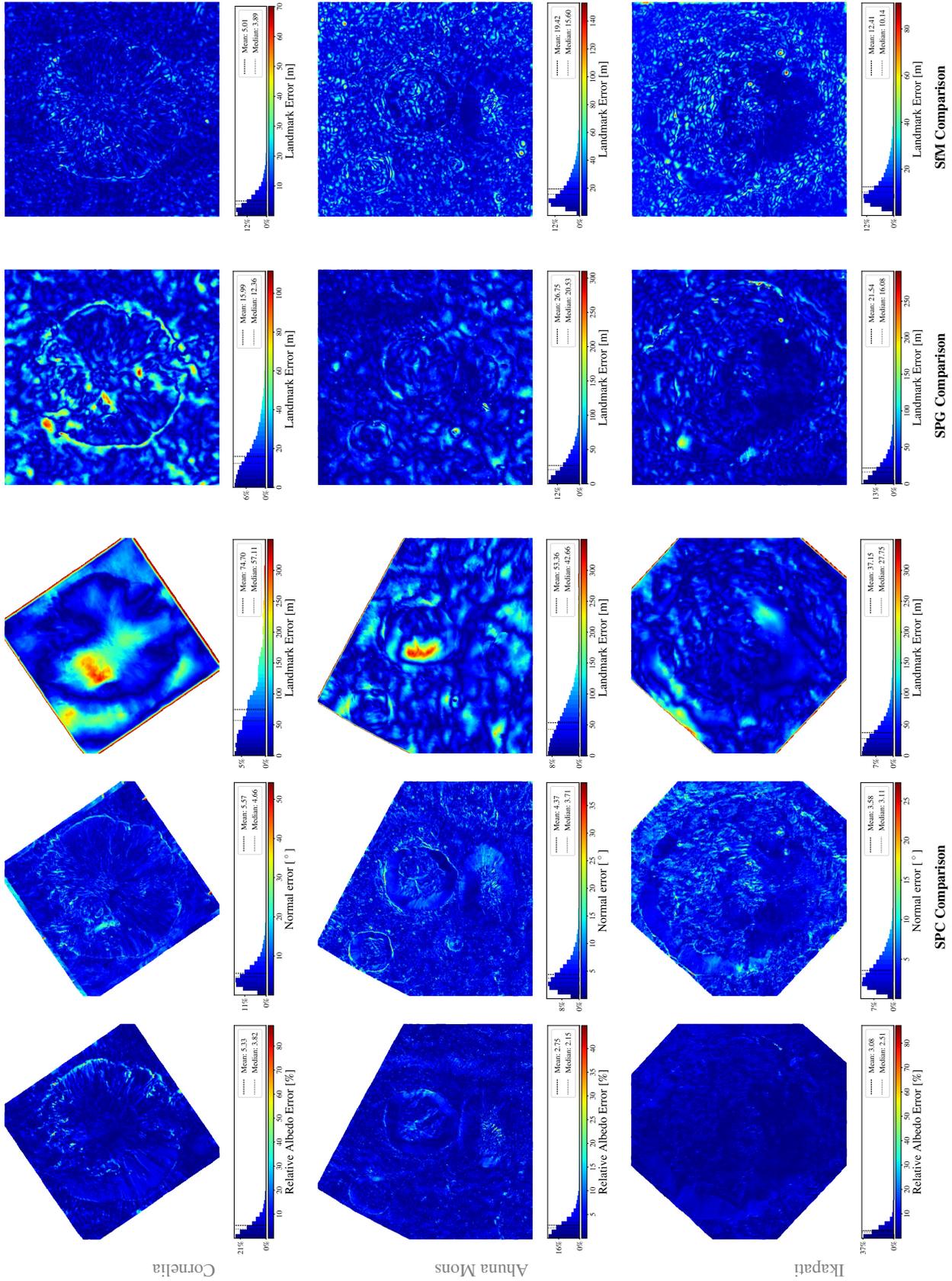
\begin{sidewaysfigure}[p]
\centering
\input{figures/spc-spg-sfm-comparison}
\caption{Comparison between the PhoMo reconstructions and each experimental baseline.}
\label{fig:spc-spg-sfm-comparison}
\end{sidewaysfigure}

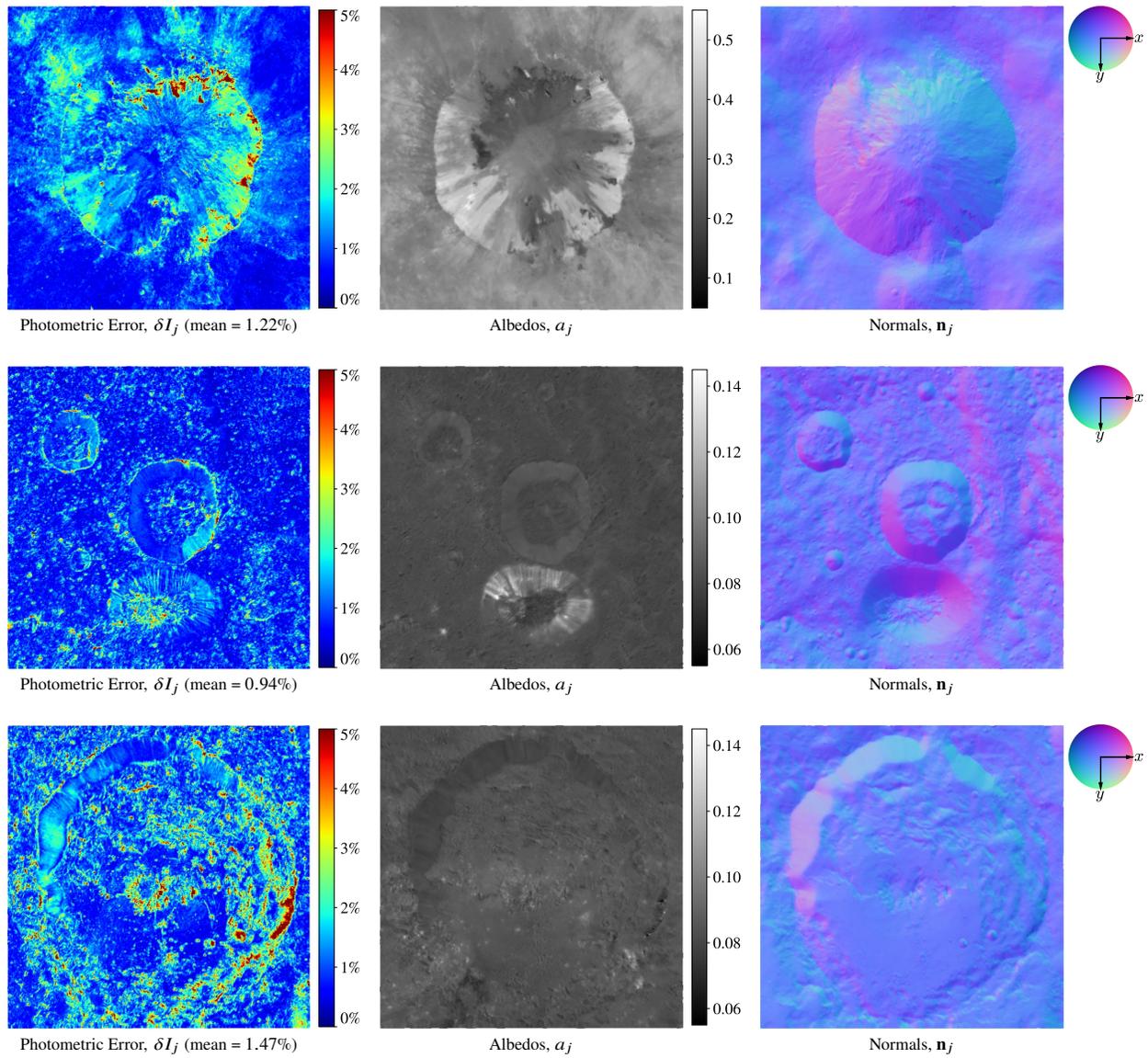
\begin{figure}[p]
    \centering
    \input{figures/phomo-albedos-normals}
    \caption{PhoMo results with the L-Lambert model.}
    \label{fig:phomo-albedos-normals}
\end{figure}

We compared our solution to the SPC, SPG, and SfM baselines, and provide the resulting photometric error metric (Equation \eqref{eq:photo-error}), for each reflectance model and site in Table \ref{tab:spc-comparison}. 
Fig.~\ref{fig:spc-spg-sfm-comparison} visualizes the resulting albedo, normal, and landmark error maps for the Lunar-Lambert model, while Fig.~\ref{fig:phomo-albedos-normals} illustrates the albedos and surface normals estimated by PhoMo with the Lunar-Lambert model, along with the corresponding photometric error map. 
For the Cornelia site, all models yield a photometric error of approximately 1.2\%, with the uncalibrated cases (Akimov and McEwen) achieving slightly lower errors. 
The albedos and surface normals align closely with the SPC baseline, with albedo errors under 6\% and normal errors below $6^\circ$.
At the Ahuna Mons site, all models produce photometric errors below 1.0\%, with the McEwen model achieving the lowest value of 0.78\%. 
Albedo and surface normal errors also show good alignment with SPC, with all models achieving albedo errors under 3\%, except for McEwen at 3.42\%, and normal errors under $5^\circ$.
For the Ikapati site, photometric errors remain below 1.5\%, with the uncalibrated models (Akimov and McEwen) again showing slightly lower errors. 
Albedo and surface normal errors are also consistent with SPC, with albedo errors below 3\% for most models, except the Minnaert model at 3.74\%, and normal errors under $4^\circ$.
Overall, for Ahuna Mons and Ikapati, the calibrated models generally achieve lower surface normal and landmark errors but exhibit slightly higher photometric and albedo errors compared to the uncalibrated models. 

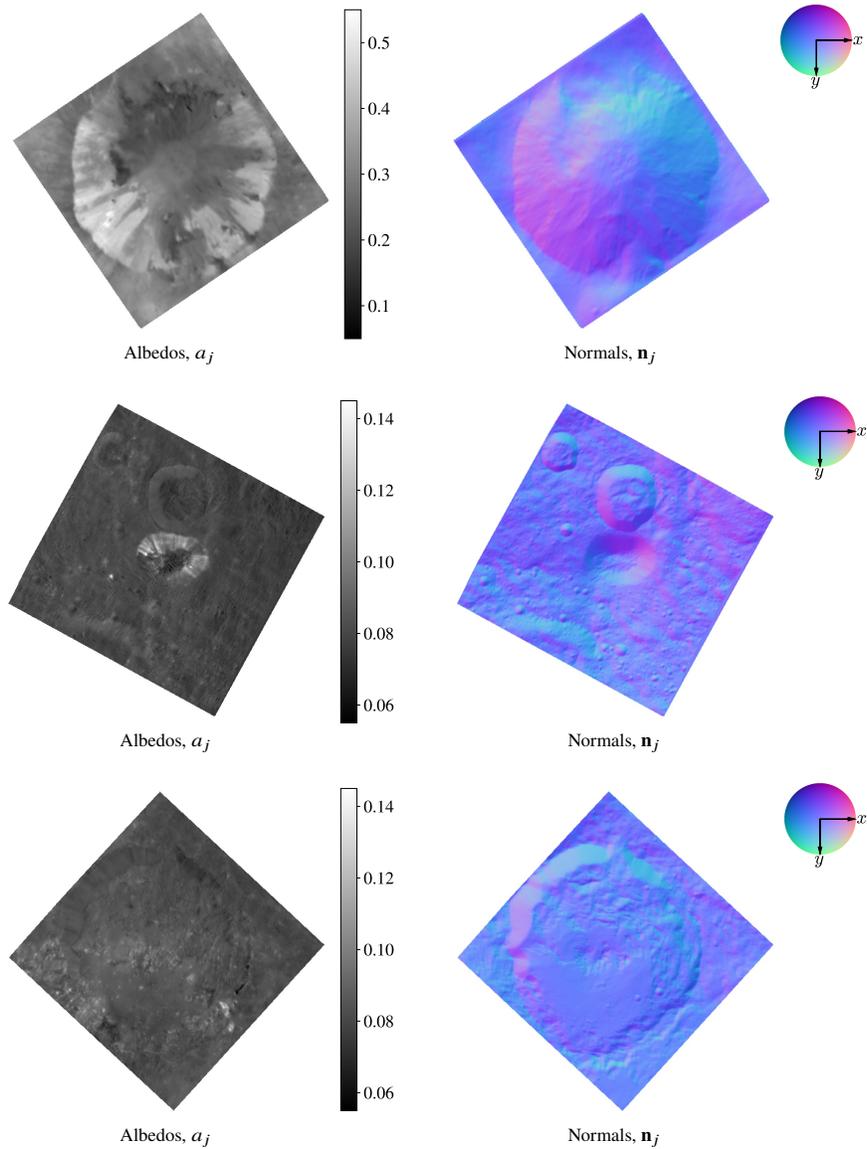
\begin{figure}[p]
    \centering
    \input{figures/spc-albedos-normals}
    \caption{SPC reconstructed albedos and normals. \normalfont{Albedos are scaled according to the PhoMo reconstruction with the L-Lambert model.}}
    \label{fig:spc-albedos-normals}
\end{figure}

All PhoMo solutions exhibit regions with relatively large landmark errors when compared to SPC. 
However, these regions of large landmark errors are not observed when compared to SfM and SPG solutions. 
Specifically, the landmark error, $\delta\ell$, is typically about twice as large compared to SPC as it is to SPG, reaching nearly four times as large for the Cornelia dataset. 
This indicates that the landmark errors likely stem from inaccuracies in the SPC solution rather than from errors in our approach, likely due to associated surface normal errors.
Indeed, since the landmark heights in the SPC solution are computed by integrating the slopes~\cite{mastrodemos2012}, errors in the slope translate to errors in the landmark positions that propagate from the points in the map where the slope errors arise towards the direction of the integration. 
A qualitative comparison of our reconstructed surface normals (Fig. \ref{fig:phomo-albedos-normals}) with those from SPC (Fig. \ref{fig:spc-albedos-normals}) reveals that the SPC normal map appears smoothed relative to the PhoMo normal map, especially for the Cornelia dataset, where the largest errors occur.
This phenomenon of surface slope smoothing by SPC, especially for areas with larger slope gradients such as craters, has also been observed in numerous other works~\cite{barnouin2020, daly2022shape}. 
Conversely, since our landmarks are not explicitly derived from the surface normal estimates, and instead are independently estimated and constrained by the keypoint measurements, our pipeline is not as susceptible to surface normal errors as the traditional SPC pipeline. 

\begin{figure}[p]
    \centering
    \includegraphics[width=\linewidth]{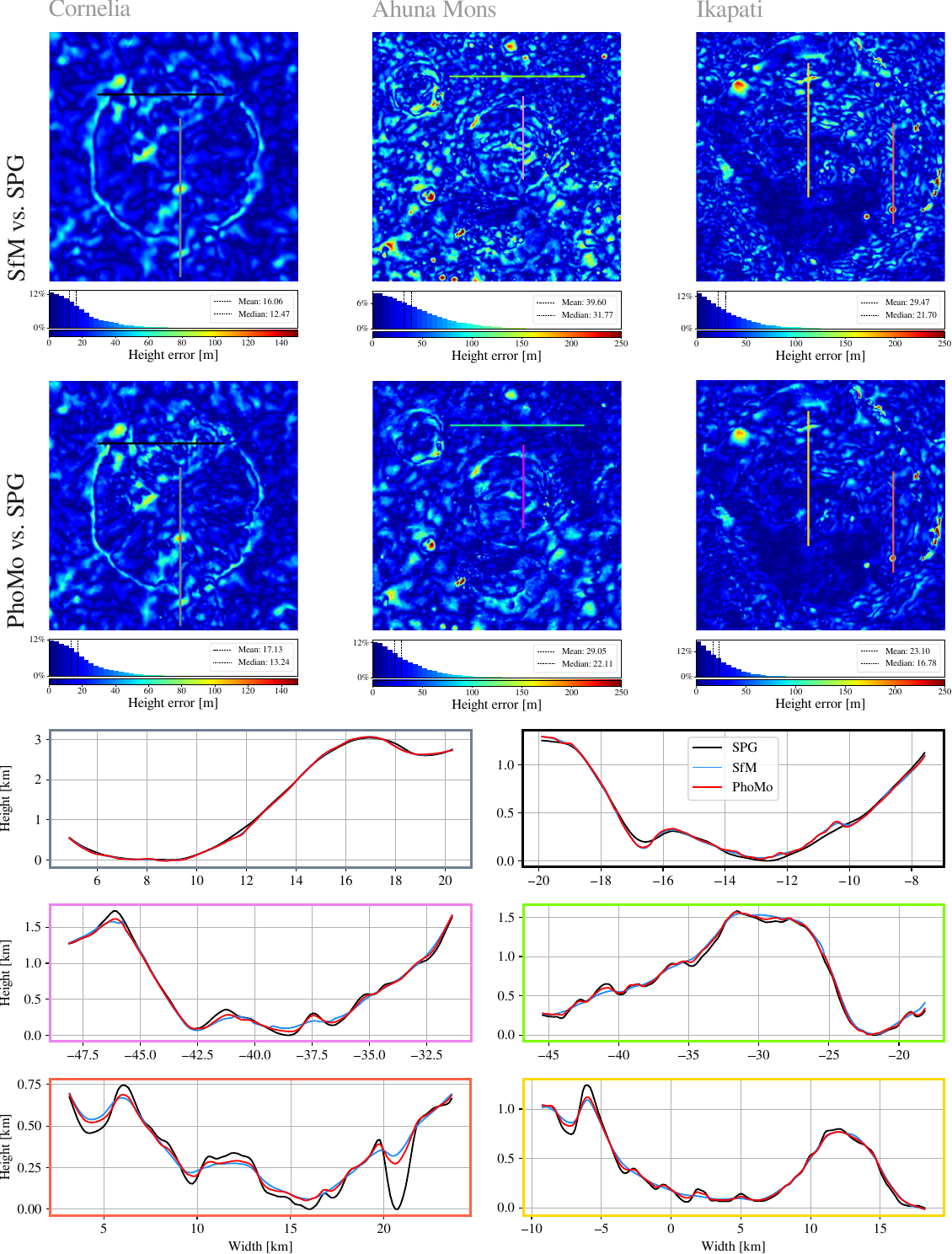}
    \caption{Height errors and line scan plots for PhoMo and SfM as compared to SPG.}
    \label{fig:height-err-phomo-sfm}
\end{figure}

Finally, we compared the estimated height maps of PhoMo and SfM with the SPG baseline in Fig. \ref{fig:height-err-phomo-sfm}. 
The height maps were derived by aligning the dense SfM and PhoMo landmark maps to the SPG reconstruction, computing the $z$-depth of each landmark in the reference image's camera frame, and defining the height as the distance in the -$z$ direction from the maximum depth. 
Recall that the dense SfM solution was obtained using the PhoMo pipeline without incorporating photoclinometry constraints. 
Thus, these results illustrate the impact of adding photoclinometry constraints to the reconstructed topography.
It can be seen from the results that PhoMo exhibits significantly lower average height error on the Ahuna Mons and Ikapati experiments compared to SfM. 
PhoMo reduces the average height error from 39.60 meters to 29.05 meters on Ahuna Mons, a 26.64\% reduction, and from 29.47 meters to 23.10 meters on Ikapati, a 21.31\% reduction. 

As previously discussed, the SPG maps for Ahuna Mons and Ikapati were generated using LAMO imagery with a ground sample distance (GSD) of approximately 30 meters/pixel. 
This higher resolution allows the SPG maps to resolve fine surface features that may not be distinguishable using the lower-resolution HAMO imagery (140 meters/pixel) used in the PhoMo and SfM reconstructions. 
Despite this limitation, the inclusion of photoclinometry constraints in the PhoMo pipeline appears to enable the resolution of some of these higher resolution features, such as small ridges and craters, which SfM alone cannot achieve, resulting in better alignment with the SPG reconstruction. 
This phenomenon is further visualized in the line-scan plots at the bottom of Fig. \ref{fig:height-err-phomo-sfm}.
However, this reduction in height error is not observed when comparing against the Cornelia SPG reconstruction, as it was constructed from HAMO images with the same spatial resolution as the PhoMo map ($\sim$70 meters/pixel). Consequently, the added detail provided by the photoclinometry constraints in the PhoMo map cannot be assessed in this case.


\subsection{Rendering Comparisons between PhoMo, NeRFs, and Gaussian Splatting}

\begin{table}[tb!]
    \centering
    \ra{1.5}
    \caption{PSNR comparison (higher is better) between PhoMo, instant-NGP (iNGP)~\cite{mueller2022ingp}, and 3D Gaussian Splatting (3DGS)~\cite{kerbl20233dgs}. \normalfont{The PhoMo result uses the L-Lambert reflectance model. The \textbf{first} and \underline{second} highest PSNRs are bolded and highlighted, respectively.}}
    \begin{adjustbox}{width=0.6\linewidth}
    \input{figures/psnr-results-nerf-3dgs}
    \end{adjustbox}
    \label{tab:psnr-results-nerf}
\end{table}

\begin{figure}[tb!]
    \centering
    \begin{adjustbox}{width=\linewidth}
    \input{figures/qual-compare-phomo-nerf-3dgs}
    \end{adjustbox}
    \caption{PhoMo, instant-NGP, and SplatFacto renderings for tain (top) and test (bottom) images. \normalfont{The PSNR for each rendering is highlighted in the top-left corner of the image. The PhoMo renderings use the L-Lambert model.}}
    \label{fig:qual-compare-phomo-nerf-3dgs}
\end{figure}
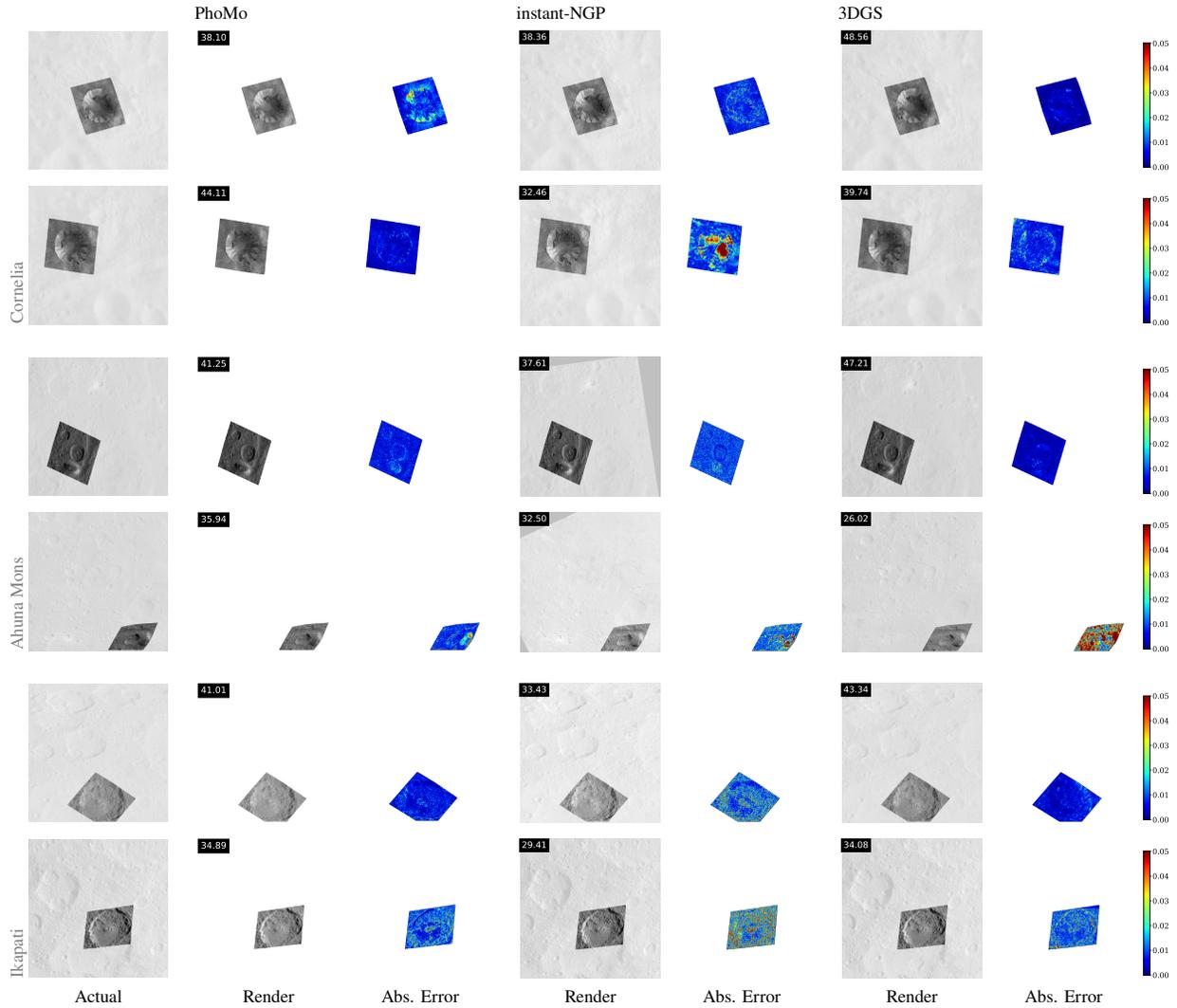

Finally, we compared our PhoMo renderings to those generated by methods based on implicit scene representations, namely, Neural Radiance Fields (NeRFs) and 3D Gaussian Splatting (3DGS) (discussed in Section \ref{sec:related-dl}). 
We chose instant-NGP~\cite{mueller2022ingp} as a representative NeRF method, as in \cite{givens2024nerf}, and the original 3DGS method~\cite{kerbl20233dgs}. 
For 3DGS, we also included optimization of a $3\times 4$ per-image affine correction term by including half of each test image in the training set as proposed in \cite{kerbl2024h3dgs}, where the half was chosen such that it did not include the portion reconstructed by PhoMo. 
For this experiment, we leveraged the estimated poses from SPC for each method, where the poses were fixed for the PhoMo method by setting strong priors for each pose. 
3DGS additionally requires an initial point cloud, where we used the dense initial point cloud from PhoMo as it was found to work better than the initial sparse point cloud from GTSfM. 
The images for each site were split according to a 90/10 train/test split. 

The rendering performance with respect to the PSNR for each method on each site is summarized in Table \ref{tab:psnr-results-nerf}, with example renderings provided in Fig. \ref{fig:qual-compare-phomo-nerf-3dgs}. 
PhoMo consistently demonstrates superior rendering quality on average for both the training and testing sets compared to instant-NGP. 
Importantly, there is very little difference between the rendering quality, measured by the PSNR, between the train and test images for PhoMo, where the difference between the average train and test PSNR is less than 0.5 for Ikapati and Cornelia and just over 3 for Ahuna Mons. 
Compared to 3DGS, PhoMo achieves a noticeably lower PSNR on the training images, but delivers a higher PSNR on the test images, which suggests that the explicit representation employed by PhoMo allowed for a superior ability to generalize to novel views.

\begin{figure}
    \centering
    \begin{subfigure}[T]{.49\linewidth}
        \includegraphics[width=\linewidth]{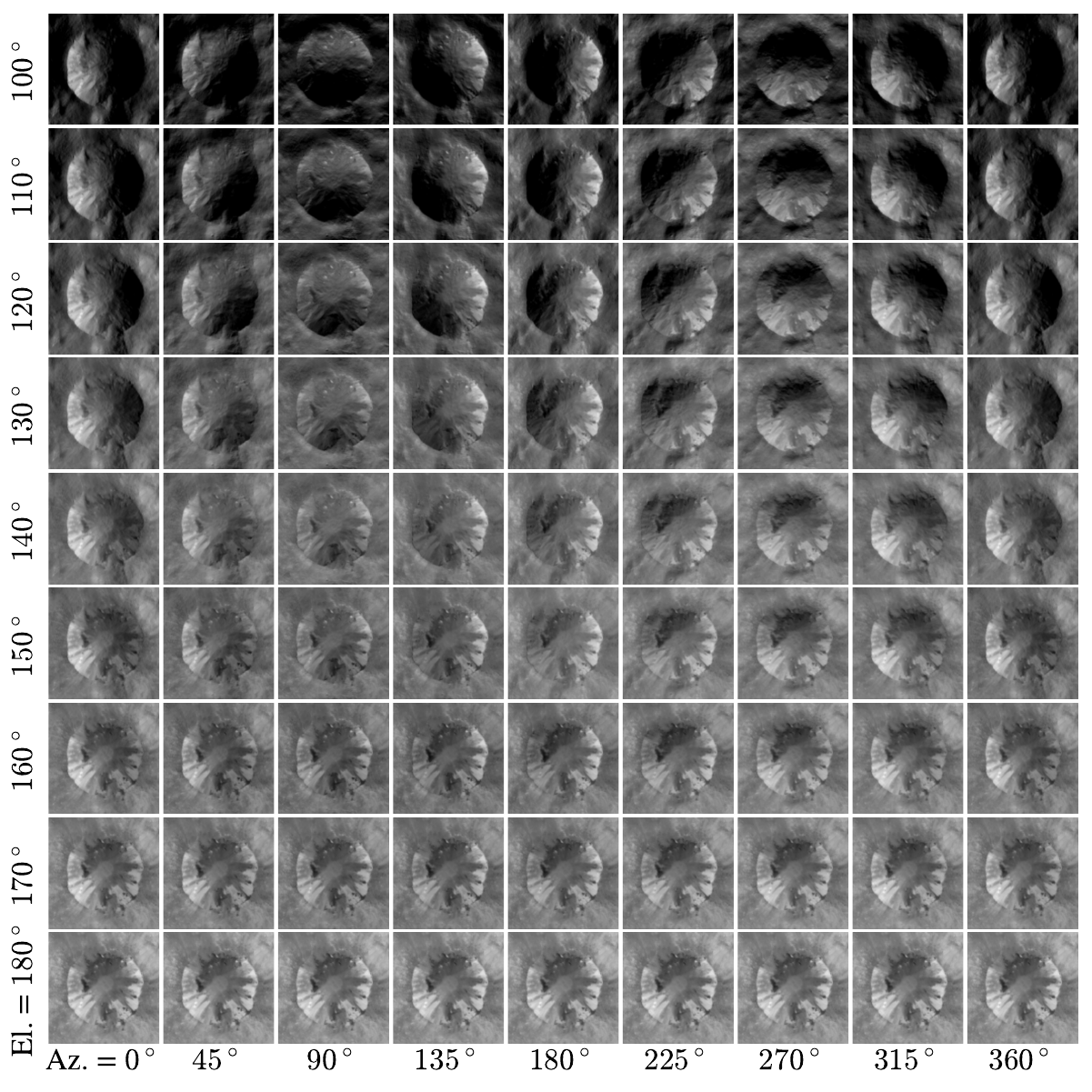}
        \vspace{-20pt}
        \caption*{Cornelia}
    \end{subfigure}%
    \hfill
    \begin{subfigure}[T]{.49\linewidth}
        \includegraphics[width=\linewidth]{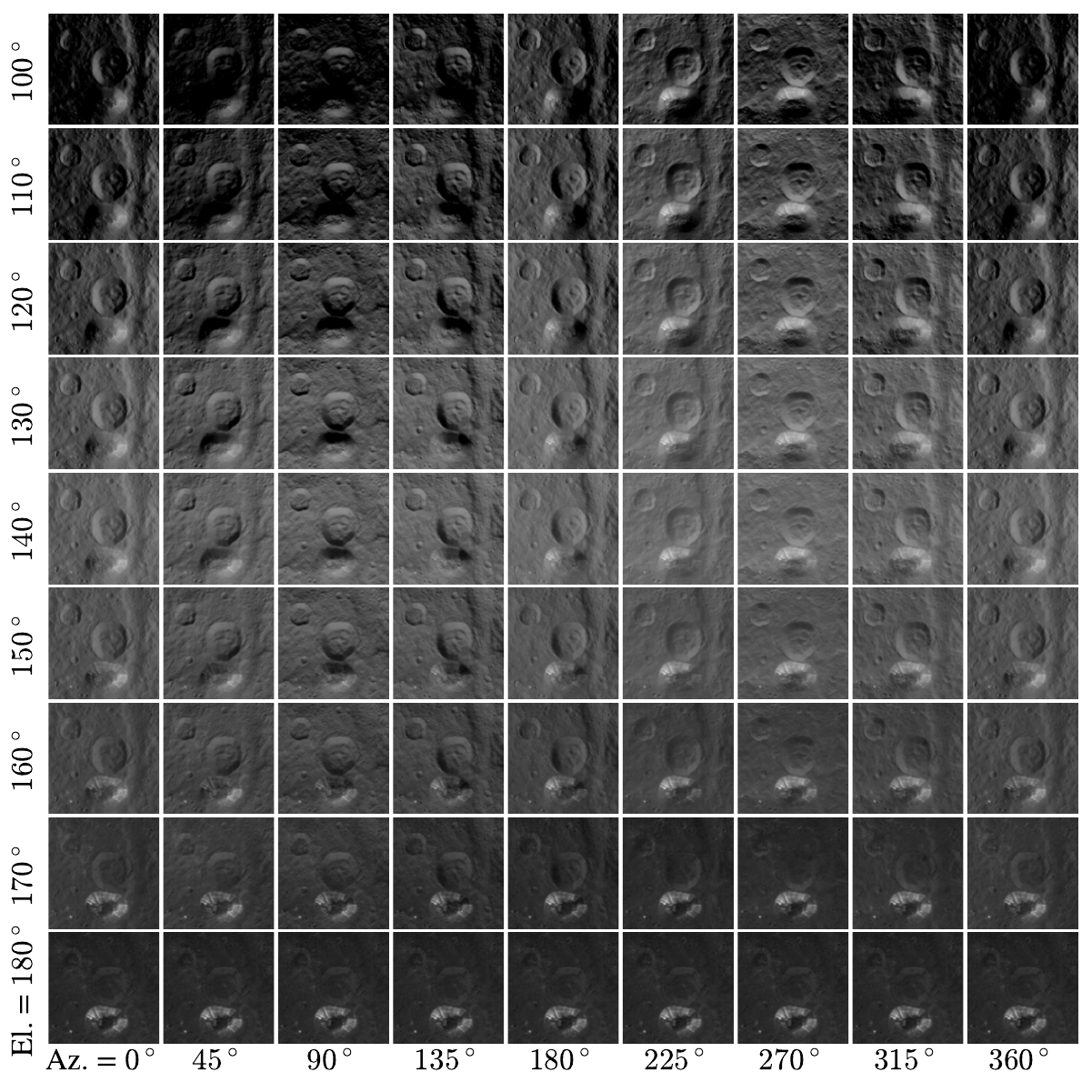}
        \vspace{-20pt}
        \caption*{Ahuna Mons}
    \end{subfigure}%
    \hfill
    \begin{subfigure}[T]{.49\linewidth}
        \includegraphics[width=\linewidth]{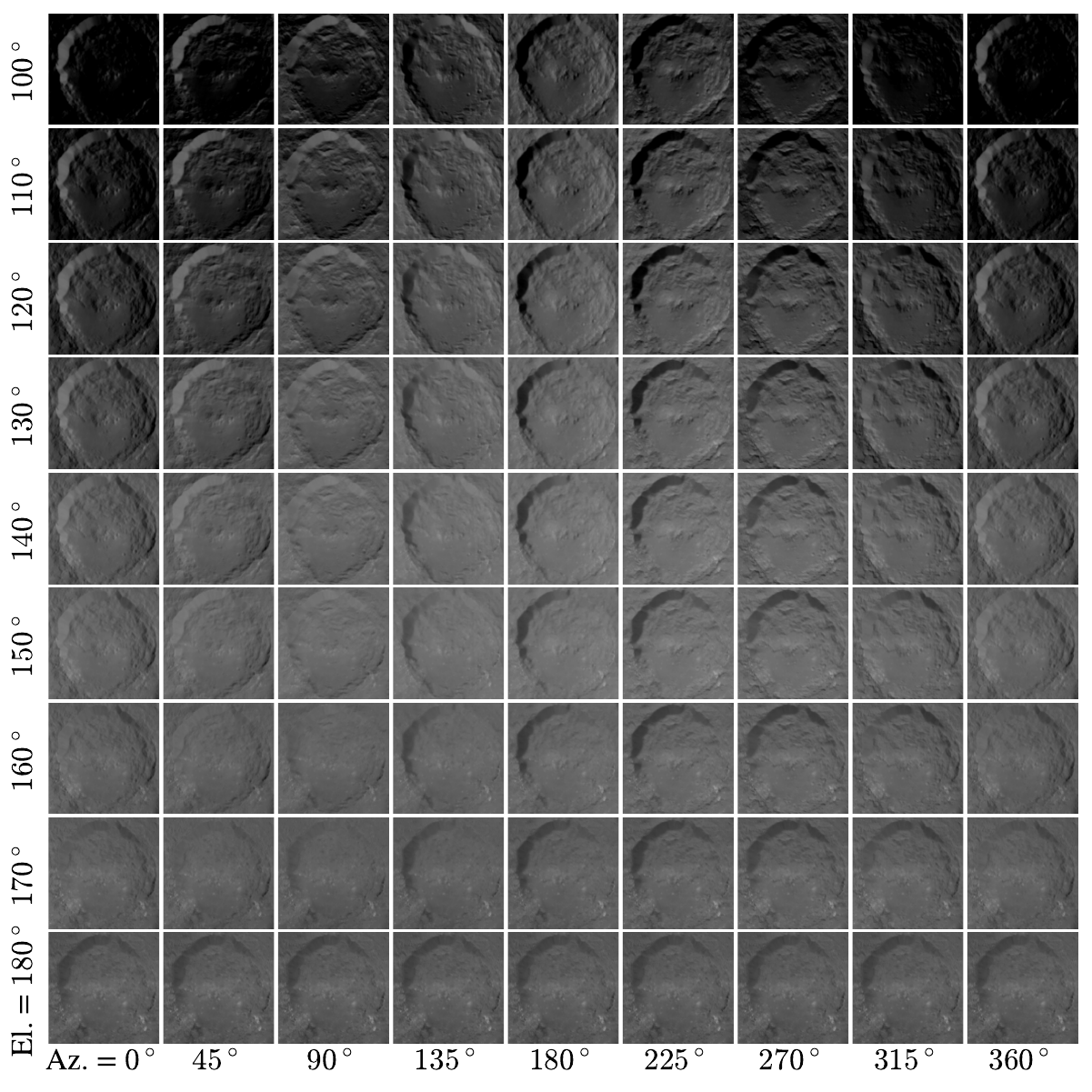}
        \vspace{-20pt}
        \caption*{Ikapati}
    \end{subfigure}%
    \caption{PhoMo rendering at various illumination azimuths and elevations.}
    \label{fig:illumination-az-el}
\end{figure}

Although 3DGS is capable of generating accurate renderings, the estimated topography, as represented by the centers of the computed 3D Gaussians, features significant amounts of noise. 
This is discussed in Appendix \ref{sec:landmark-err-3dgs}. 
Moreover, since PhoMo explicitly represents the photometric properties of the scene using physical quantities, namely the Sun vector, the PhoMo reconstruction can be used to render each site under specified illumination conditions as illustrated in Fig. \ref{fig:illumination-az-el}. 
The results also demonstrate that the surface reflectance properties of small bodies can be reliably represented using relatively simple models with few tunable parameters. 
Indeed, PhoMo represents the photometric properties of each landmark with just 3 parameters, 2 for the surface normal and 1 for the albedo, combined with a \textit{global} reflectance model, while 3DGS estimates 55 parameters for each landmark (3D Gaussian center), including 7 parameters for the scale and orientation of the 3D Gaussian and 44 spherical harmonic parameters to represent the directional appearance component of the radiance field. 
Incorporating Sun vector information and the semi-empirical reflectance models used in this work into the 3DGS framework would be an interesting direction for future research.

%% file: figures/psnr-results-spc.tex
\begin{tabular}{lrrrrr>{\columncolor{blue!10}}r}\toprule
 & \multicolumn{5}{c}{PhoMo (Ours)} \\
\cmidrule{2-6}
             & Akimov & McEwen & Akimov+ & L-Lambert & Minnaert & SPC \\ 
\midrule
Cornelia     & \cc{gray!45} 40.36 & \cc{gray!50} 40.42 & \cc{gray!00} 39.82 & \cc{gray!28} 40.16 & \cc{gray!38} 40.28 & 33.09 \\
Ahuna Mons   & \cc{gray!00} 38.69 & \cc{gray!47} 38.99 & \cc{gray!22} 38.83 & \cc{gray!50} 39.01 & \cc{gray!44} 38.97 & 36.41 \\
Ikapati      & \cc{gray!40} 37.21 & \cc{gray!50} 37.26 & \cc{gray!33} 37.17 & \cc{gray!46} 37.24 & \cc{gray!00} 37.00 & 33.49 \\
\midrule
Average      & \cc{gray!25} 38.75 & \cc{gray!50} 38.89 & \cc{gray!00} 38.61 & \cc{gray!34} 38.80 & \cc{gray!25} 38.75 & 33.49 \\
\bottomrule
\end{tabular}

%% file: figures/qual-compare-phomo-spc-cornelia.tex
\centering
\setlength{\extrarowheight}{-10pt}
\setlength{\tabcolsep}{1pt} 
\begin{tabular}{cp{2.9cm}p{.3cm}p{2.9cm}p{2.9cm}p{.3cm}p{2.9cm}p{2.9cm}p{.6cm}}
    &&& \multicolumn{2}{l}{\textcolor{gray}{\scriptsize{PhoMo}}} && \multicolumn{2}{l}{\textcolor{gray}{\scriptsize{SPC}}} \\
    &
    \includegraphics[width=\linewidth]{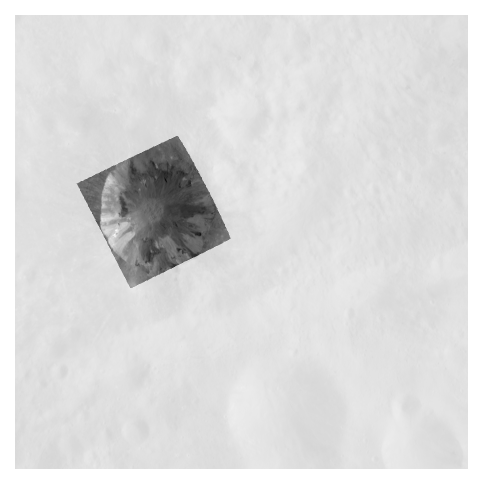} &&
    \includegraphics[width=\linewidth]{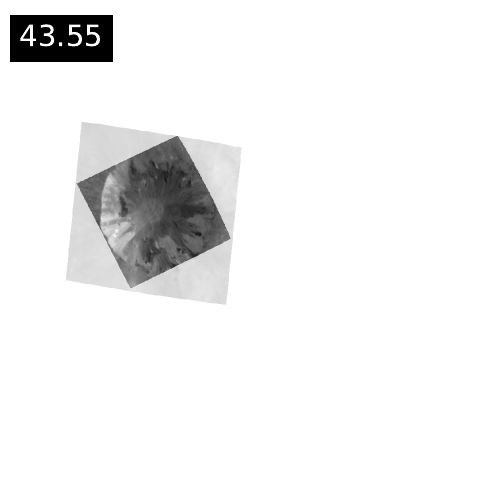} &
    \includegraphics[width=\linewidth]{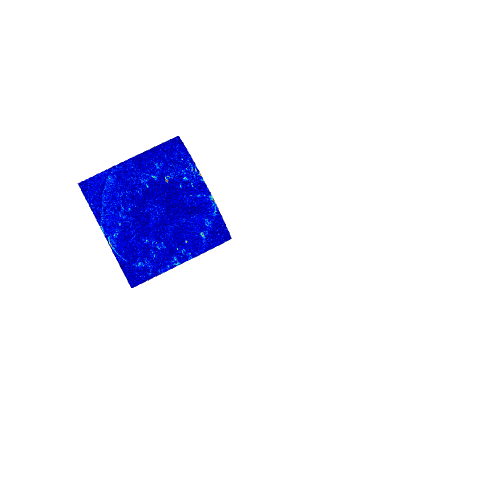} &&
    \includegraphics[width=\linewidth]{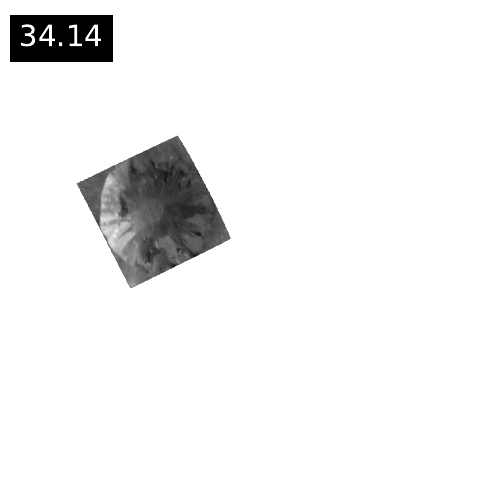} &
    \includegraphics[width=\linewidth]{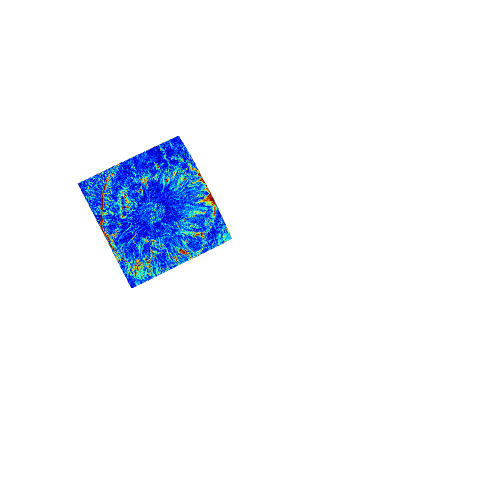} &
    \includegraphics[width=\linewidth]{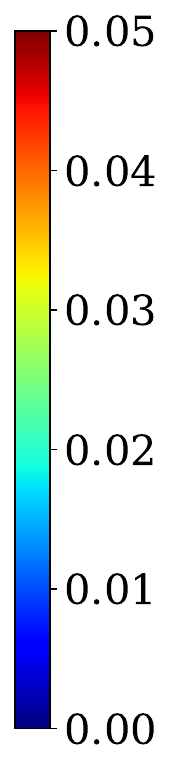} \\
    \parbox[t]{2mm}{\rotatebox[origin=l]{90}{\textcolor{gray}{\scriptsize{Cornelia}}}} &
    \includegraphics[width=\linewidth]{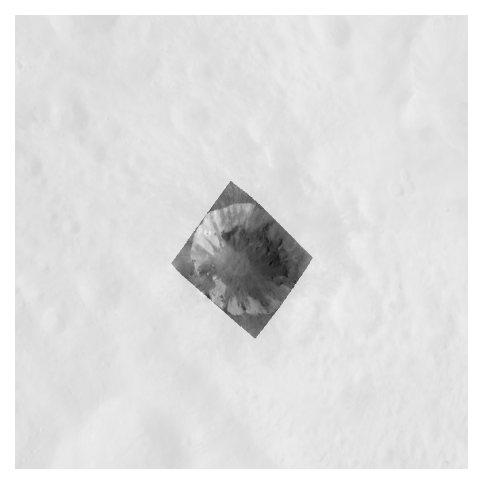} &&
    \includegraphics[width=\linewidth]{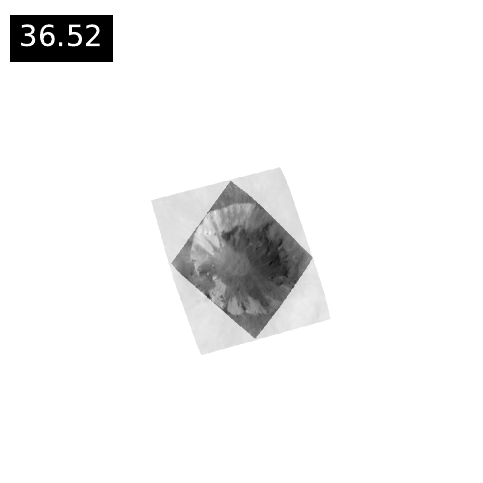} &
    \includegraphics[width=\linewidth]{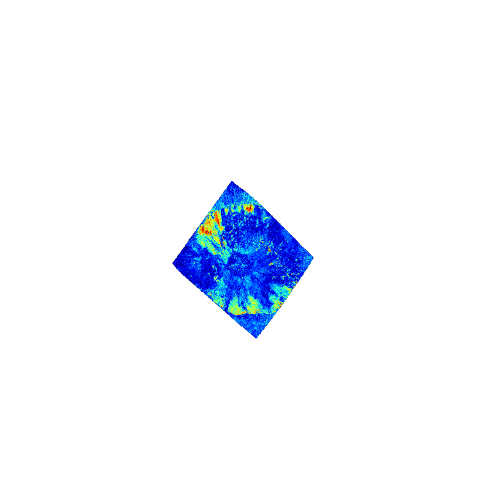} &&
    \includegraphics[width=\linewidth]{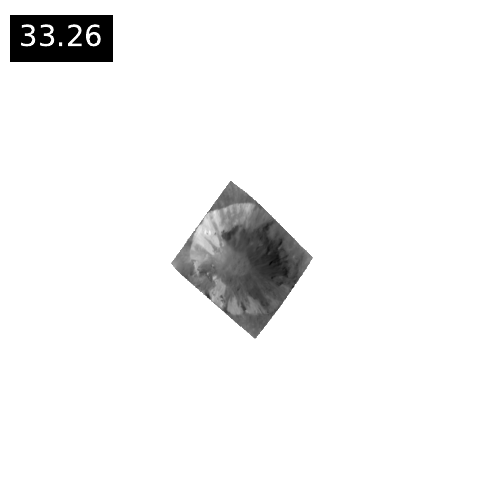} &
    \includegraphics[width=\linewidth]{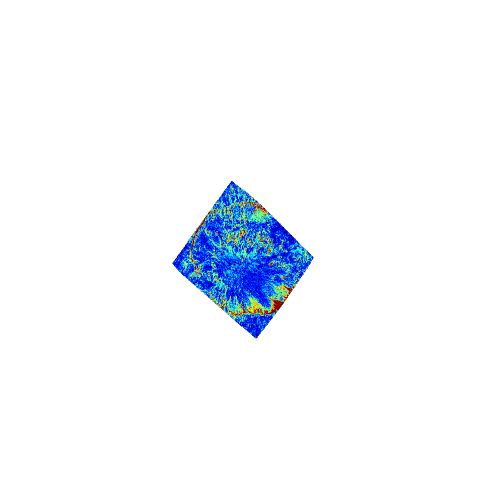} &
    \includegraphics[width=\linewidth]{figures/phomo_results/cornelia/lunar_lambert/images_rendered/err-cb.pdf} \\
    & & & & \\ 
    & & & & \\ 
    &
    \includegraphics[width=\linewidth]{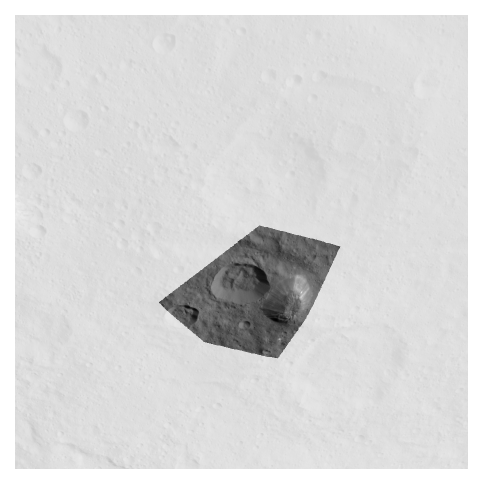} &&
    \includegraphics[width=\linewidth]{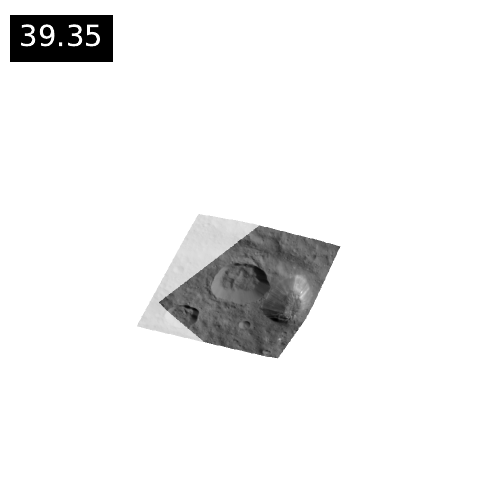} &
    \includegraphics[width=\linewidth]{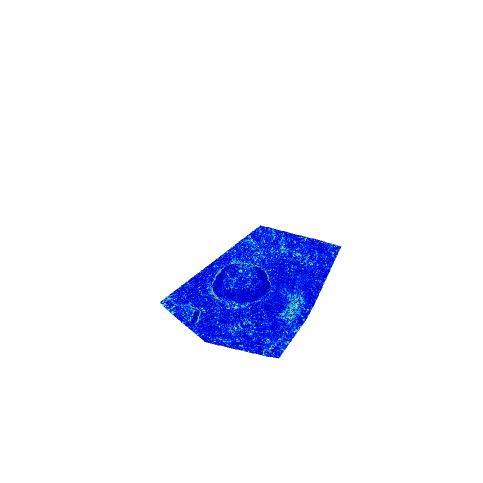} &&
    \includegraphics[width=\linewidth]{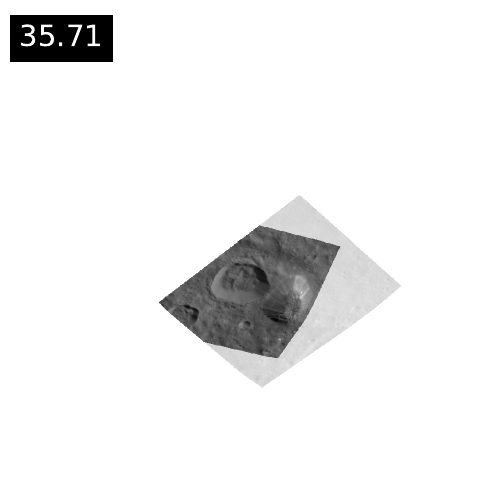} &
    \includegraphics[width=\linewidth]{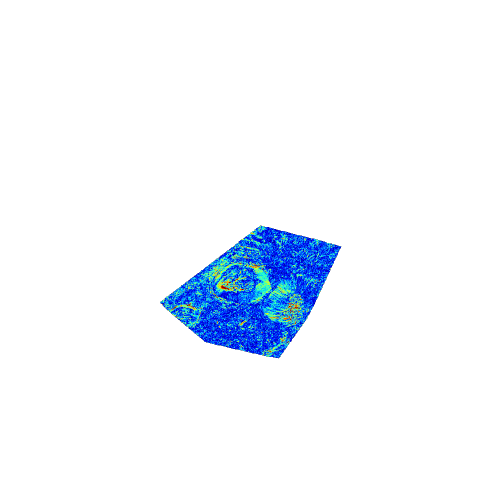} &
    \includegraphics[width=\linewidth]{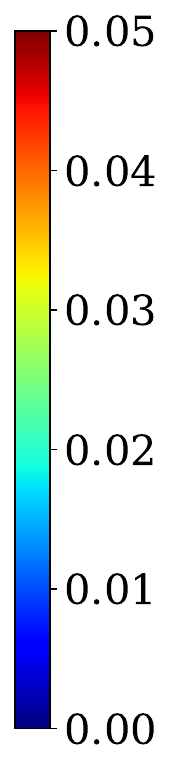} \\
    \parbox[t]{2mm}{\rotatebox[origin=l]{90}{\textcolor{gray}{\scriptsize{Ahuna Mons}}}} &
    \includegraphics[width=\linewidth]{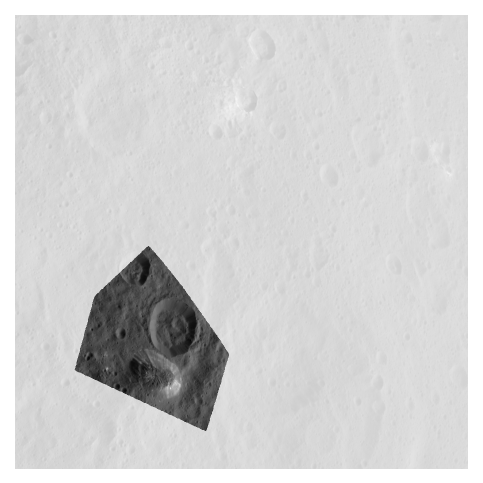} &&
    \includegraphics[width=\linewidth]{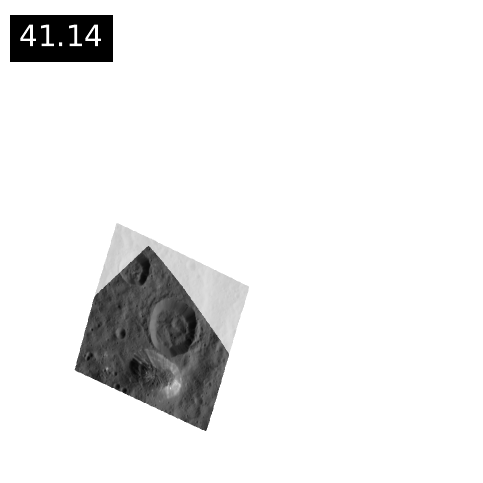} &
    \includegraphics[width=\linewidth]{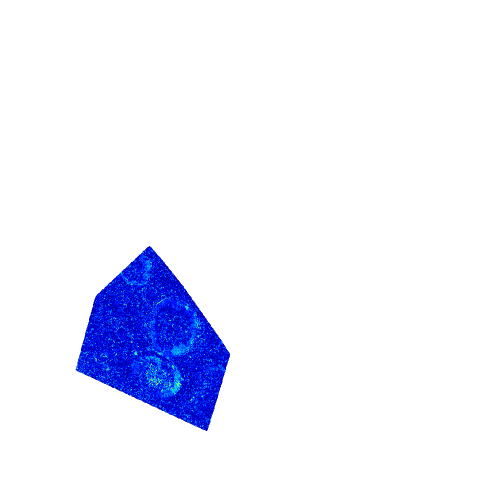} &&
    \includegraphics[width=\linewidth]{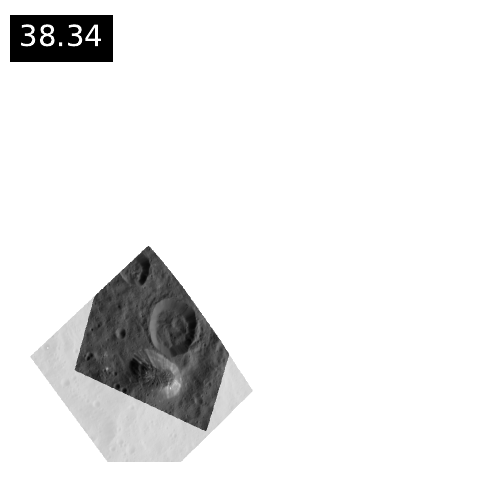} &
    \includegraphics[width=\linewidth]{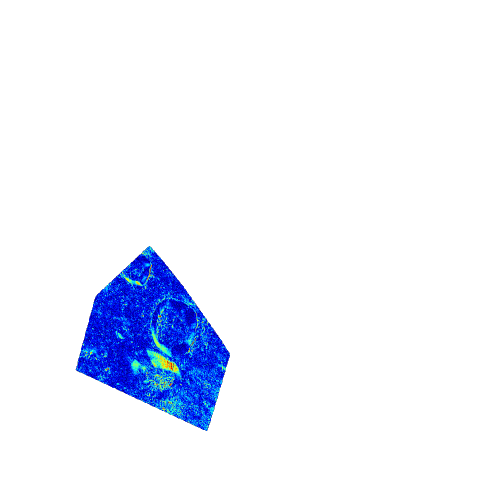} &
    \includegraphics[width=\linewidth]{figures/phomo_results/ahunamons/lunar_lambert/images_rendered/err-cb.pdf} \\
    & & & & \\ 
    & & & & \\ 
    &
    \includegraphics[width=\linewidth]{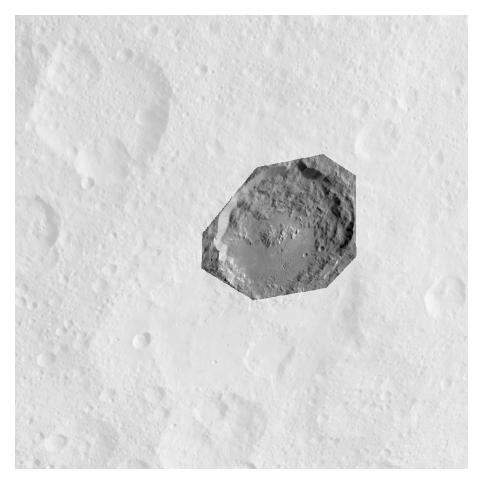} &&
    \includegraphics[width=\linewidth]{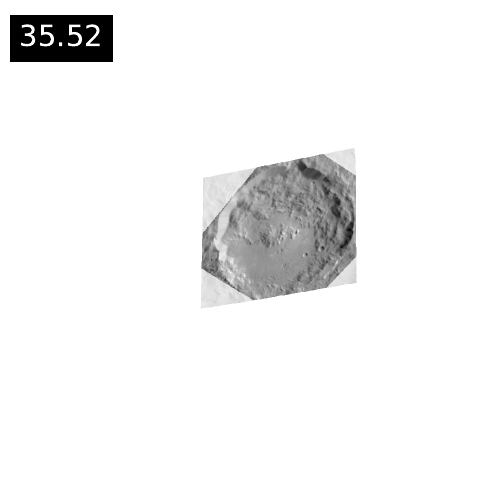} &
    \includegraphics[width=\linewidth]{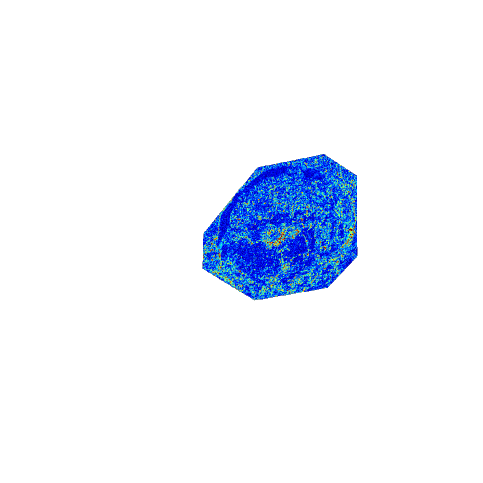} &&
    \includegraphics[width=\linewidth]{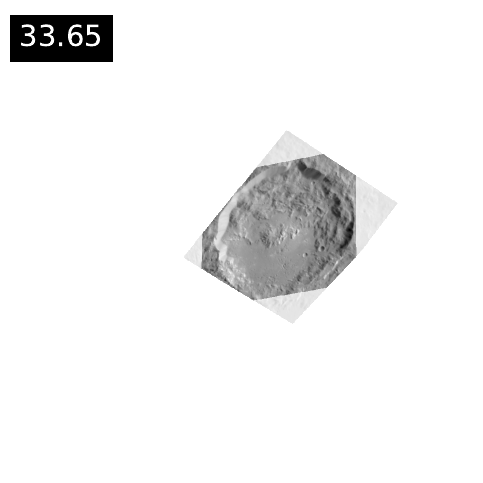} &
    \includegraphics[width=\linewidth]{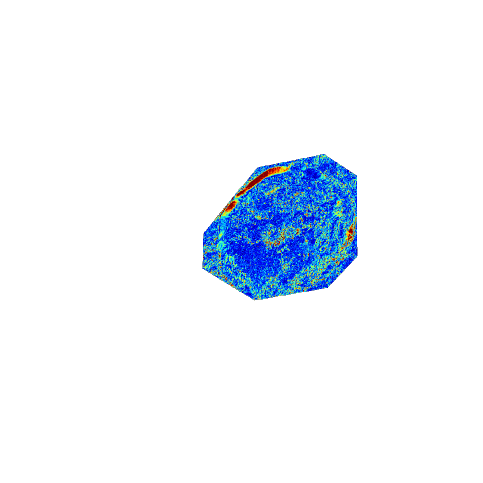} &
    \includegraphics[width=\linewidth]{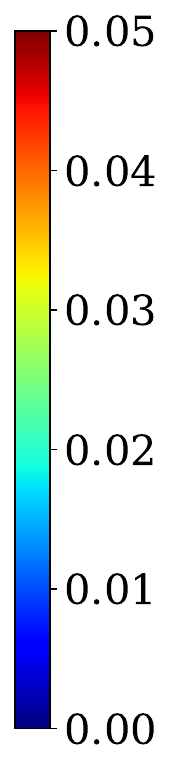} \\
    \parbox[t]{2mm}{\rotatebox[origin=l]{90}{\textcolor{gray}{\scriptsize{Ikapati}}}} &
    \includegraphics[width=\linewidth]{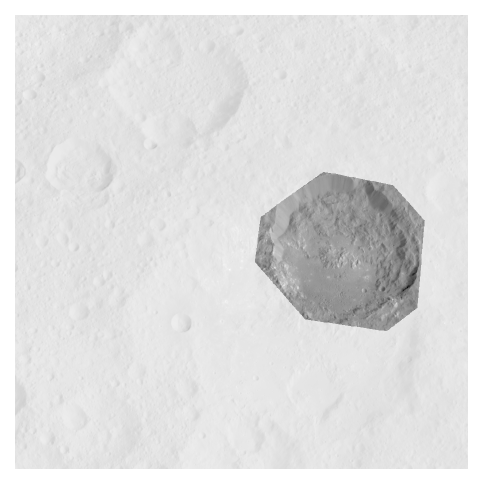} &&
    \includegraphics[width=\linewidth]{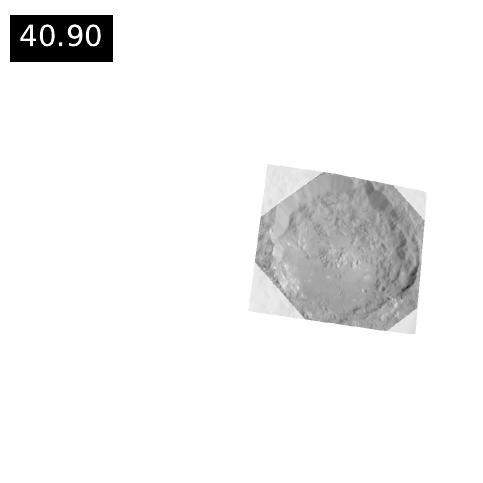} &
    \includegraphics[width=\linewidth]{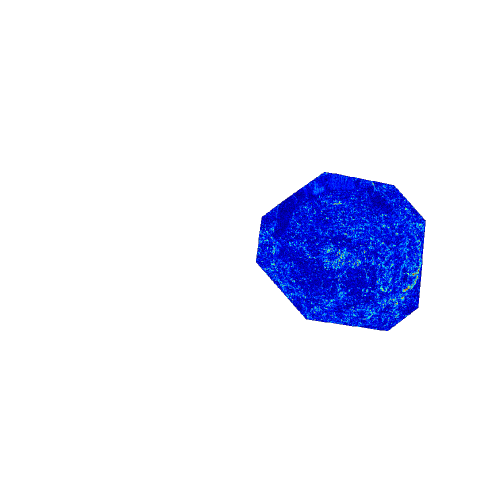} &&
    \includegraphics[width=\linewidth]{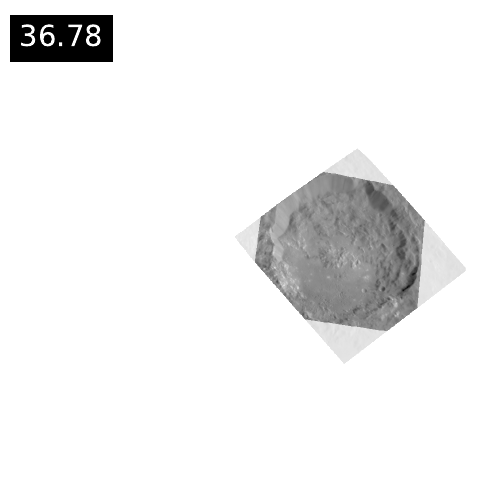} &
    \includegraphics[width=\linewidth]{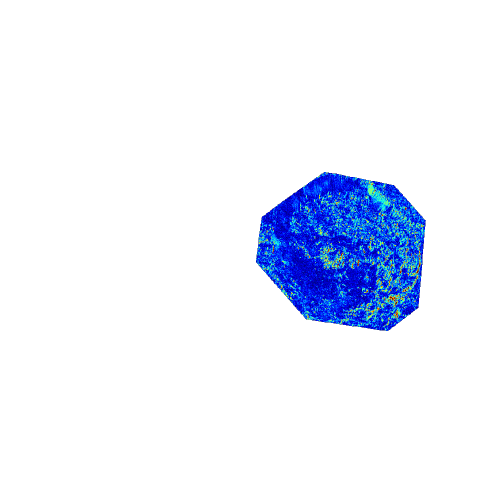} &
    \includegraphics[width=\linewidth]{figures/phomo_results/ikapati/lunar_lambert/images_rendered/err-cb.pdf} \\
    & \centering\scriptsize{Actual} && \centering\scriptsize{Render} & \centering\scriptsize{Abs. Error} && \centering\scriptsize{Render} & \centering\scriptsize{Abs. Error} \\
\end{tabular}

%% file: figures/spc-comparison.tex
\begin{tabular}{p{1pt}lcccccc}\toprule
&&& \multicolumn{3}{c}{vs. SPC} & vs. SPG & vs. SfM \\
\cmidrule(lr){4-6} \cmidrule(lr){7-7} \cmidrule(lr){8-8}
          && $\delta I$ [\%] & $\delta a$ [\%] & $\delta \vvec{n}$ [$^\circ$] & $\delta\vvec{\ell}$ [m] & $\delta\vvec{\ell}$ [m] & $\delta\vvec{\ell}$ [m] \\ 
\midrule
\multirow{5}{*}{\parbox[t]{1mm}{\rotatebox[origin=l]{90}{Cornelia}}} 
& Akimov    & \cc{gray!50} 1.19 & \cc{gray!34} 5.12 & \cc{gray!36} 5.22 & \cc{gray!38} 74.54 & \cc{gray!42} 15.53 & \cc{gray!36} 4.22 \\
& McEwen    & \cc{gray!50} 1.19 & \cc{gray!50} 4.99 & \cc{gray!15} 5.49 & \cc{gray!50} 74.42 & \cc{gray!00} 16.29 & \cc{gray!00} 5.26 \\
& Akimov+   & \cc{gray!00} 1.26 & \cc{gray!39} 5.08 & \cc{gray!50} 5.04 & \cc{gray!23} 74.68 & \cc{gray!50} 15.38 & \cc{gray!50} 3.83 \\
& L-Lambert & \cc{gray!29} 1.22 & \cc{gray!07} 5.33 & \cc{gray!09} 5.57 & \cc{gray!21} 74.70 & \cc{gray!16} 15.99 & \cc{gray!09} 5.01 \\
& Minnaert  & \cc{gray!21} 1.23 & \cc{gray!00} 5.39 & \cc{gray!00} 5.68 & \cc{gray!00} 74.89 & \cc{gray!14} 16.03 & \cc{gray!06} 5.08 \\
\midrule
\midrule
\multirow{5}{*}{\parbox[t]{1mm}{\rotatebox[origin=l]{90}{Ahuna Mons}}} 
& Akimov    & \cc{gray!00} 0.99 & \cc{gray!50} 2.21 & \cc{gray!35} 4.39 & \cc{gray!07} 54.09 & \cc{gray!01} 29.11 & \cc{gray!48} 15.65 \\
& McEwen    & \cc{gray!50} 0.78 & \cc{gray!00} 3.42 & \cc{gray!10} 4.60 & \cc{gray!00} 54.26 & \cc{gray!00} 29.17 & \cc{gray!39} 16.98 \\
& Akimov+   & \cc{gray!02} 0.98 & \cc{gray!41} 2.43 & \cc{gray!50} 4.27 & \cc{gray!02} 54.22 & \cc{gray!16} 28.41 & \cc{gray!50} 15.41 \\
& L-Lambert & \cc{gray!12} 0.94 & \cc{gray!28} 2.75 & \cc{gray!38} 4.37 & \cc{gray!35} 53.36 & \cc{gray!50} 26.75 & \cc{gray!22} 19.42 \\
& Minnaert  & \cc{gray!10} 0.95 & \cc{gray!24} 2.85 & \cc{gray!00} 4.68 & \cc{gray!50} 52.99 & \cc{gray!47} 26.91 & \cc{gray!00} 22.68 \\
\midrule
\midrule
\multirow{5}{*}{\parbox[t]{1mm}{\rotatebox[origin=l]{90}{Ikapati}}} 
& Akimov    & \cc{gray!47} 1.33 & \cc{gray!50} 2.10 & \cc{gray!00} 4.19 & \cc{gray!00} 39.16 & \cc{gray!00} 24.76 & \cc{gray!00} 13.08 \\
& McEwen    & \cc{gray!50} 1.32 & \cc{gray!40} 2.43 & \cc{gray!43} 3.54 & \cc{gray!24} 37.74 & \cc{gray!23} 23.12 & \cc{gray!50} 11.60 \\
& Akimov+   & \cc{gray!06} 1.46 & \cc{gray!32} 2.70 & \cc{gray!50} 3.44 & \cc{gray!35} 37.08 & \cc{gray!41} 21.82 & \cc{gray!06} 12.91 \\
& L-Lambert & \cc{gray!03} 1.47 & \cc{gray!20} 3.08 & \cc{gray!41} 3.58 & \cc{gray!34} 37.15 & \cc{gray!44} 21.54 & \cc{gray!23} 12.41 \\
& Minnaert  & \cc{gray!00} 1.48 & \cc{gray!00} 3.74 & \cc{gray!39} 3.60 & \cc{gray!50} 36.21 & \cc{gray!50} 21.14 & \cc{gray!01} 13.04 \\
\bottomrule
\end{tabular}

%% file: figures/spc-spg-sfm-comparison.tex

\centering
\setlength{\extrarowheight}{-10pt}
\setlength{\tabcolsep}{2pt} 
\begin{tabular}{cR{4.1cm}R{4.1cm}R{4.1cm}p{.3cm}R{4.1cm}p{.3cm}R{4.1cm}}
    \parbox[t]{2mm}{\rotatebox[origin=l]{90}{\textcolor{gray}{\small{Cornelia}}}} &
    \includegraphics[width=.93\linewidth]{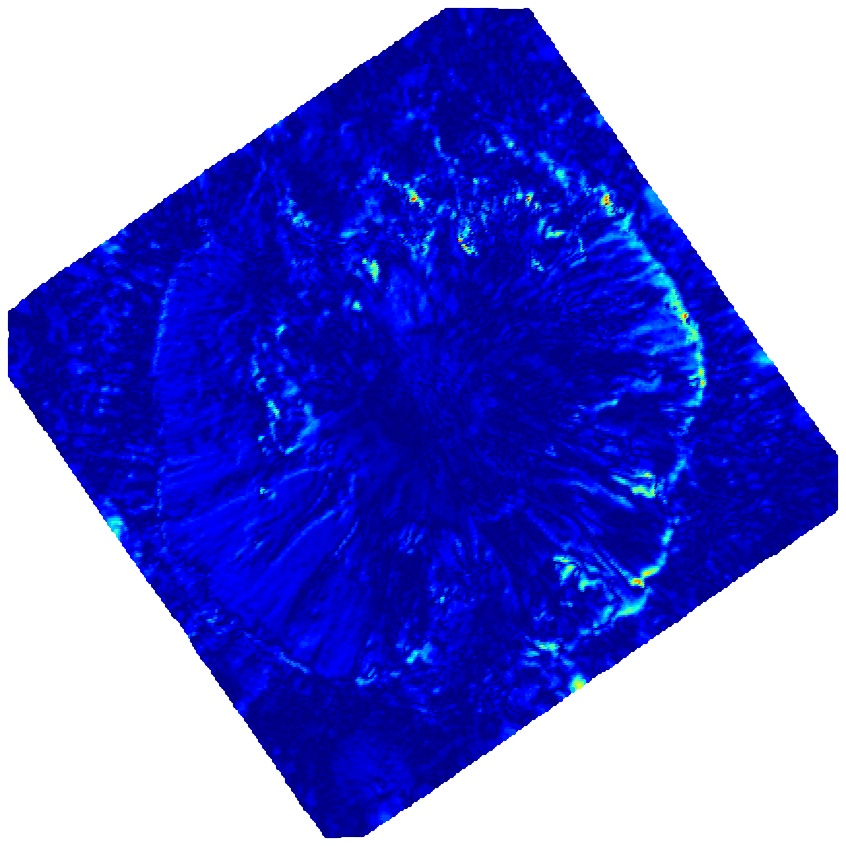} &
    \includegraphics[width=.93\linewidth]{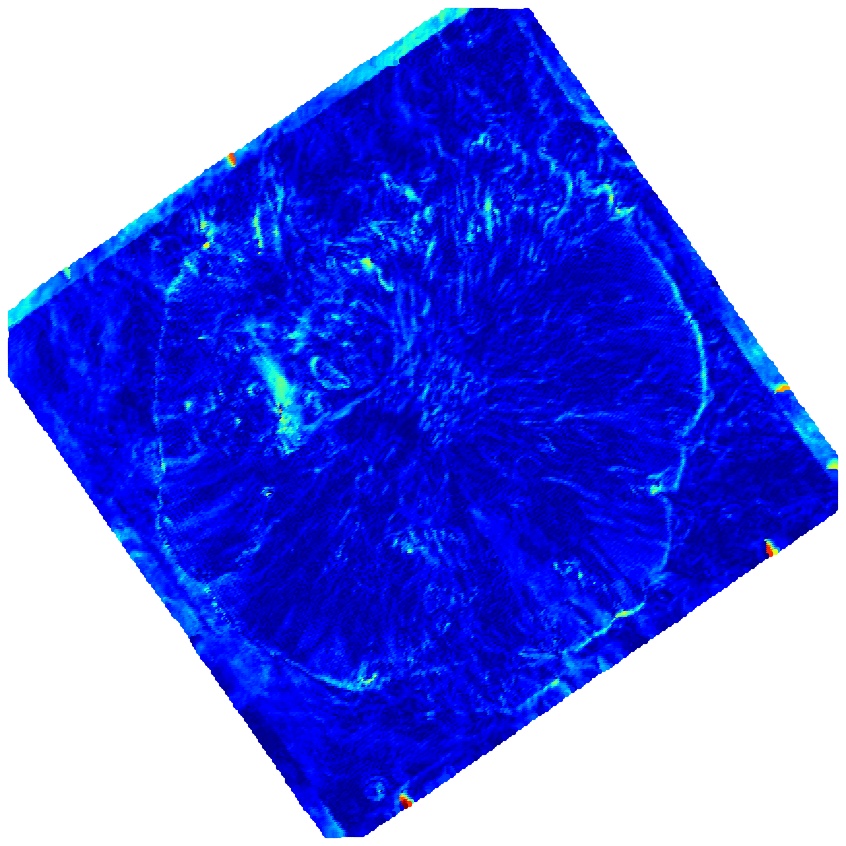} &
    \includegraphics[width=.93\linewidth]{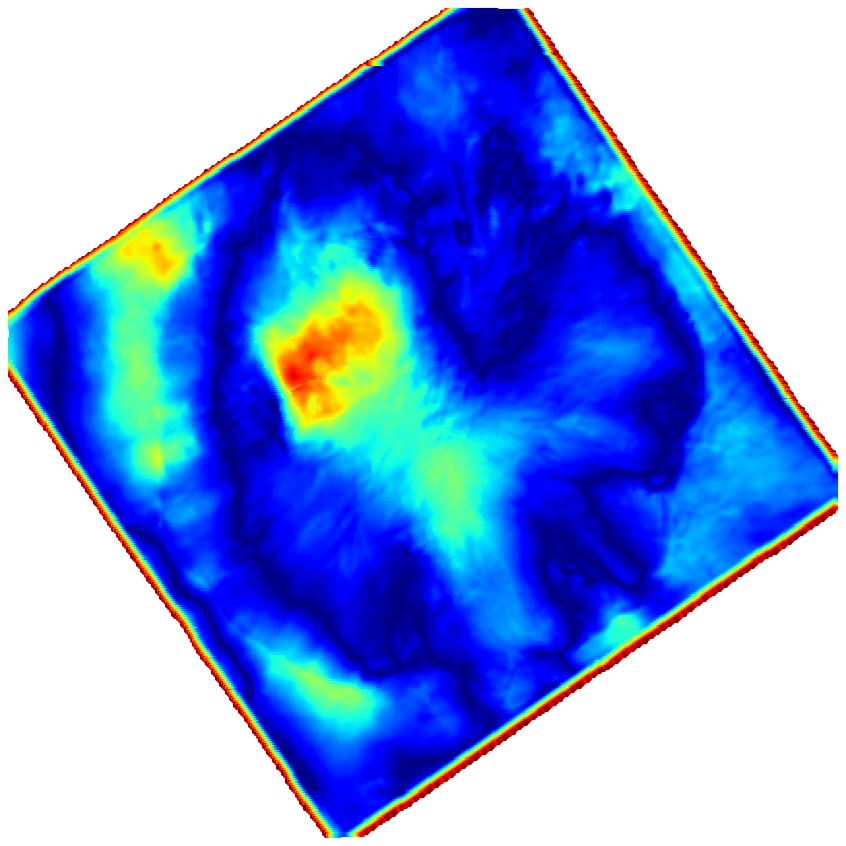} &&
    \includegraphics[width=.93\linewidth]{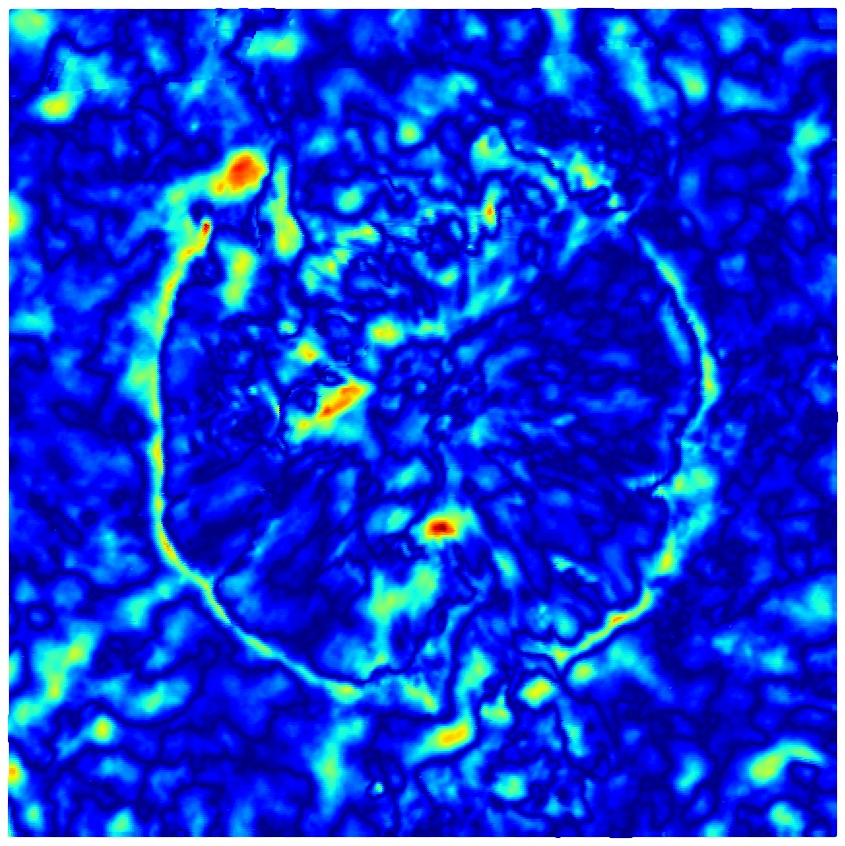} &&
    \includegraphics[width=.93\linewidth]{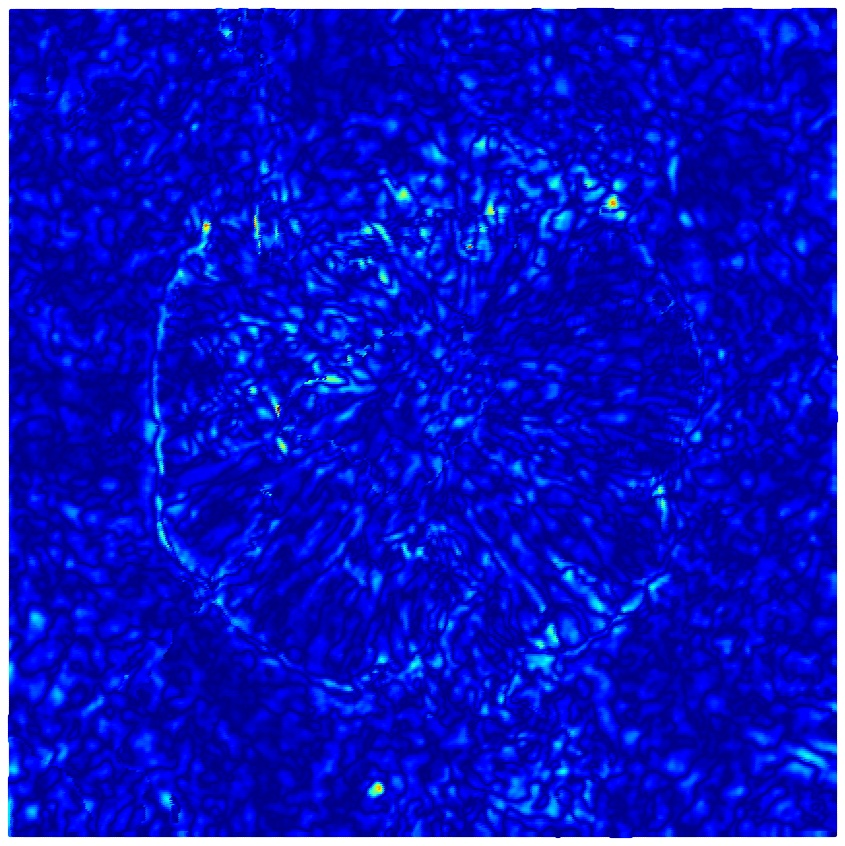} \\
    &
    \includegraphics[width=\linewidth]{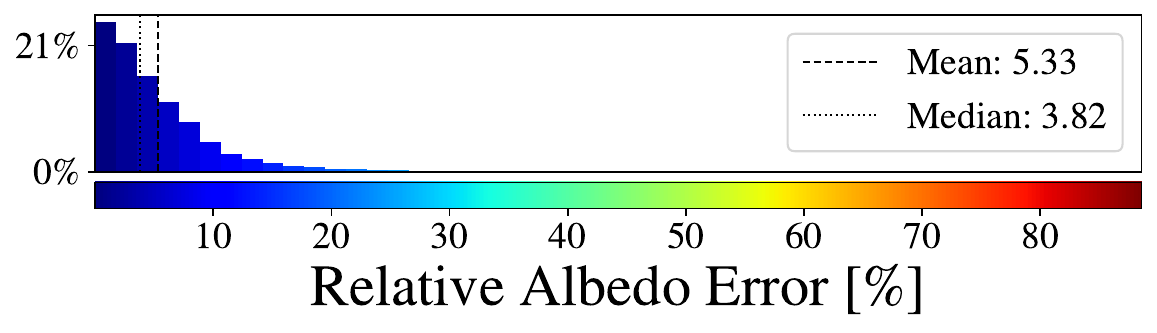} &
    \includegraphics[width=\linewidth]{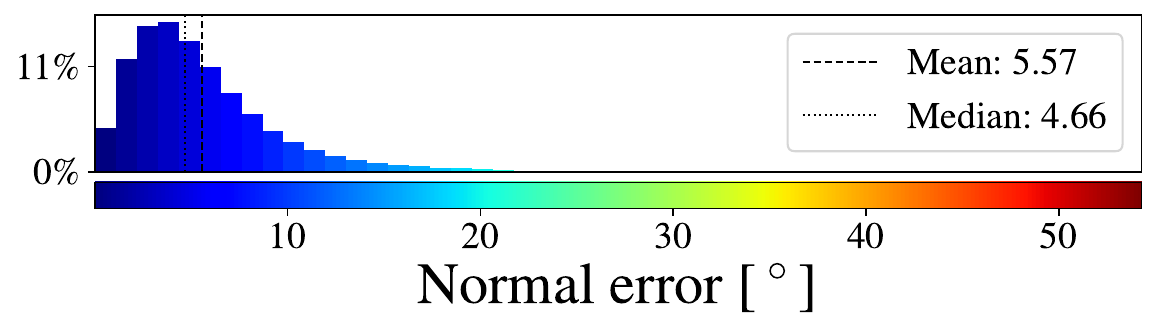} &
    \includegraphics[width=\linewidth]{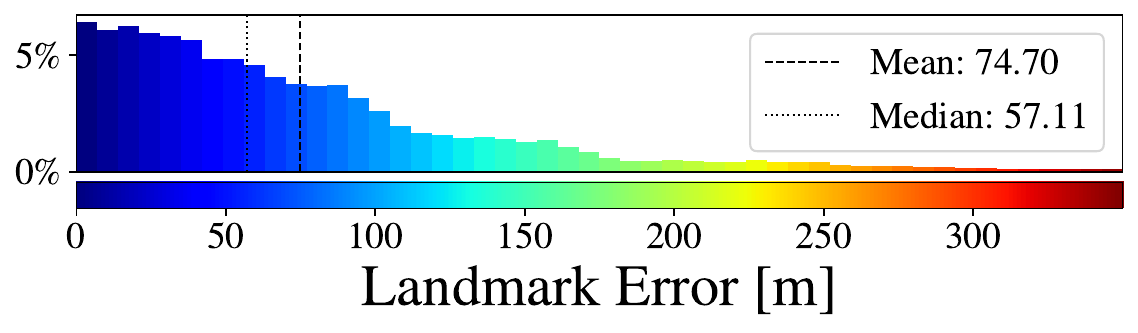} &&
    \includegraphics[width=\linewidth]{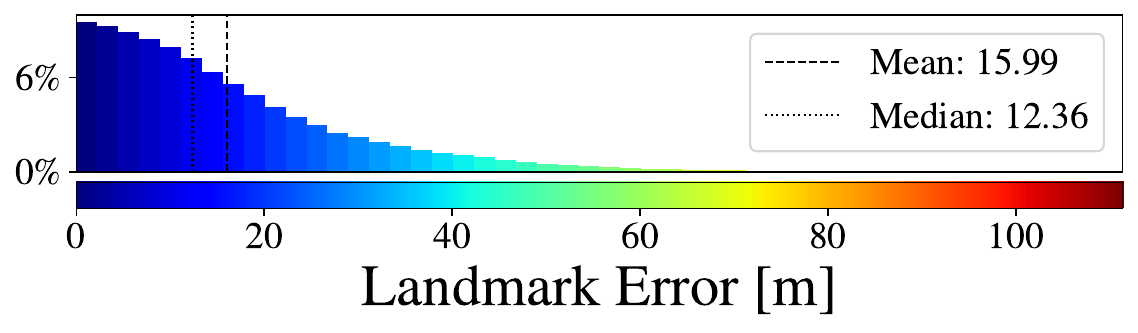} &&
    \includegraphics[width=\linewidth]{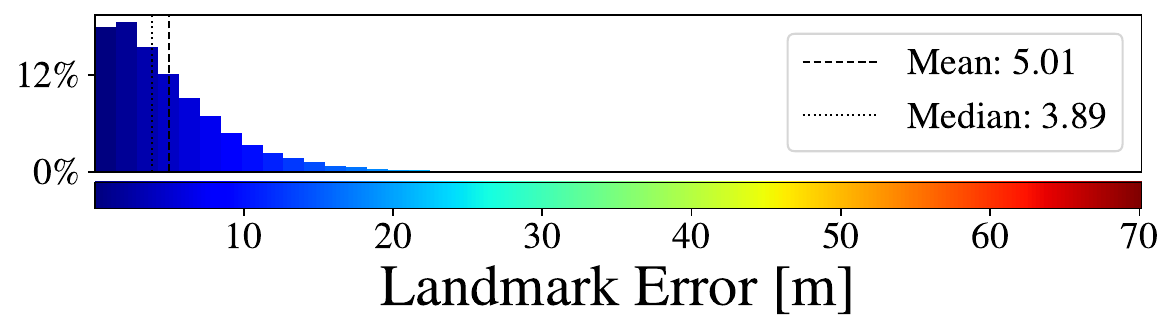} \\ \\
    \parbox[t]{2mm}{\rotatebox[origin=l]{90}{\textcolor{gray}{\small{Ahuna Mons}}}} &
    \includegraphics[width=.93\linewidth]{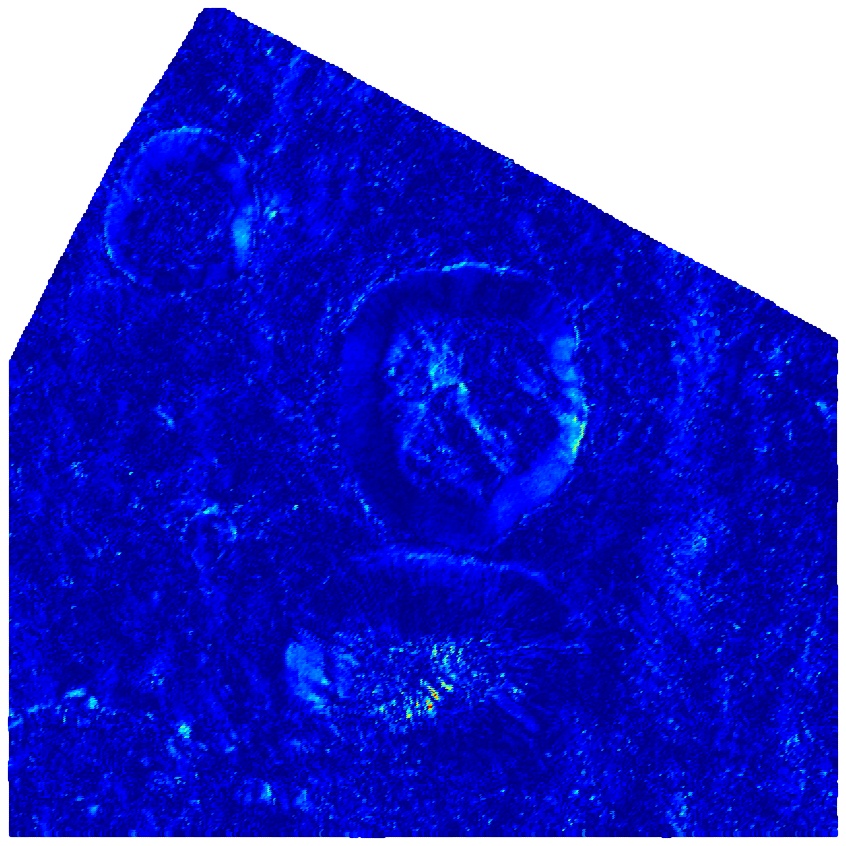} &
    \includegraphics[width=.93\linewidth]{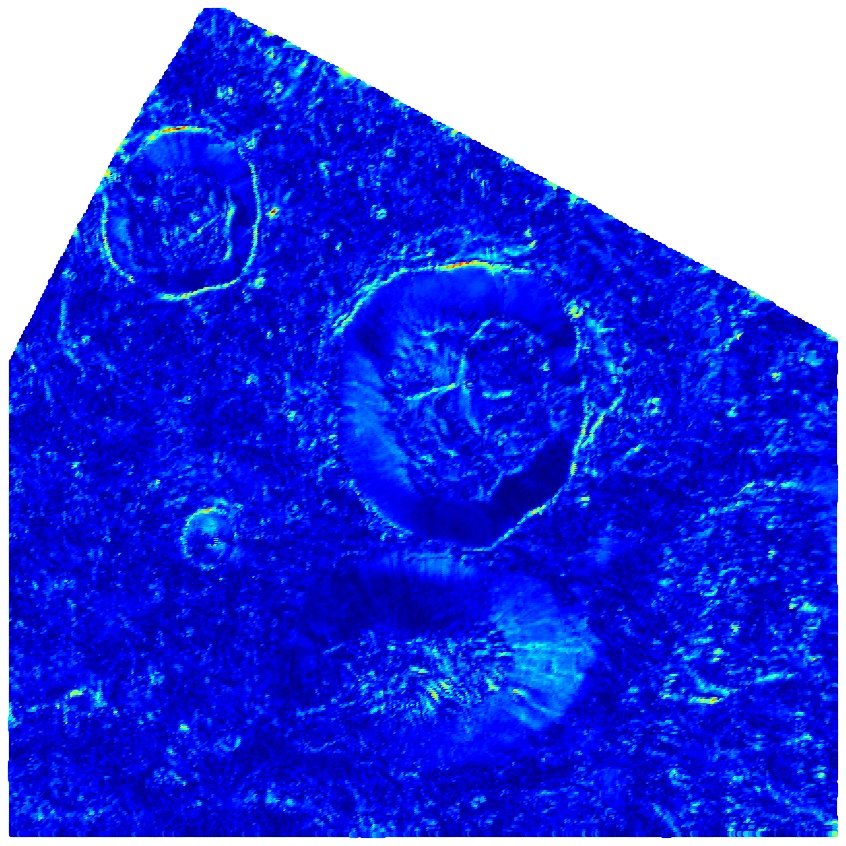} &
    \includegraphics[width=.93\linewidth]{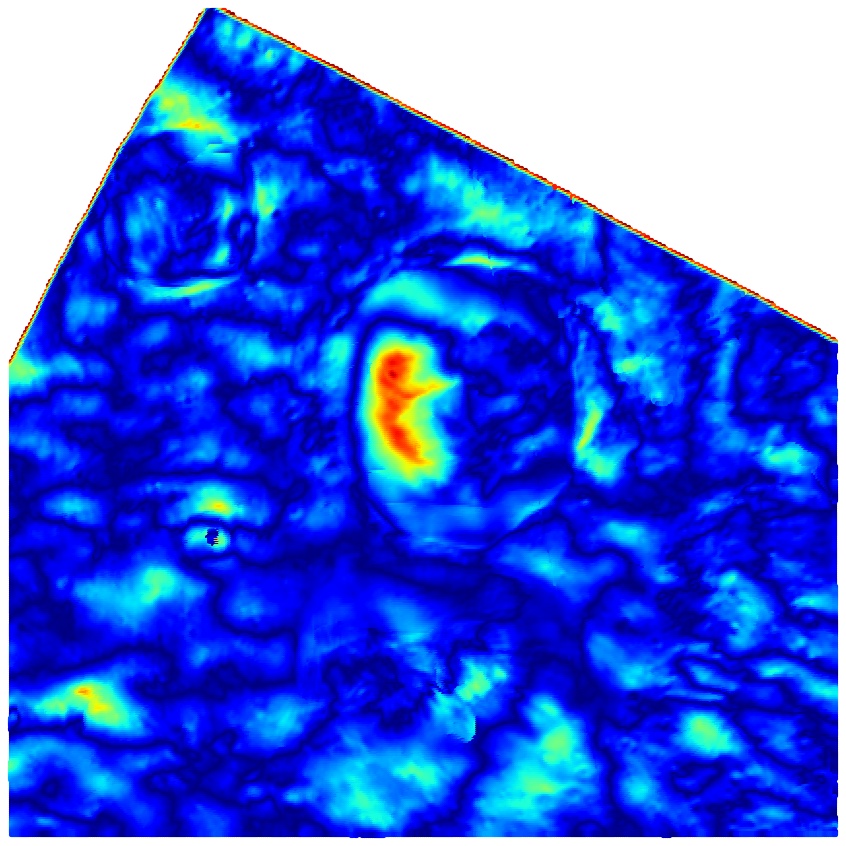} &&
    \includegraphics[width=.93\linewidth]{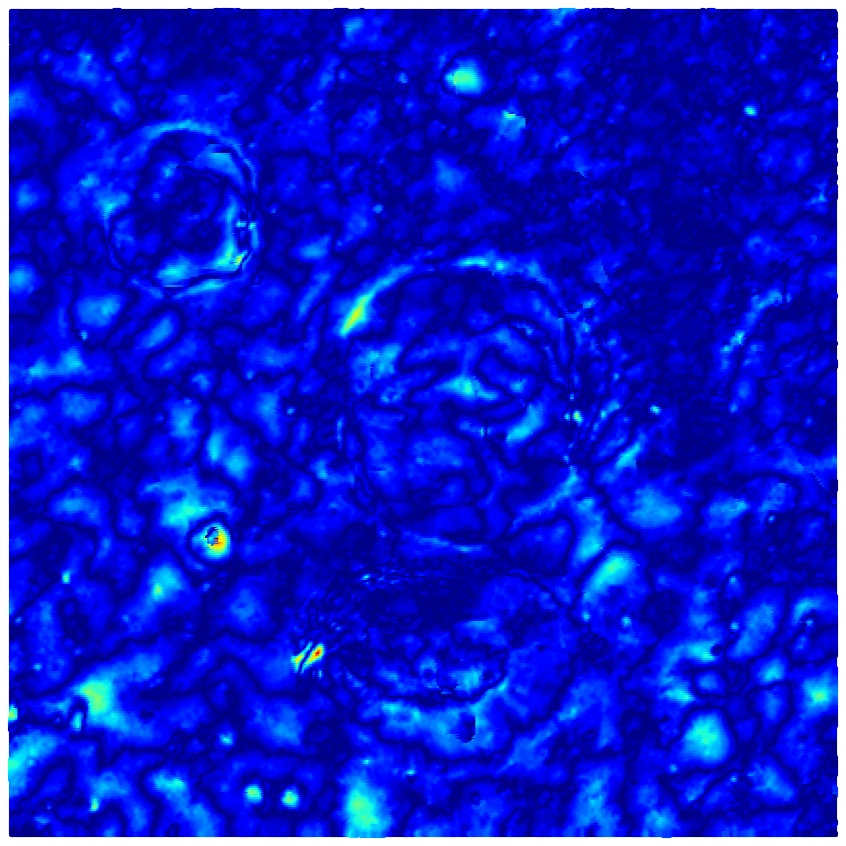} &&
    \includegraphics[width=.93\linewidth]{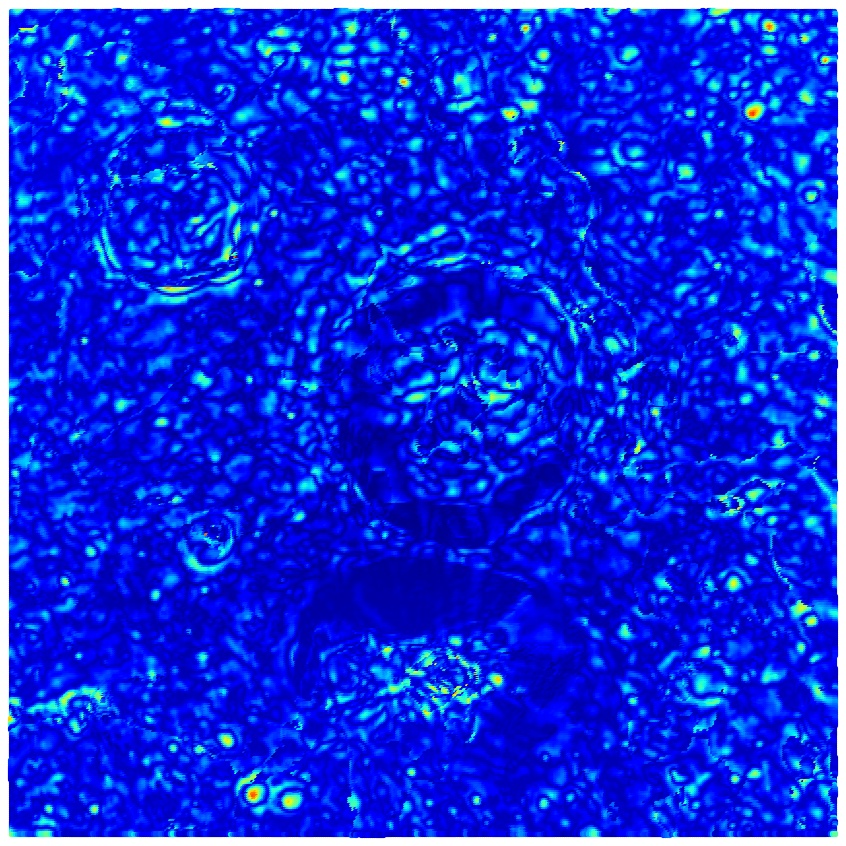} \\
    &
    \includegraphics[width=\linewidth]{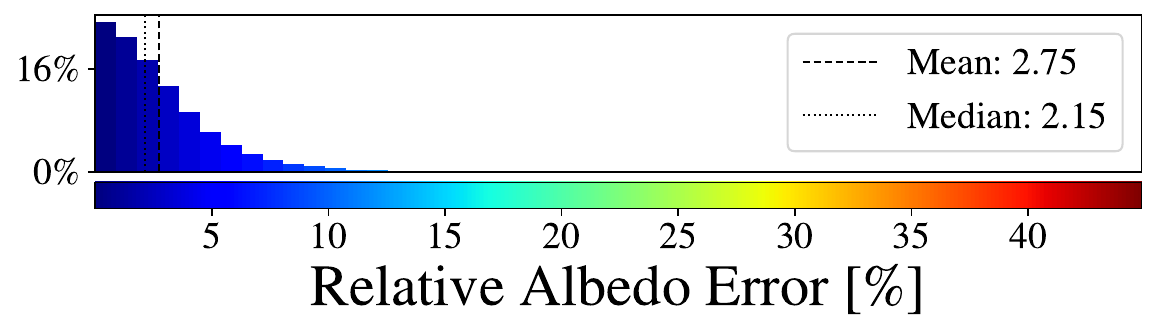} &
    \includegraphics[width=\linewidth]{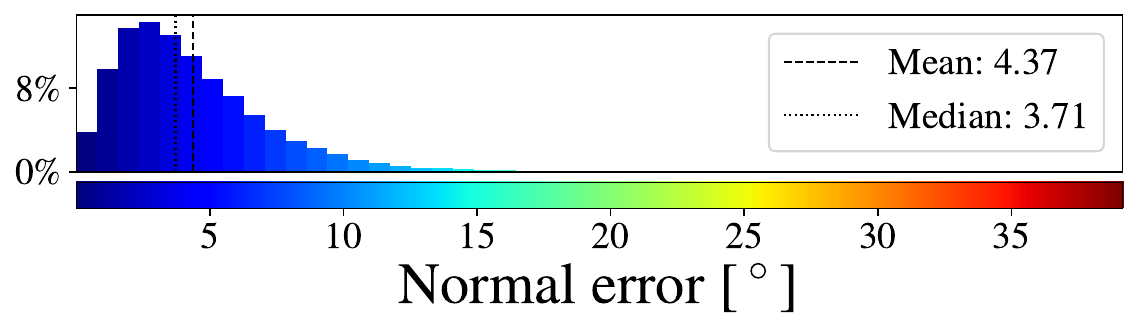} &
    \includegraphics[width=\linewidth]{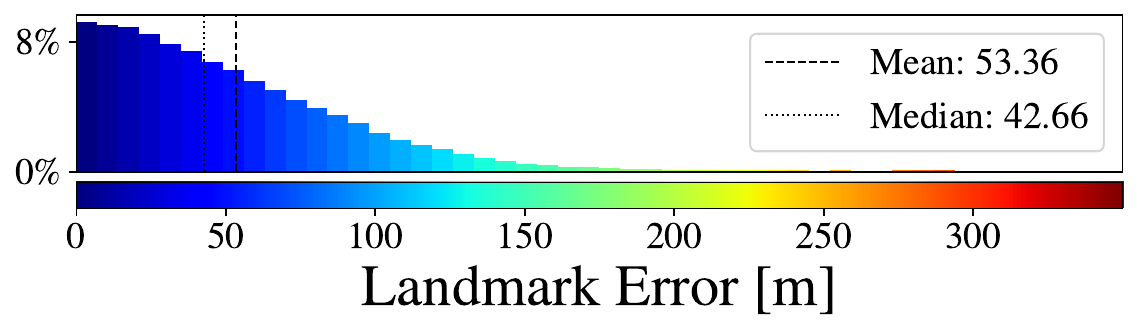} &&
    \includegraphics[width=\linewidth]{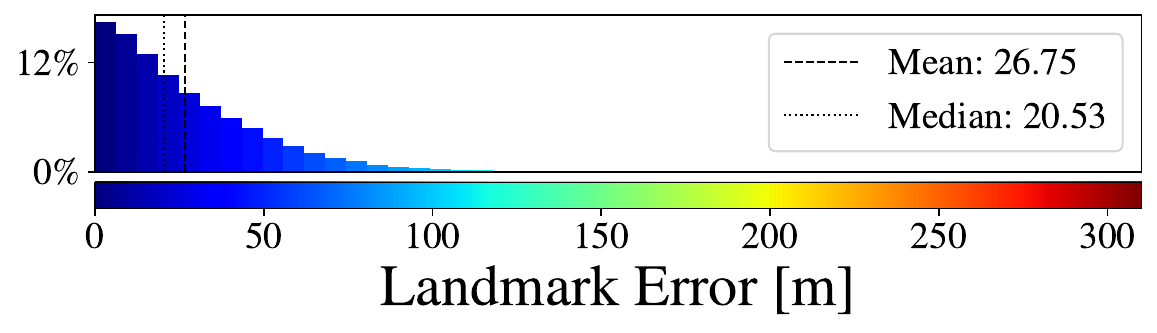} &&
    \includegraphics[width=\linewidth]{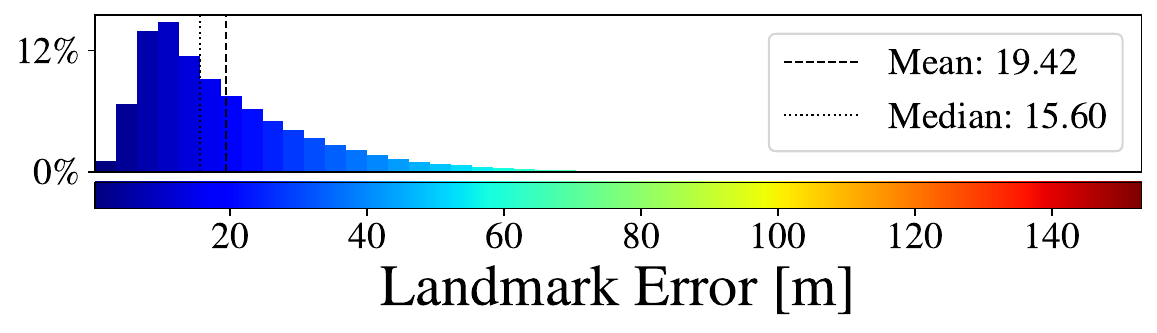} \\ \\
    \parbox[t]{2mm}{\rotatebox[origin=l]{90}{\textcolor{gray}{\small{Ikapati}}}} &
    \includegraphics[width=.93\linewidth]{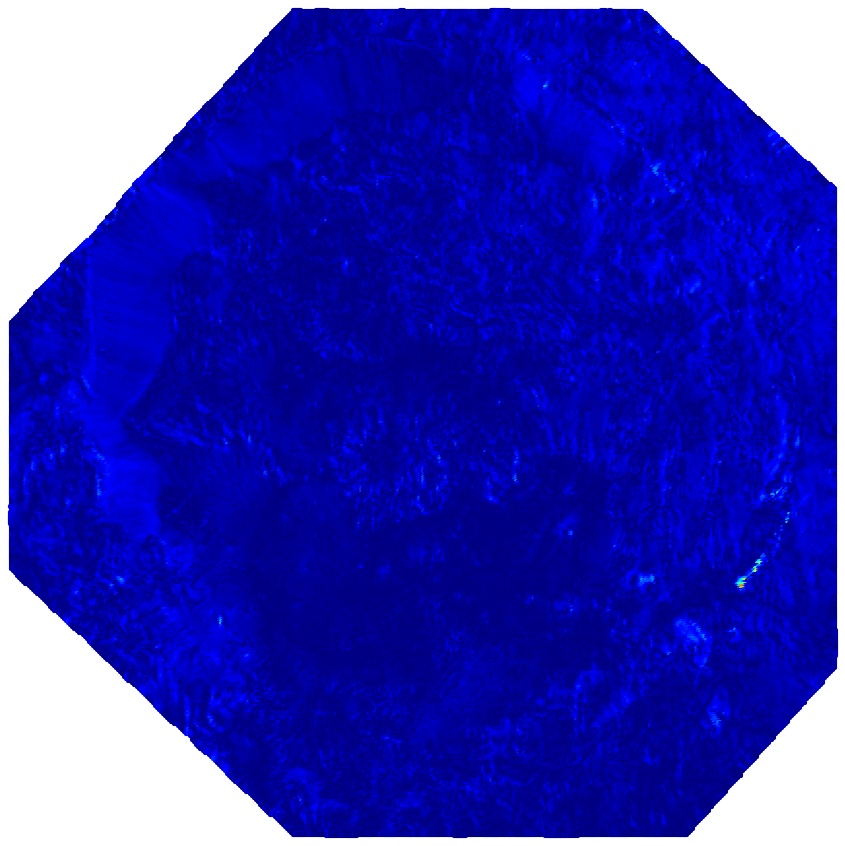} &
    \includegraphics[width=.93\linewidth]{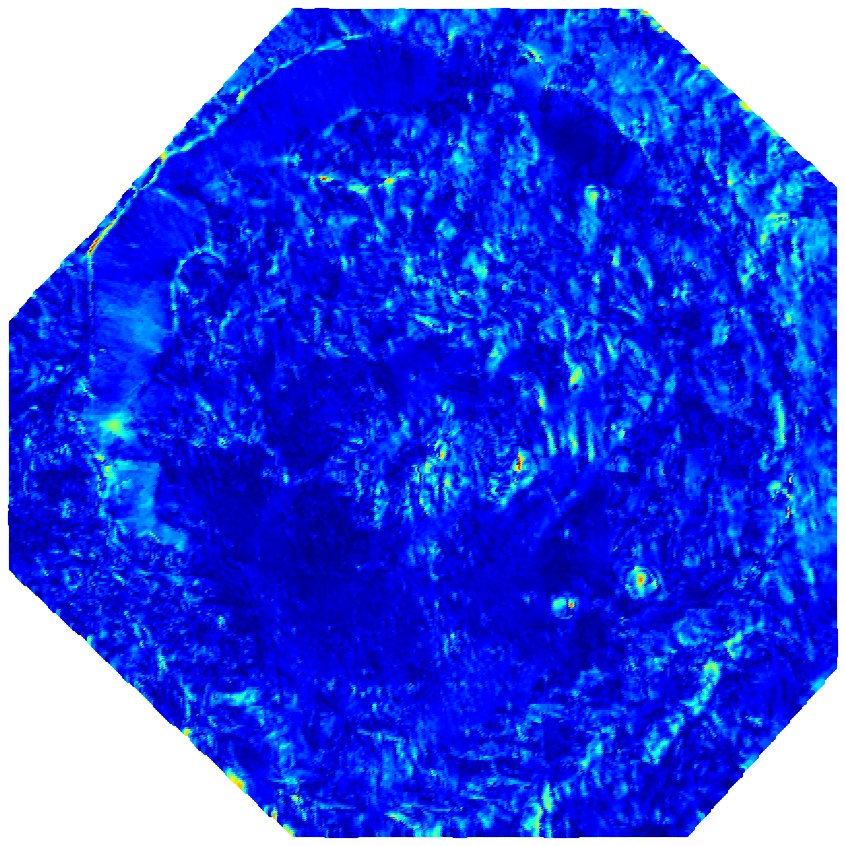} &
    \includegraphics[width=.93\linewidth]{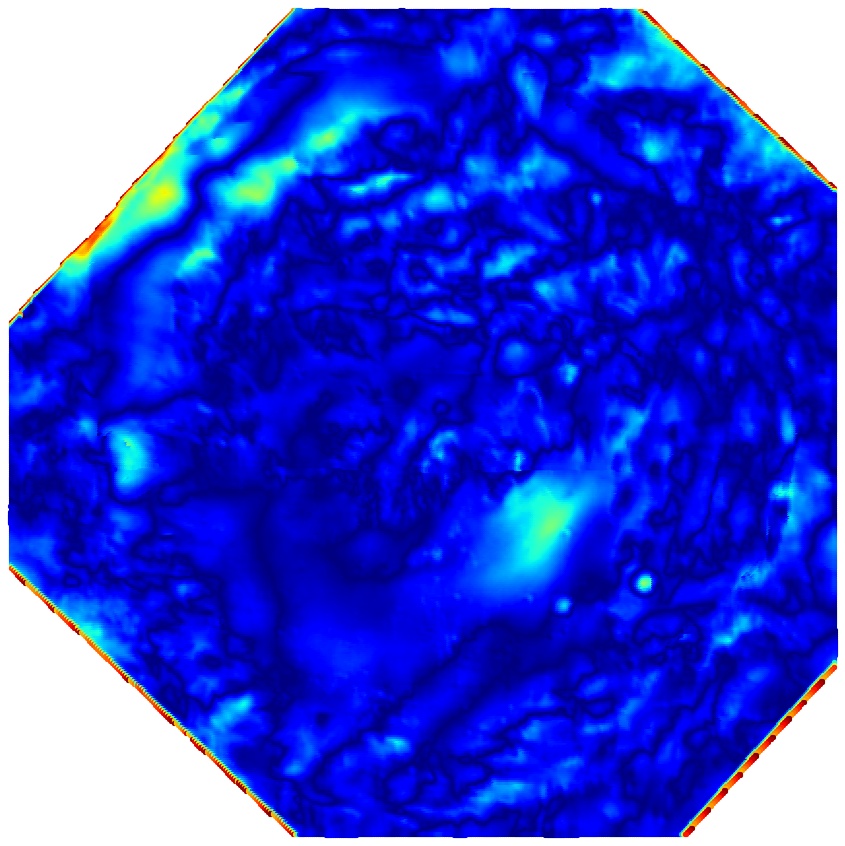} &&
    \includegraphics[width=.93\linewidth]{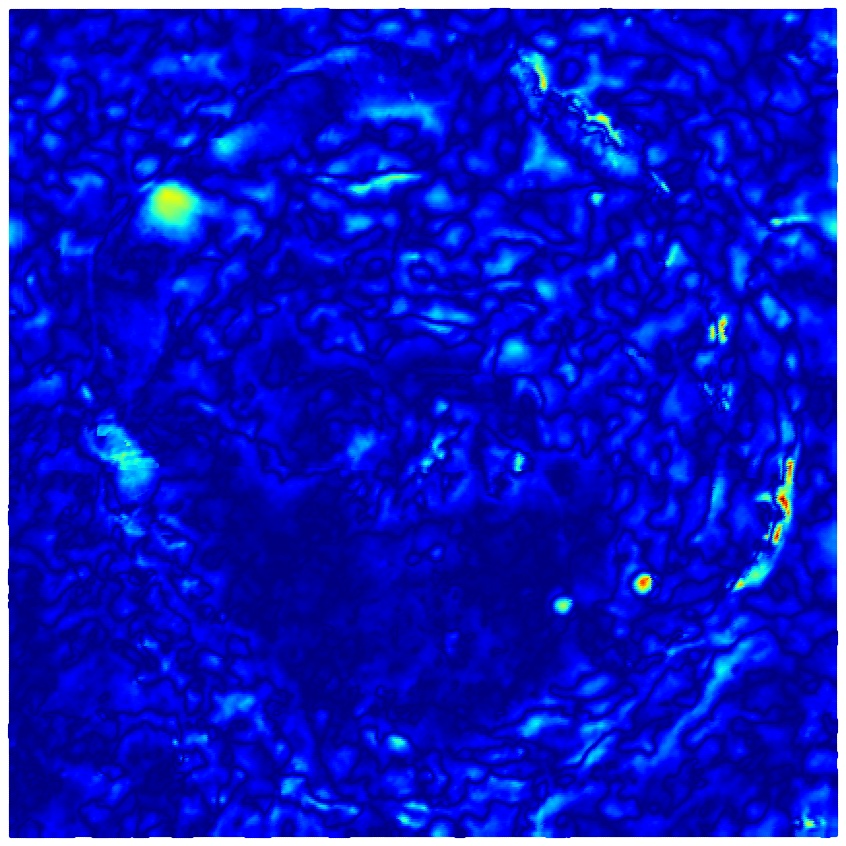} &&
    \includegraphics[width=.93\linewidth]{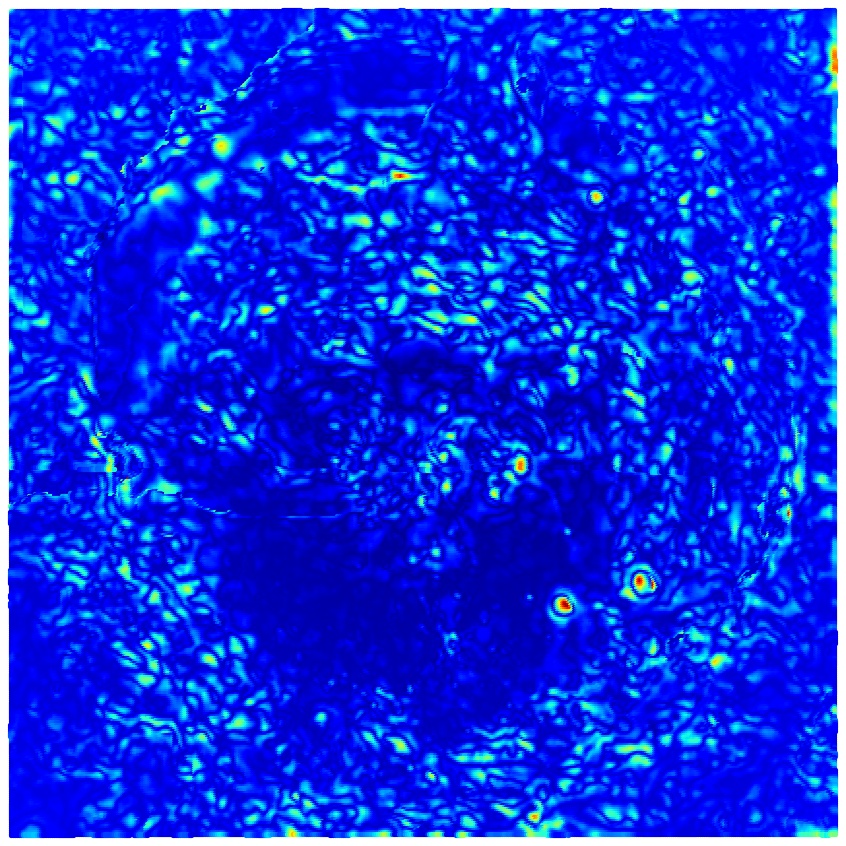} \\
    &
    \includegraphics[width=\linewidth]{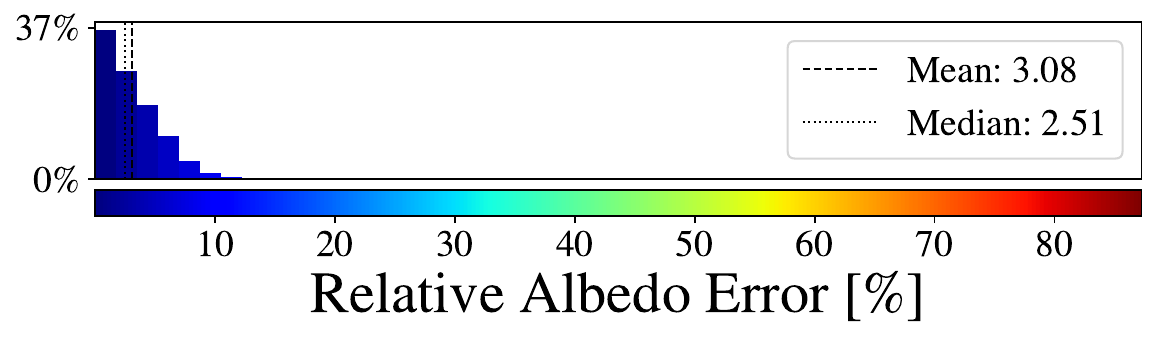} &
    \includegraphics[width=\linewidth]{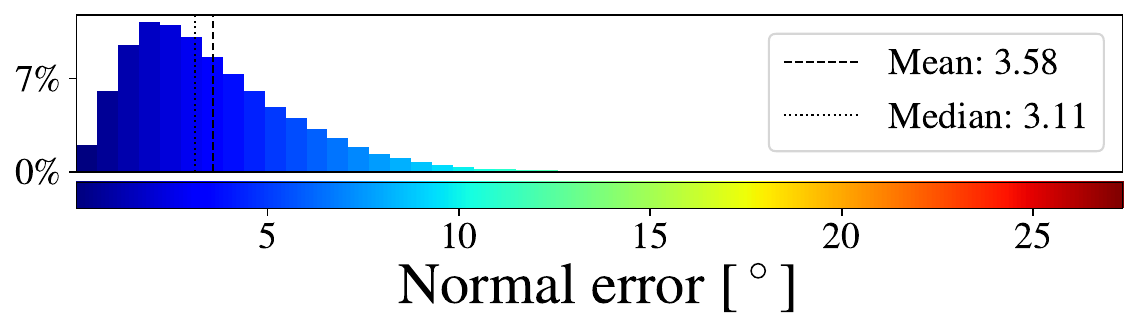} &
    \includegraphics[width=\linewidth]{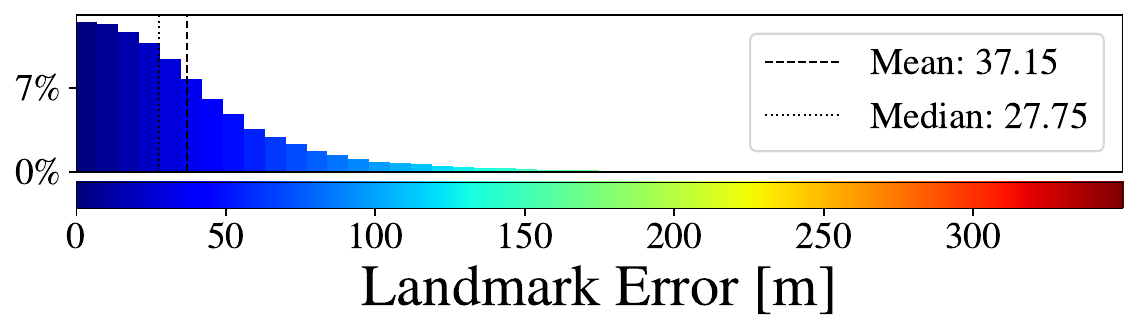} &&
    \includegraphics[width=\linewidth]{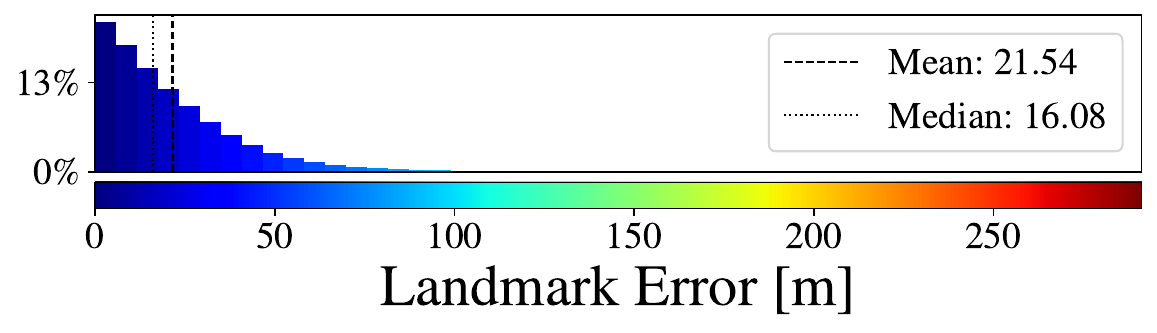} &&
    \includegraphics[width=\linewidth]{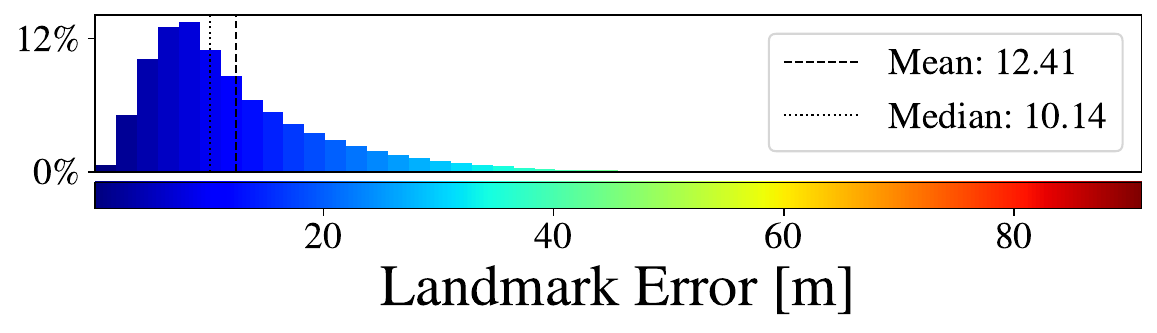} \\
    & 
    \multicolumn{3}{c}{\scriptsize{\textbf{SPC Comparison}}} && \multicolumn{1}{c}{\scriptsize{\textbf{SPG Comparison}}} && \multicolumn{1}{c}{\scriptsize{\textbf{SfM Comparison}}} \\
\end{tabular}

%% file: figures/phomo-albedos-normals.tex
\begin{subfigure}[t]{\linewidth}
\centering
  \begin{subfigure}[t]{0.27\linewidth}
    \vskip 0pt
    \includegraphics[width=\linewidth]{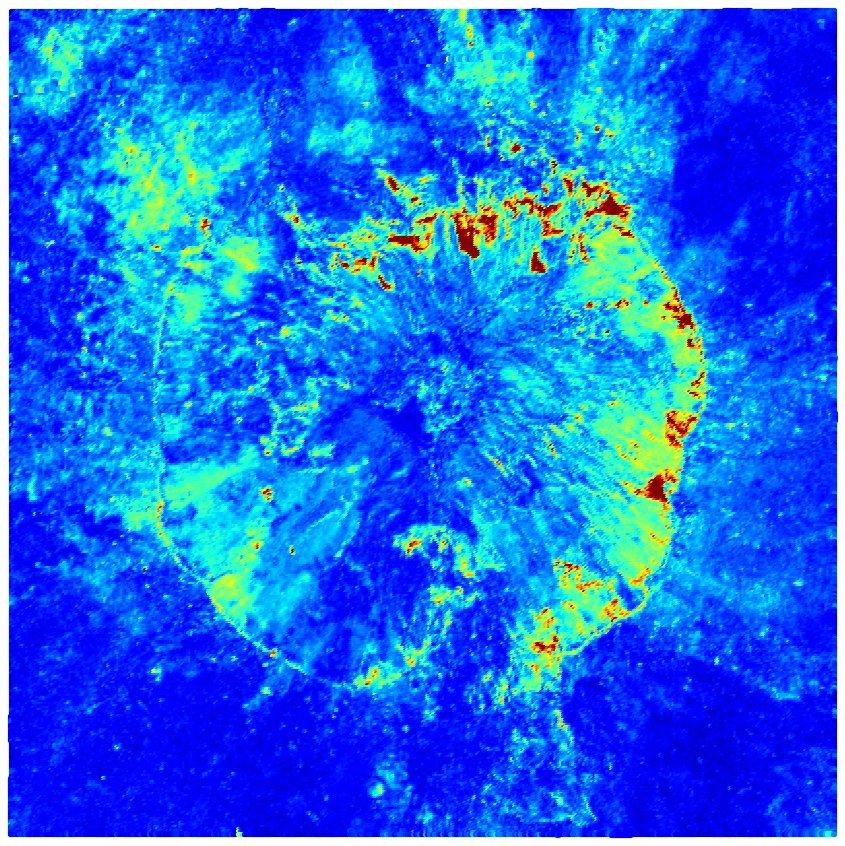}
    \vspace{-18pt}
    \caption*{\scriptsize{\textnormal{Photometric Error, $\delta I_j$ (mean = $1.22$\%)}}}
  \end{subfigure}%
  \begin{subfigure}[t]{0.04725\linewidth}
    \vskip 0pt
    \includegraphics[width=\linewidth]{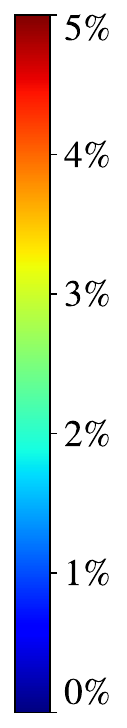}
  \end{subfigure}%
  \hspace{2pt}
  \begin{subfigure}[t]{0.27\linewidth}
    \vskip 0pt
    \includegraphics[width=\linewidth]{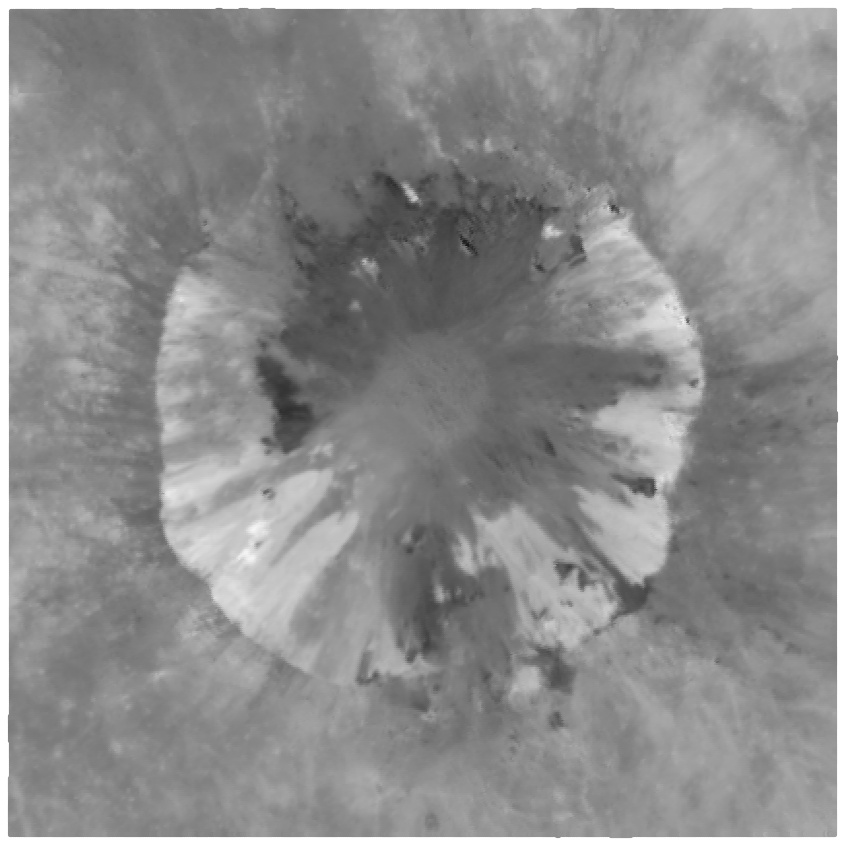}
    \vspace{-18pt}
    \caption*{\scriptsize{\textnormal{Albedos, $a_j$}}}
  \end{subfigure}%
  \begin{subfigure}[t]{0.04725\linewidth}
    \vskip 0pt
    \includegraphics[width=\linewidth]{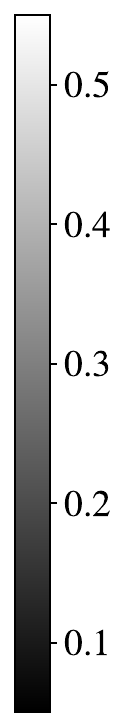}
  \end{subfigure}%
  \hfill
  \begin{subfigure}[t]{0.27\linewidth}
    \vskip 0pt
    \includegraphics[width=\linewidth]{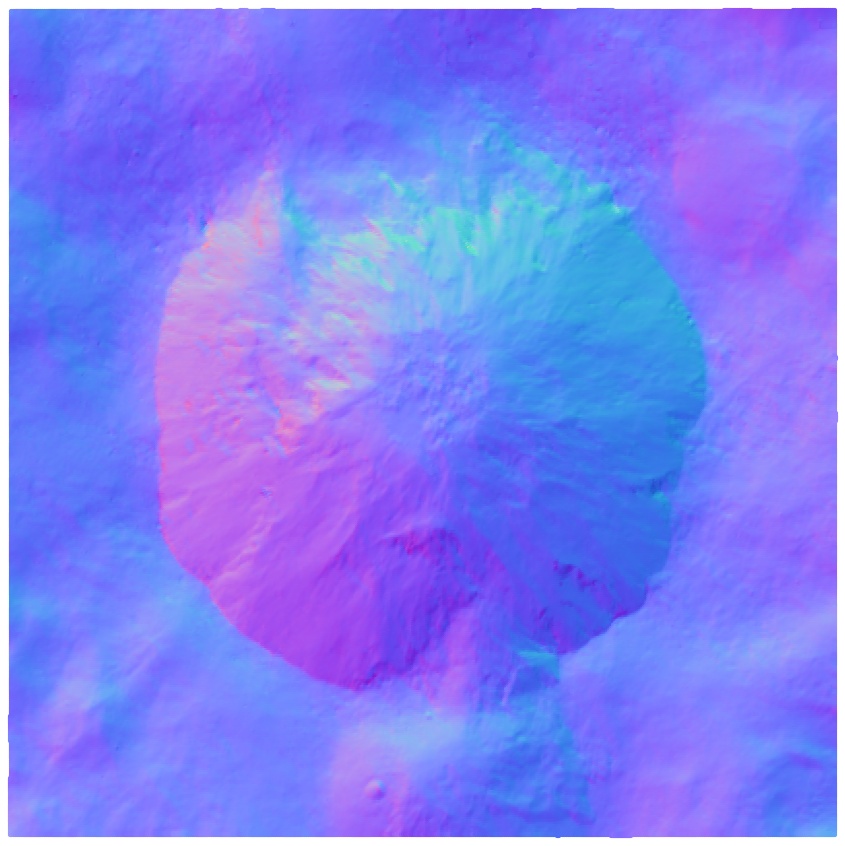}
    \vspace{-18pt}
    \caption*{\scriptsize{\textnormal{Normals, $\vvec{n}_j$}}}
  \end{subfigure}%
  \begin{subfigure}[t]{0.07\linewidth}
    \vskip 0pt
    \includegraphics[width=\linewidth]{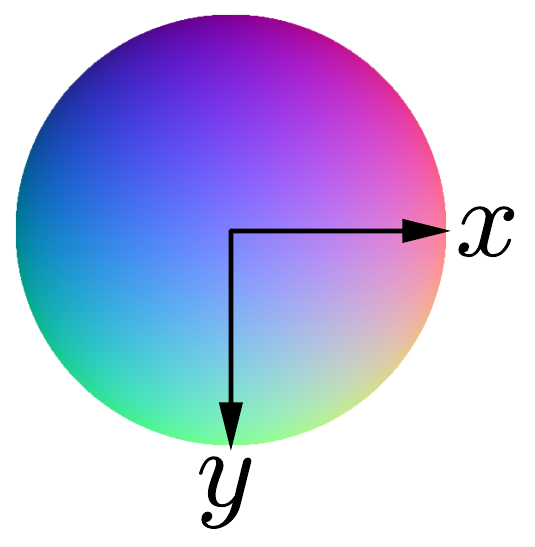}
  \end{subfigure}%
\end{subfigure}\\
\vspace{10pt}
\begin{subfigure}[t]{\linewidth}
\centering
  \begin{subfigure}[t]{0.27\linewidth}
    \vskip 0pt
    \includegraphics[width=\linewidth]{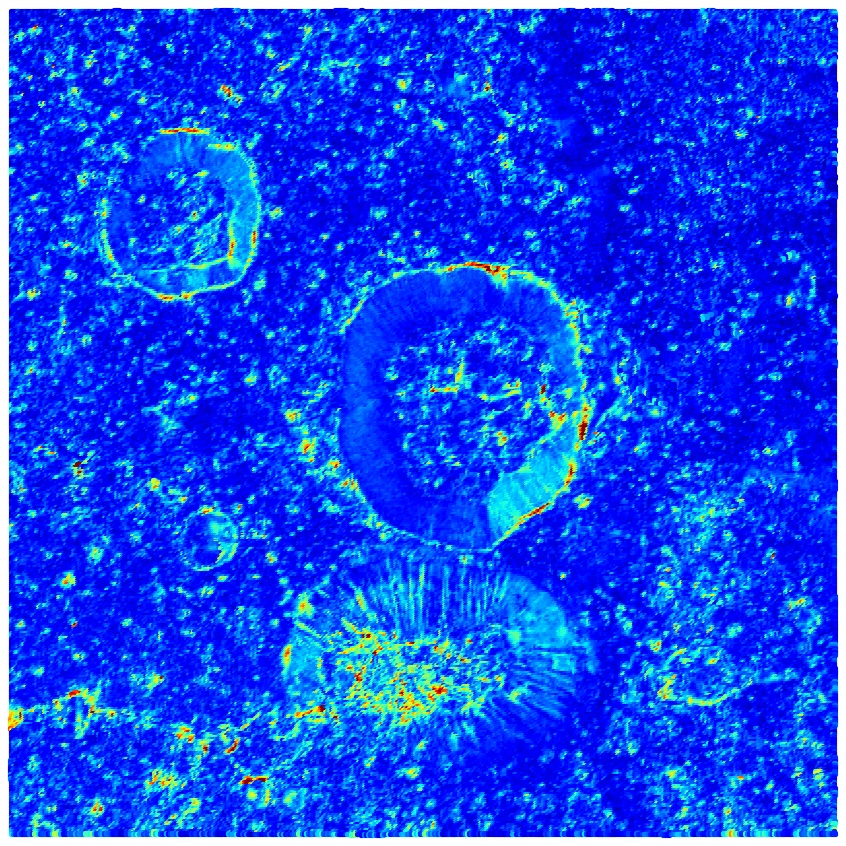}
    \vspace{-18pt}
    \caption*{\scriptsize{\textnormal{Photometric Error, $\delta I_j$ (mean = $0.94$\%)}}}
  \end{subfigure}%
  \begin{subfigure}[t]{0.04725\linewidth}
    \vskip 0pt
    \includegraphics[width=\linewidth]{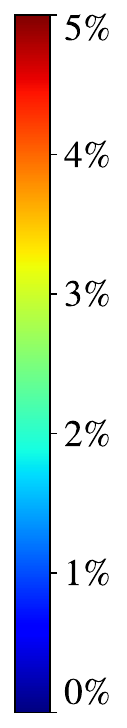}
  \end{subfigure}%
  \hspace{2pt}
  \begin{subfigure}[t]{0.27\linewidth}
    \vskip 0pt
    \includegraphics[width=\linewidth]{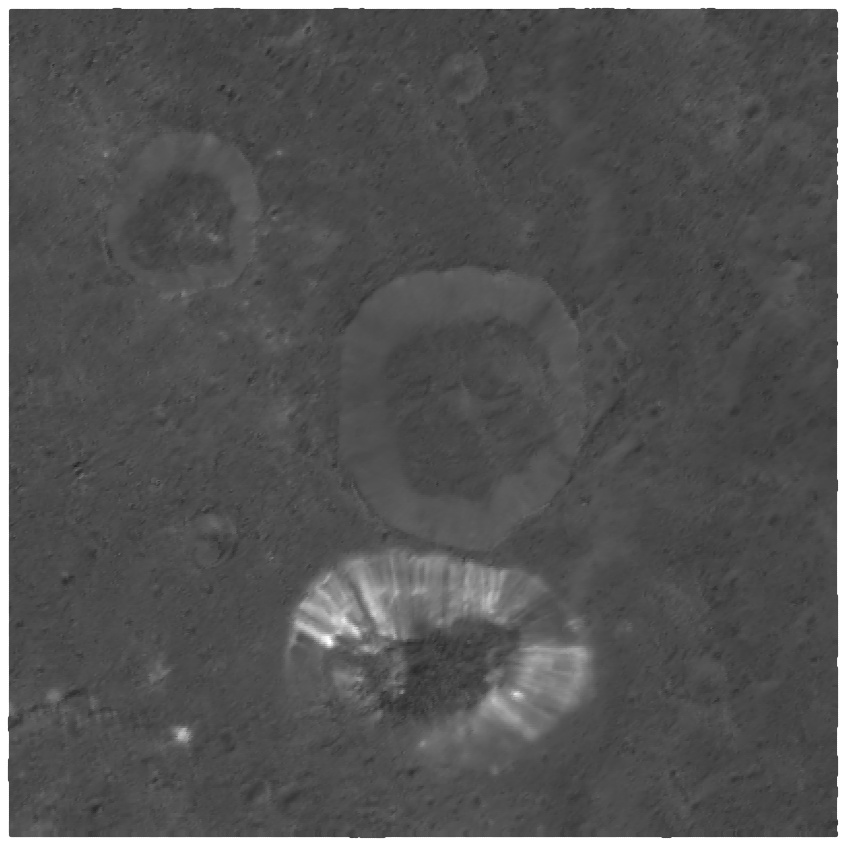}
    \vspace{-18pt}
    \caption*{\scriptsize{\textnormal{Albedos, $a_j$}}}
  \end{subfigure}%
  \begin{subfigure}[t]{0.054\linewidth}
    \vskip 0pt
    \includegraphics[width=\linewidth]{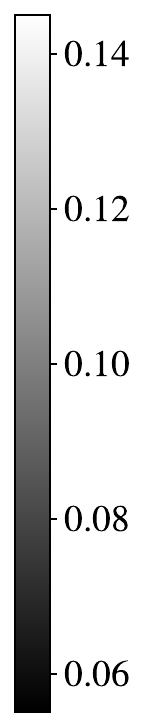}
  \end{subfigure}%
  \hfill
  \begin{subfigure}[t]{0.27\linewidth}
    \vskip 0pt
    \includegraphics[width=\linewidth]{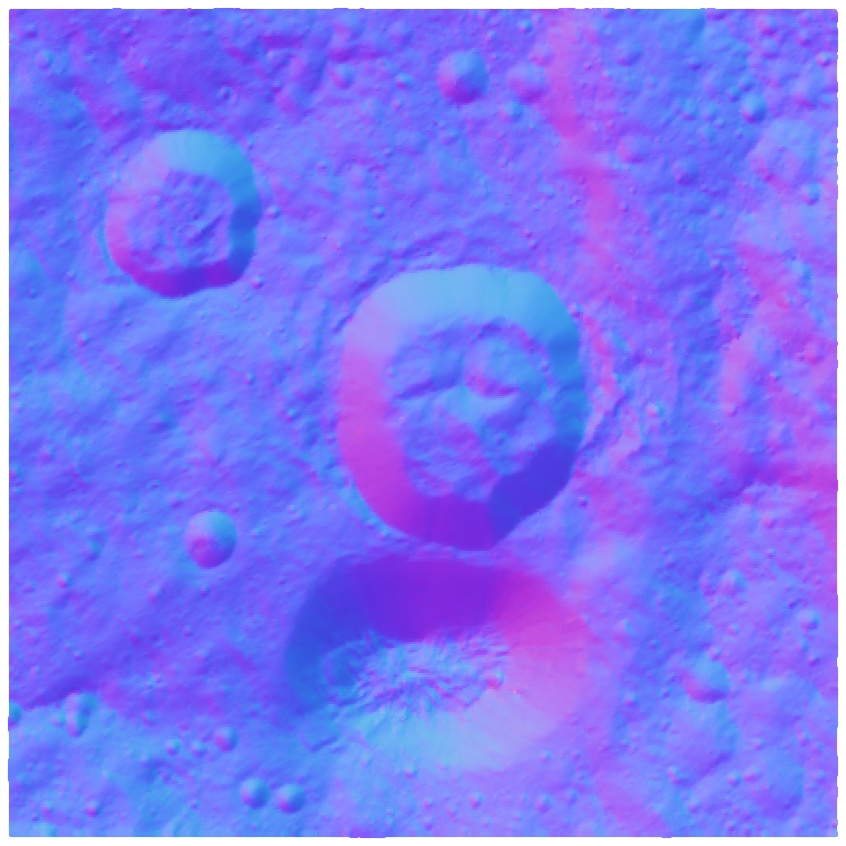}
    \vspace{-18pt}
    \caption*{\scriptsize{\textnormal{Normals, $\vvec{n}_j$}}}
  \end{subfigure}%
  \begin{subfigure}[t]{0.07\linewidth}
    \vskip 0pt
    \includegraphics[width=\linewidth]{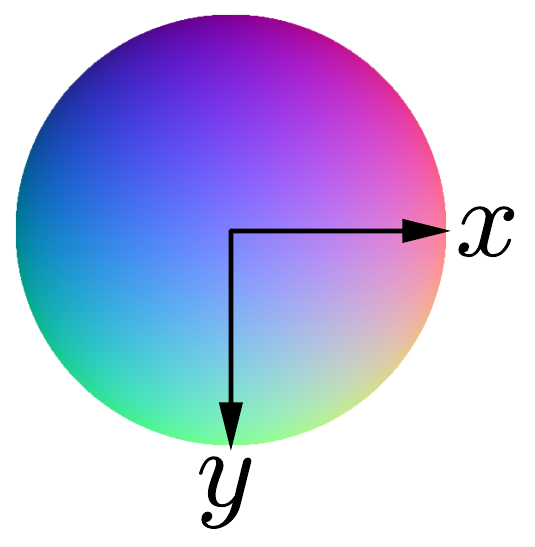}
  \end{subfigure}%
\end{subfigure}\\
\vspace{10pt}
\begin{subfigure}[t]{\linewidth}
\centering
  \begin{subfigure}[t]{0.27\linewidth}
    \vskip 0pt
    \includegraphics[width=\linewidth]{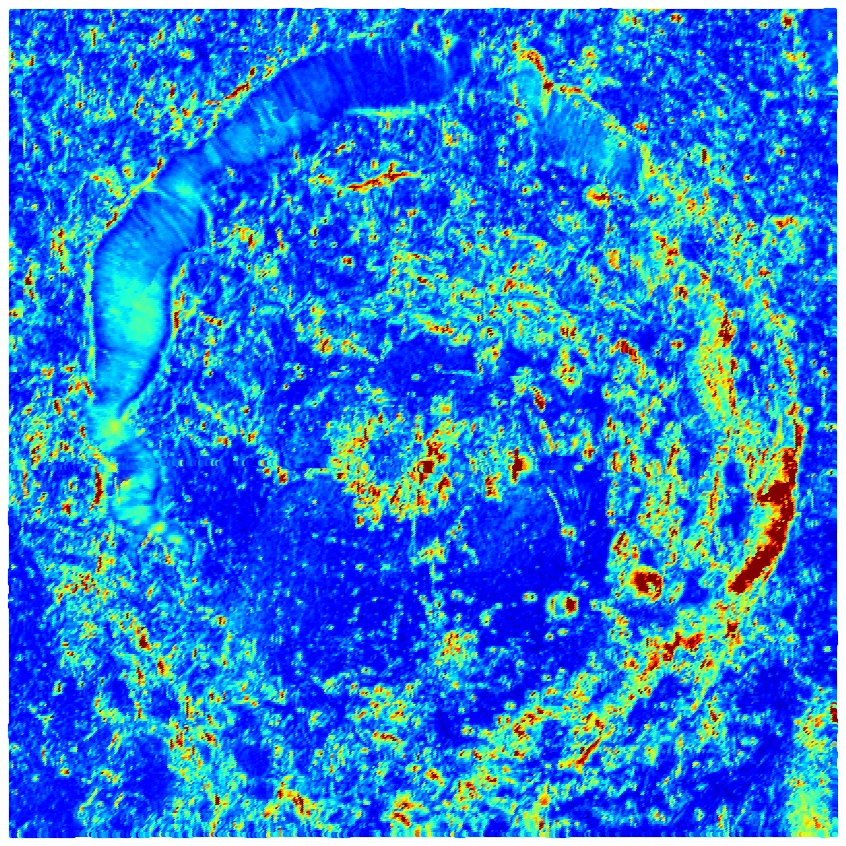}
    \vspace{-18pt}
    \caption*{\scriptsize{\textnormal{Photometric Error, $\delta I_j$ (mean = $1.47$\%)}}}
  \end{subfigure}%
  \begin{subfigure}[t]{0.04725\linewidth}
    \vskip 0pt
    \includegraphics[width=\linewidth]{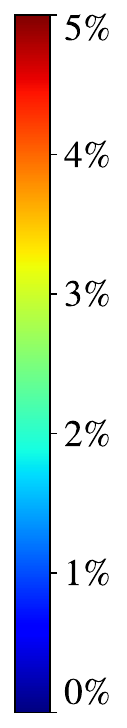}
  \end{subfigure}%
  \hfill
  \begin{subfigure}[t]{0.27\linewidth}
    \vskip 0pt
    \includegraphics[width=\linewidth]{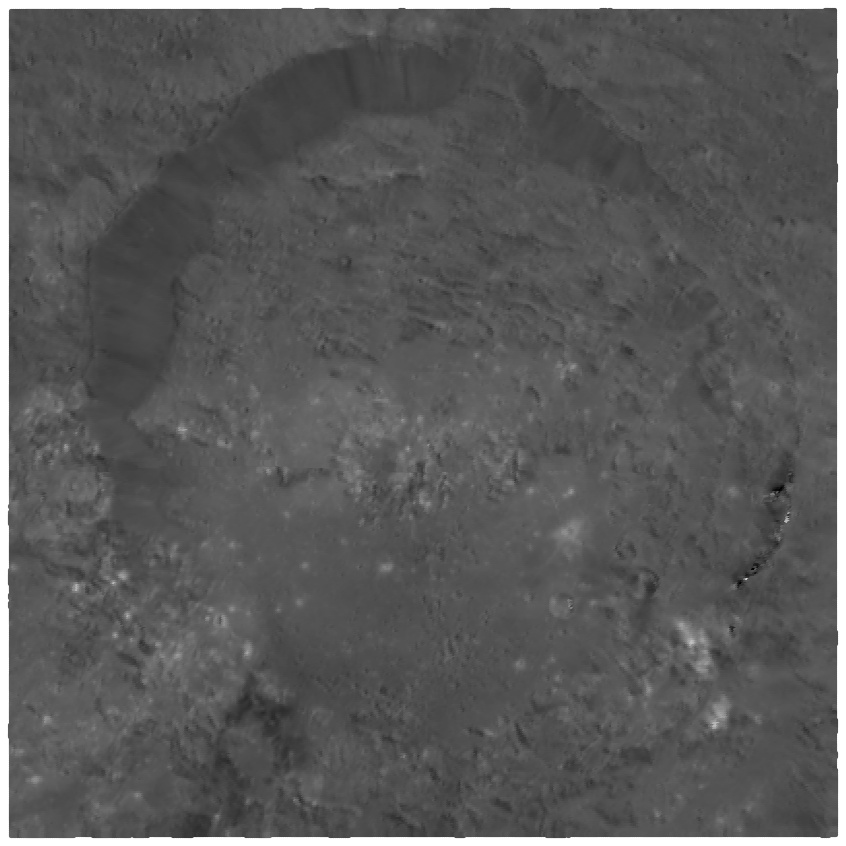}
    \vspace{-18pt}
    \caption*{\scriptsize{\textnormal{Albedos, $a_j$}}}
  \end{subfigure}%
  \begin{subfigure}[t]{0.054\linewidth}
    \vskip 0pt
    \includegraphics[width=\linewidth]{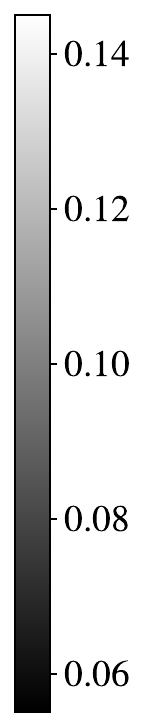}
  \end{subfigure}%
  \hfill
  \begin{subfigure}[t]{0.27\linewidth}
    \vskip 0pt
    \includegraphics[width=\linewidth]{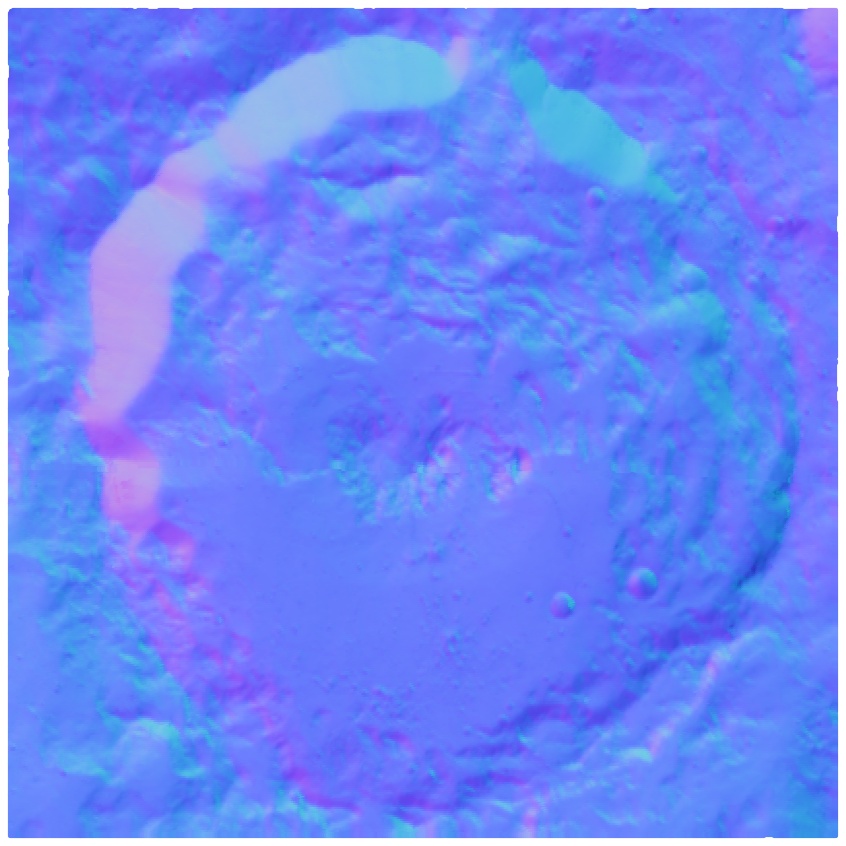}
    \vspace{-18pt}
    \caption*{\scriptsize{\textnormal{Normals, $\vvec{n}_j$}}}
  \end{subfigure}%
  \begin{subfigure}[t]{0.07\linewidth}
    \vskip 0pt
    \includegraphics[width=\linewidth]{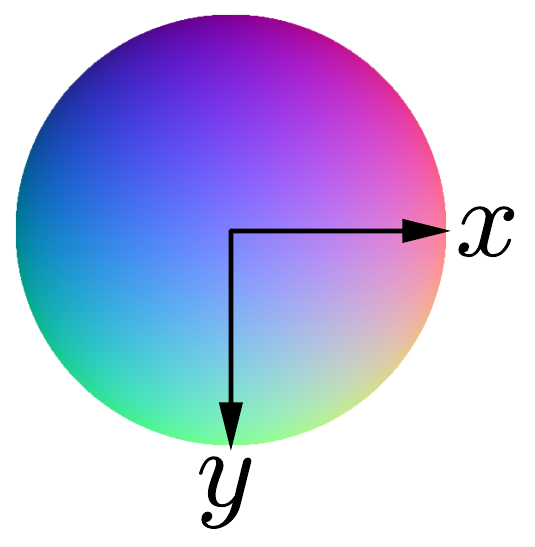}
  \end{subfigure}%
\end{subfigure}\\

%% file: figures/spc-albedos-normals.tex
\begin{subfigure}[t]{\linewidth}
\centering
  \begin{subfigure}[t]{0.27\linewidth}
    \vskip 0pt
    \includegraphics[width=\linewidth]{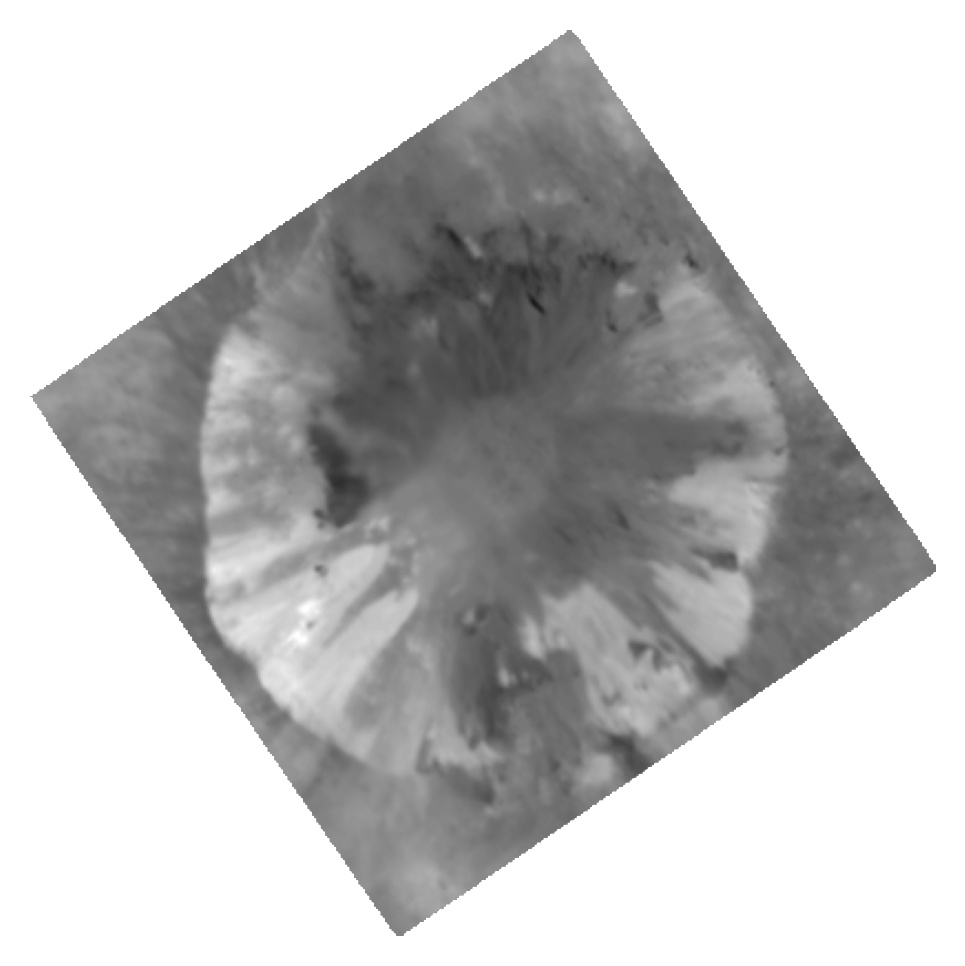}
    \vspace{-18pt}
    \caption*{\scriptsize{\textnormal{Albedos, $a_j$}}}
  \end{subfigure}%
  \begin{subfigure}[t]{0.048\linewidth}
    \vskip 0pt
    \includegraphics[width=\linewidth]{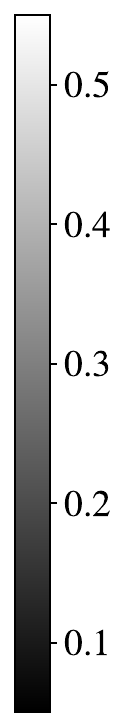}
  \end{subfigure}%
  \hspace{15pt}
  \begin{subfigure}[t]{0.27\linewidth}
    \vskip 0pt
    \includegraphics[width=\linewidth]{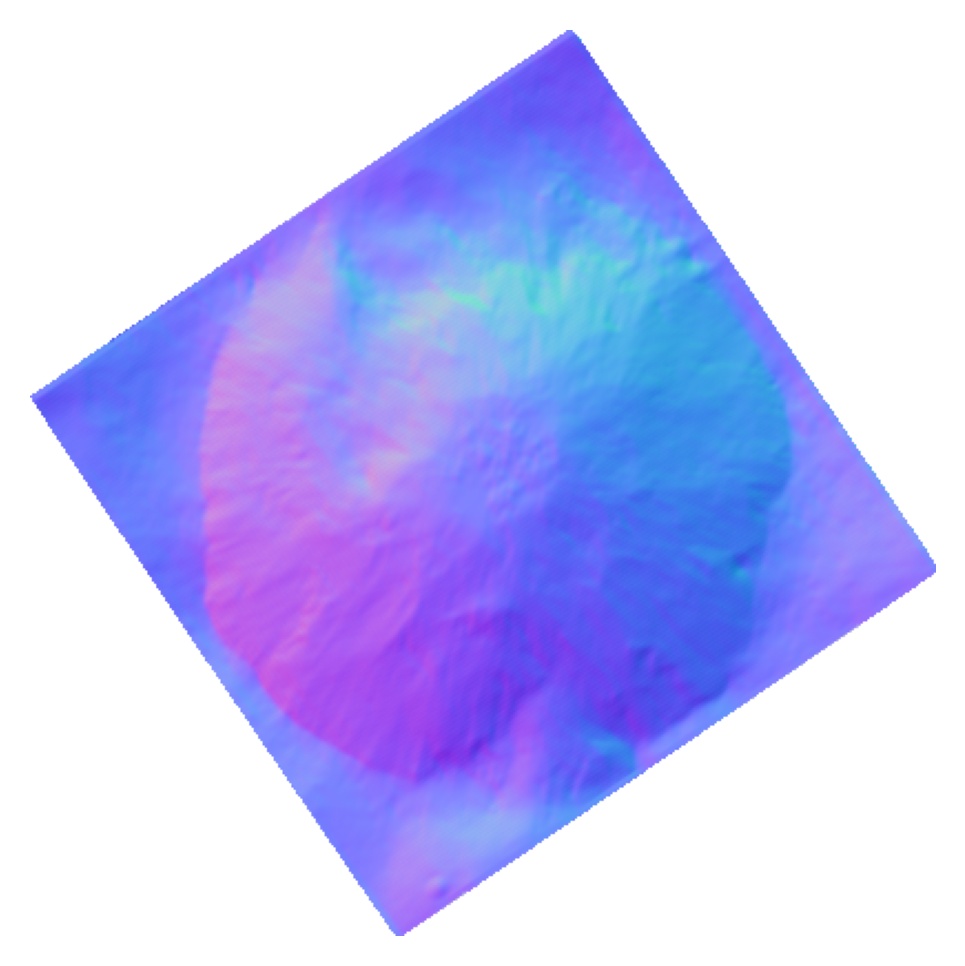}
    \vspace{-18pt}
    \caption*{\scriptsize{\textnormal{Normals, $\vvec{n}_j$}}}
  \end{subfigure}%
  \begin{subfigure}[t]{0.07\linewidth}
    \vskip 0pt
    \includegraphics[width=\linewidth]{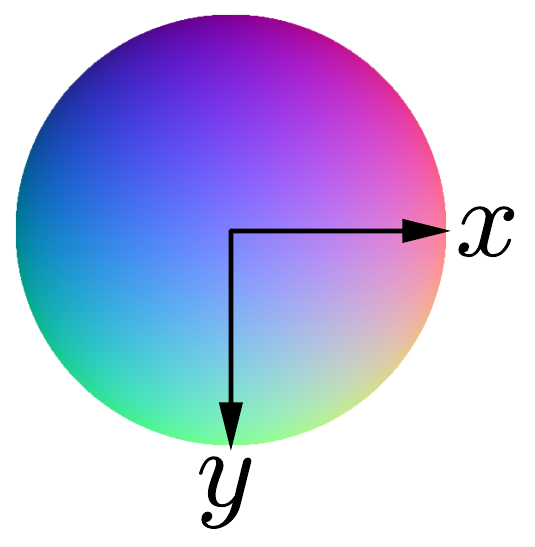}
  \end{subfigure}%
\end{subfigure}\\
\vspace{10pt}
\begin{subfigure}[t]{\linewidth}
\centering
  \begin{subfigure}[t]{0.27\linewidth}
    \vskip 0pt
    \includegraphics[width=\linewidth]{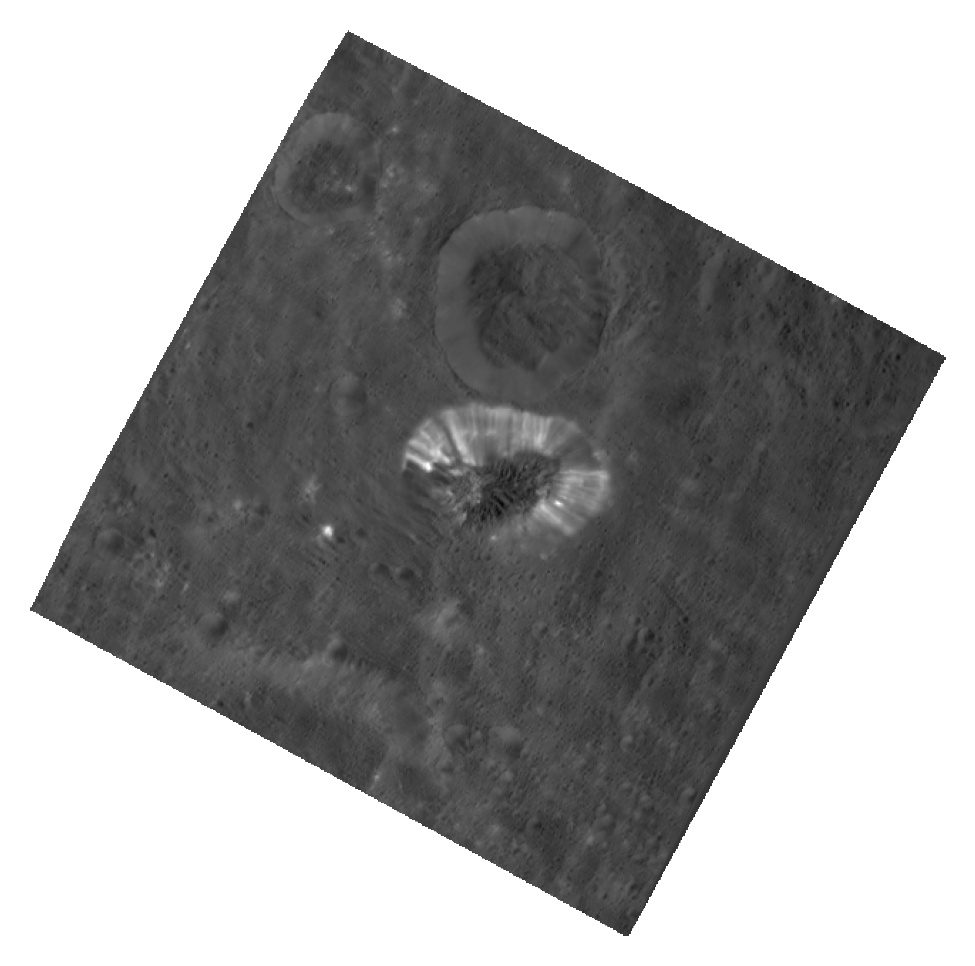}
    \vspace{-18pt}
    \caption*{\scriptsize{\textnormal{Albedos, $a_j$}}}
  \end{subfigure}%
  \begin{subfigure}[t]{0.054\linewidth}
    \vskip 0pt
    \includegraphics[width=\linewidth]{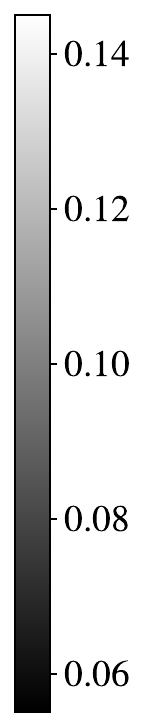}
  \end{subfigure}%
  \hspace{15pt}
  \begin{subfigure}[t]{0.27\linewidth}
    \vskip 0pt
    \includegraphics[width=\linewidth]{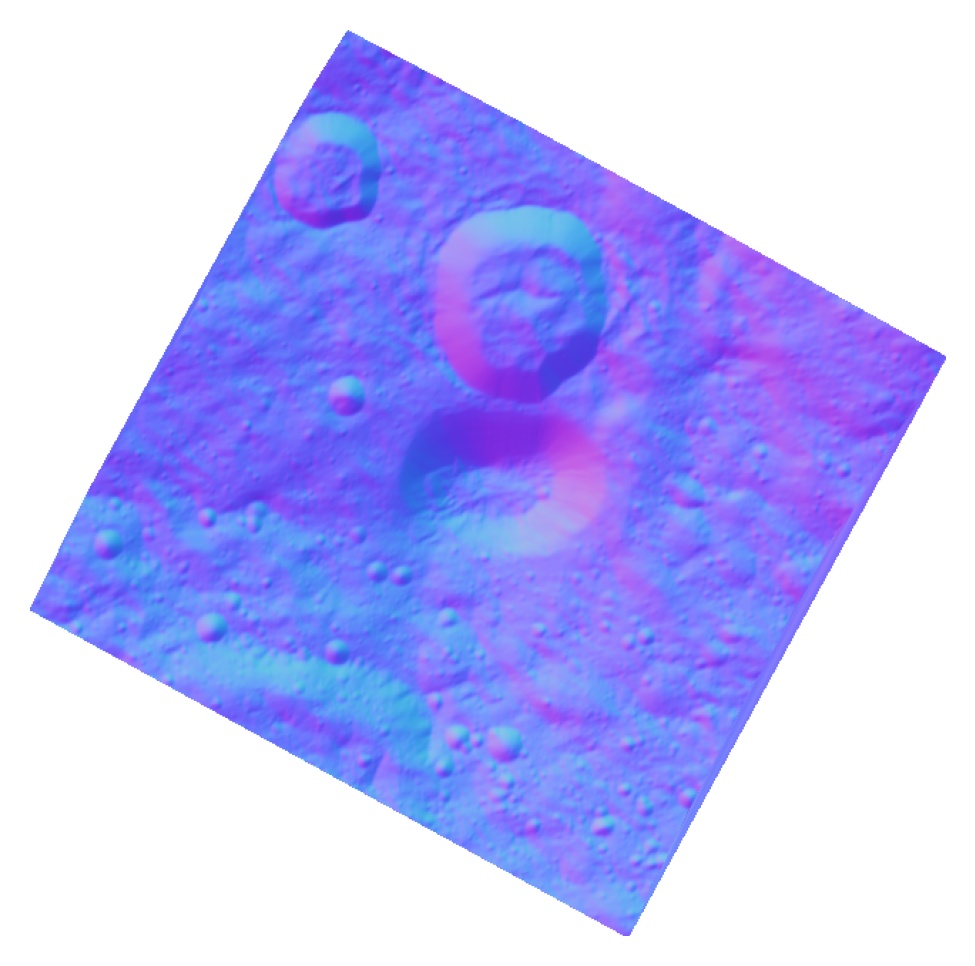}
    \vspace{-18pt}
    \caption*{\scriptsize{\textnormal{Normals, $\vvec{n}_j$}}}
  \end{subfigure}%
  \begin{subfigure}[t]{0.07\linewidth}
    \vskip 0pt
    \includegraphics[width=\linewidth]{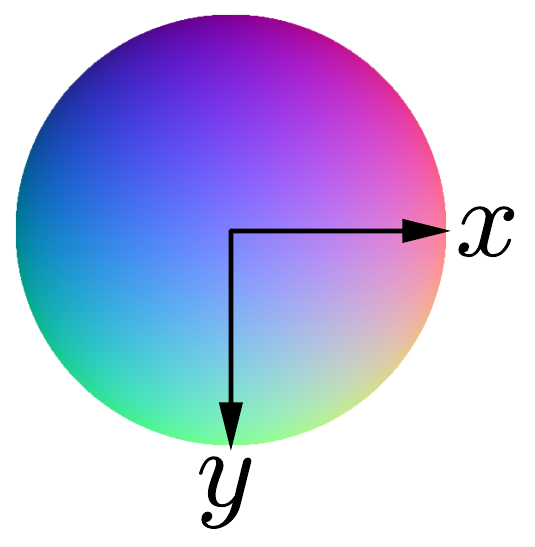}
  \end{subfigure}%
\end{subfigure}\\
\vspace{10pt}
\begin{subfigure}[t]{\linewidth}
\centering
  \begin{subfigure}[t]{0.27\linewidth}
    \vskip 0pt
    \includegraphics[width=\linewidth]{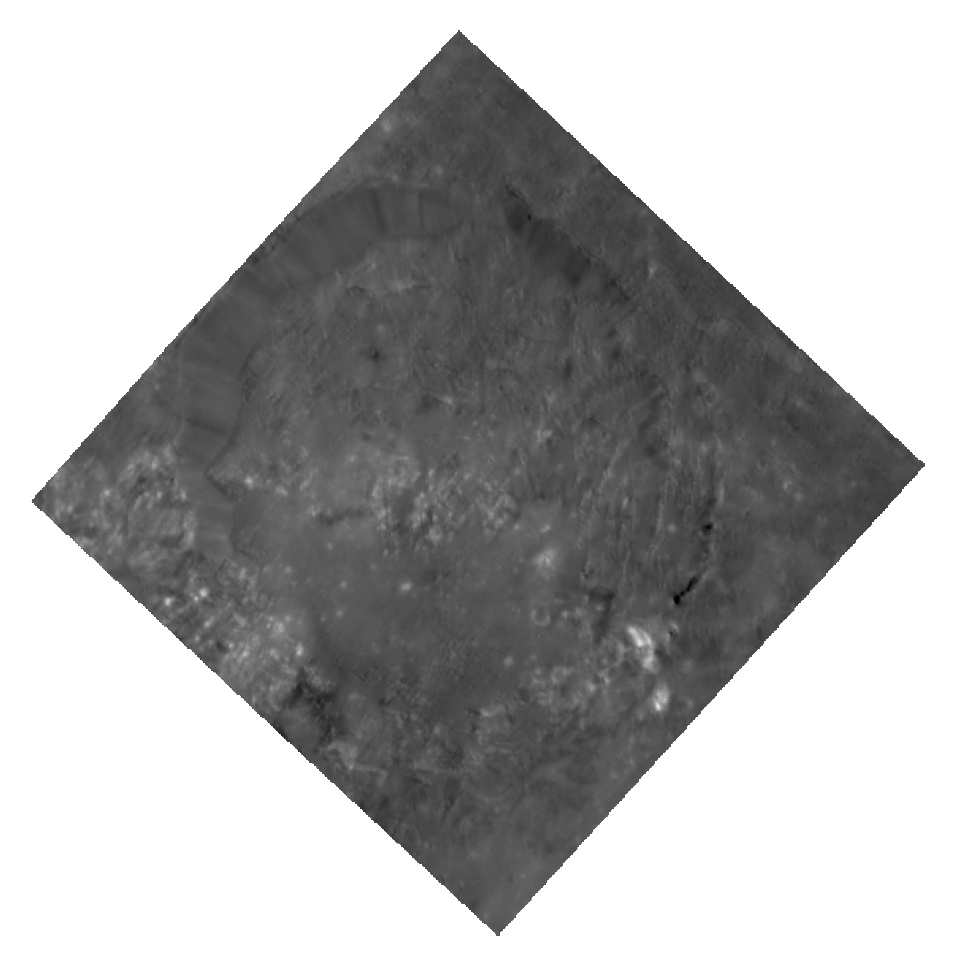}
    \vspace{-18pt}
    \caption*{\scriptsize{\textnormal{Albedos, $a_j$}}}
  \end{subfigure}%
  \begin{subfigure}[t]{0.054\linewidth}
    \vskip 0pt
    \includegraphics[width=\linewidth]{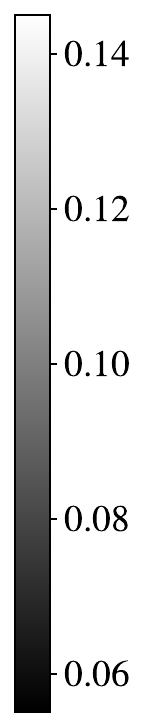}
  \end{subfigure}%
  \hspace{15pt}
  \begin{subfigure}[t]{0.27\linewidth}
    \vskip 0pt
    \includegraphics[width=\linewidth]{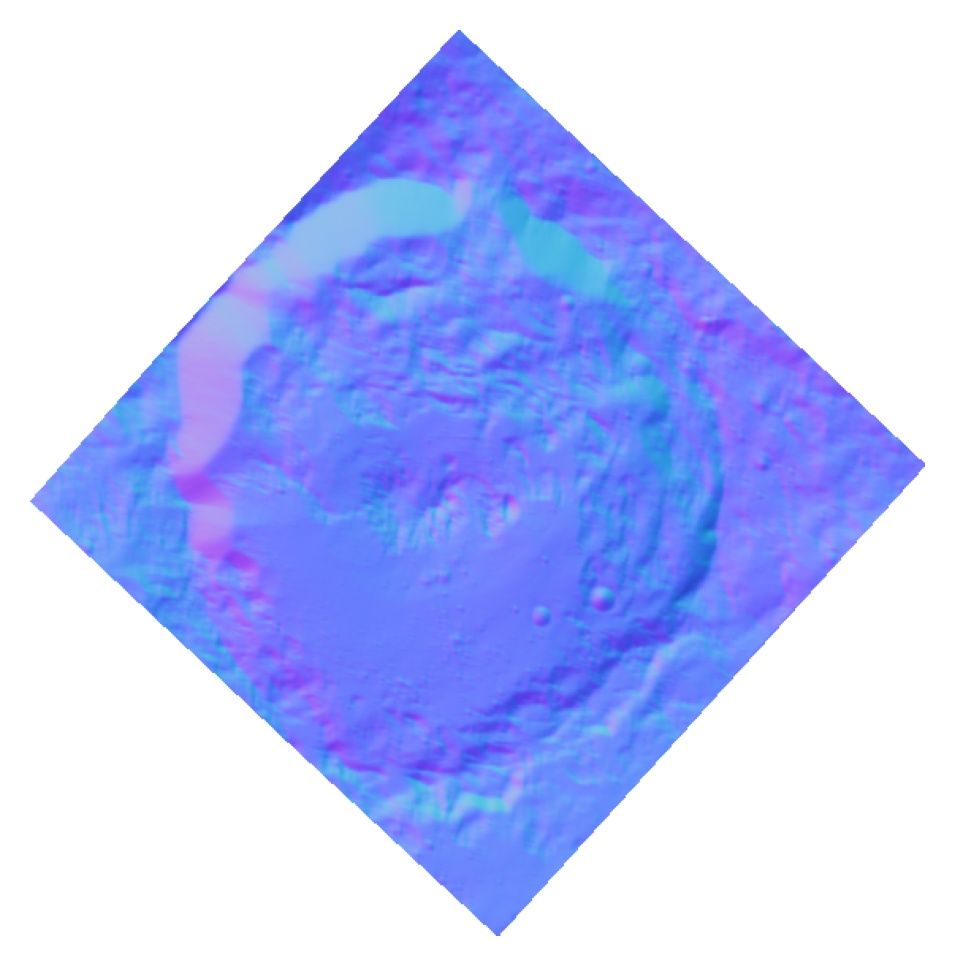}
    \vspace{-18pt}
    \caption*{\scriptsize{\textnormal{Normals, $\vvec{n}_j$}}}
  \end{subfigure}%
  \begin{subfigure}[t]{0.07\linewidth}
    \vskip 0pt
    \includegraphics[width=\linewidth]{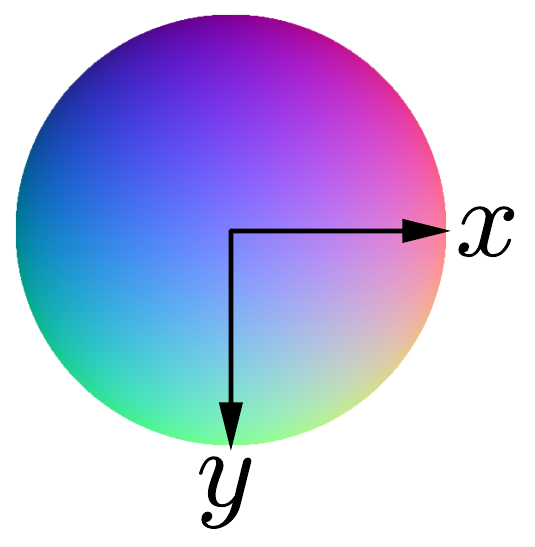}
  \end{subfigure}%
\end{subfigure}\\

%% file: figures/psnr-results-nerf-3dgs.tex
\begin{tabular}{l>{\columncolor[gray]{0.9}}c>{\columncolor[gray]{0.9}}c>{\columncolor[gray]{0.9}}cccc}
\toprule
& \multicolumn{3}{c}{\cellcolor[gray]{0.9} Train} & \multicolumn{3}{c}{Test} \\
\cmidrule(lr){2-4} \cmidrule(lr){5-7}
                 & PhoMo & iNGP & 3DGS & PhoMo & iNGP & 3DGS \\ 
\midrule
Ahuna Mons       & \underline{38.89} & 35.64 & \textbf{45.37} & \textbf{35.70} & 30.82 & \underline{32.79} \\
Ikapati          & \underline{35.64} & 30.83 & \textbf{41.43} & \textbf{35.09} & 28.72 & \underline{31.79} \\
Cornelia         & \underline{41.01} & 38.94 & \textbf{45.79} & \textbf{39.59} & 31.20 & \underline{35.90} \\
\midrule
Average          & \underline{38.51} & 35.14 & \textbf{44.20} & \textbf{36.79} & 30.25 & \underline{33.49} \\
\bottomrule
\end{tabular}

%% file: figures/qual-compare-phomo-nerf-3dgs.tex
\centering
\setlength{\extrarowheight}{-10pt}
\setlength{\tabcolsep}{1pt} 
\begin{tabular}{cp{2.9cm}p{.3cm}p{2.9cm}p{2.9cm}p{.3cm}p{2.9cm}p{2.9cm}p{.3cm}p{2.9cm}p{2.9cm}p{.6cm}}
    &&& \multicolumn{2}{l}{\textcolor{black}{\normalsize{PhoMo}}} && \multicolumn{2}{l}{\textcolor{black}{\normalsize{instant-NGP}}} && \multicolumn{2}{l}{\textcolor{black}{\normalsize{3DGS}}} \\
    &
    \includegraphics[width=\linewidth]{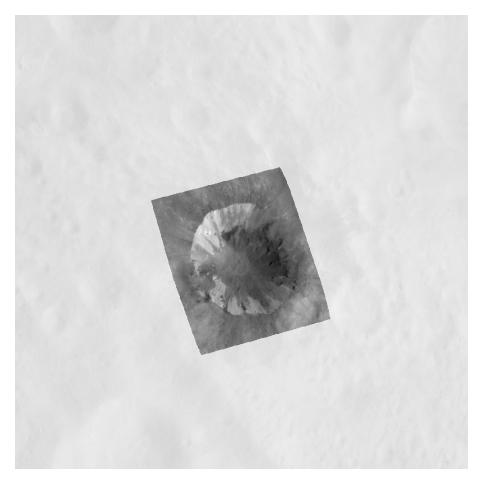} &&
    \includegraphics[width=\linewidth]{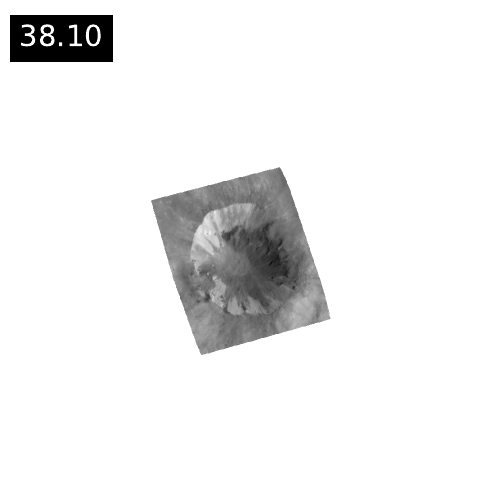} &
    \includegraphics[width=\linewidth]{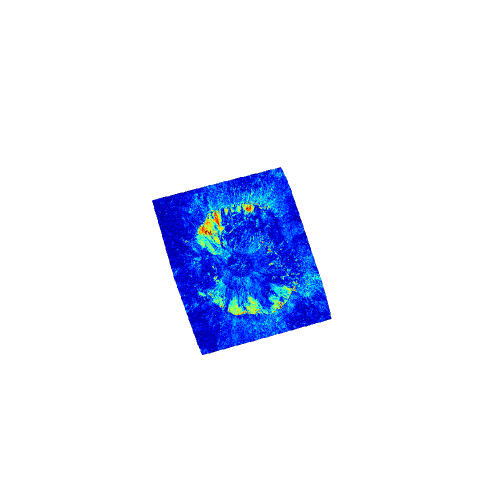} &&
    \includegraphics[width=\linewidth]{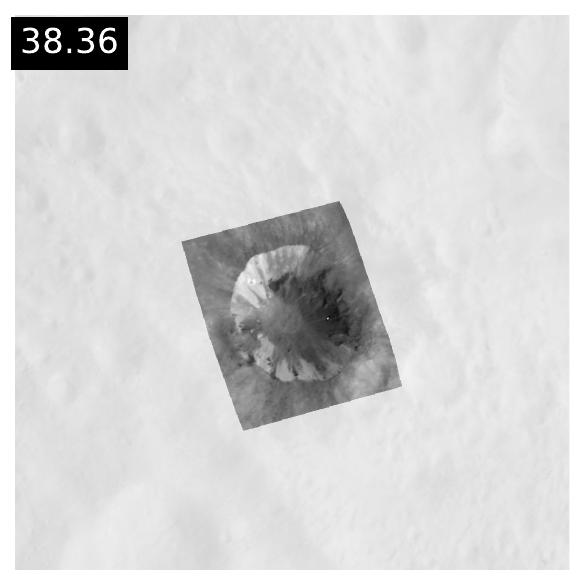} &
    \includegraphics[width=\linewidth]{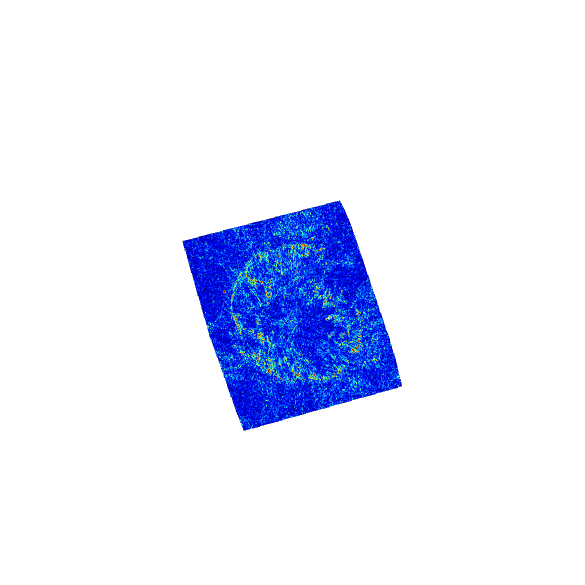} &&
    \includegraphics[width=\linewidth]{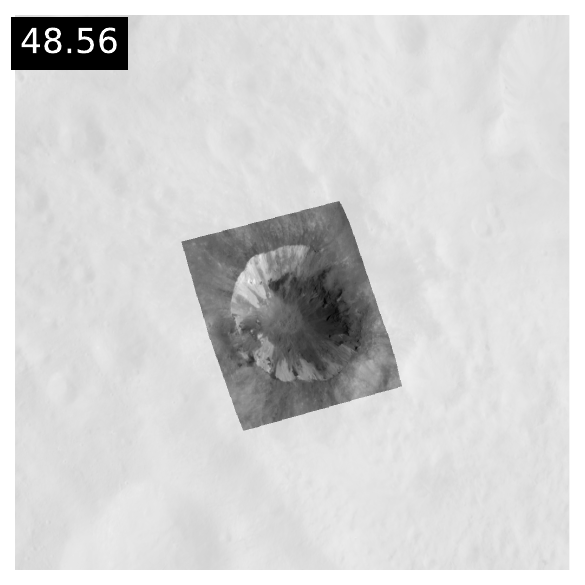} &
    \includegraphics[width=\linewidth]{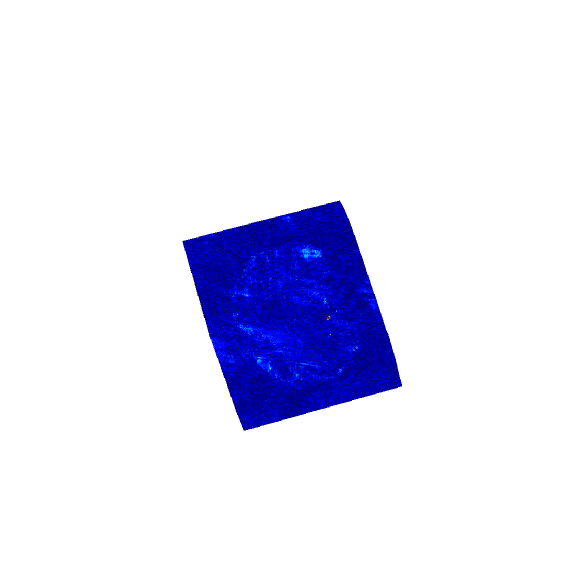} &
    \includegraphics[width=\linewidth]{figures/phomo_results/ahunamons/lunar_lambert/images_rendered/err-cb.pdf} \\
    \parbox[t]{2mm}{\rotatebox[origin=l]{90}{\textcolor{gray}{\normalsize{Cornelia}}}} &
    \includegraphics[width=\linewidth]{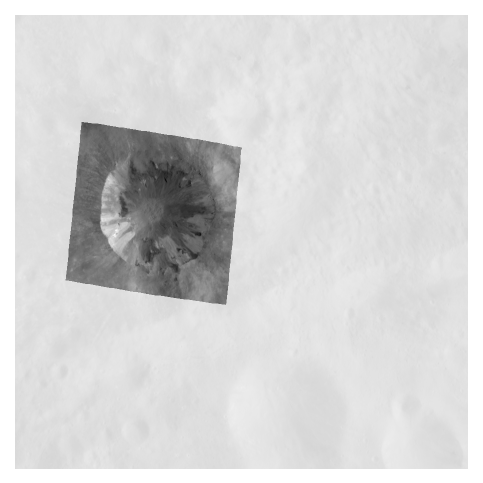} &&
    \includegraphics[width=\linewidth]{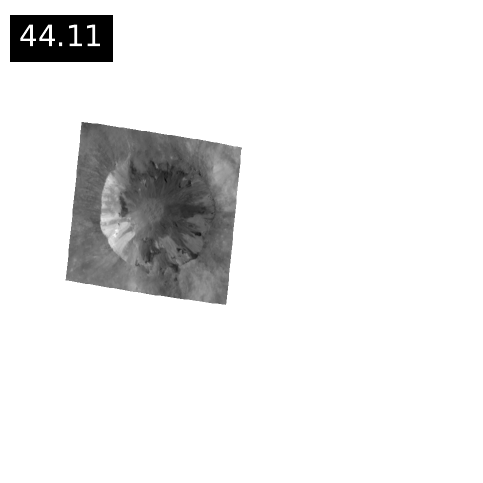} &
    \includegraphics[width=\linewidth]{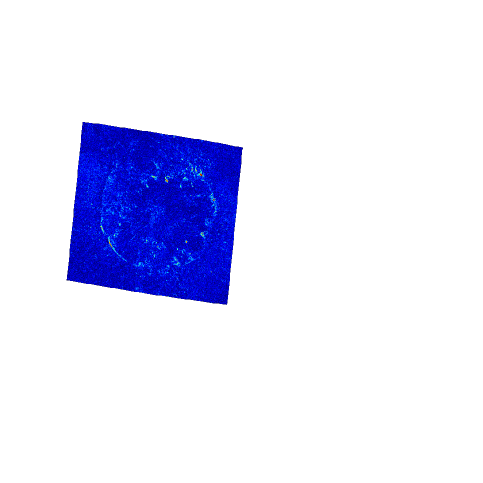} &&
    \includegraphics[width=\linewidth]{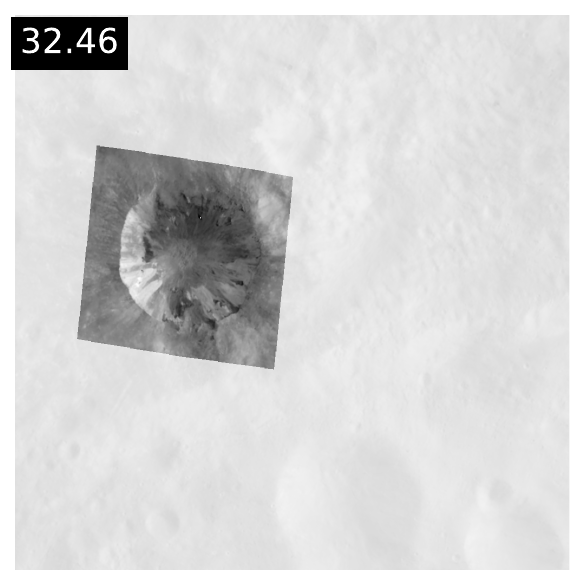} &
    \includegraphics[width=\linewidth]{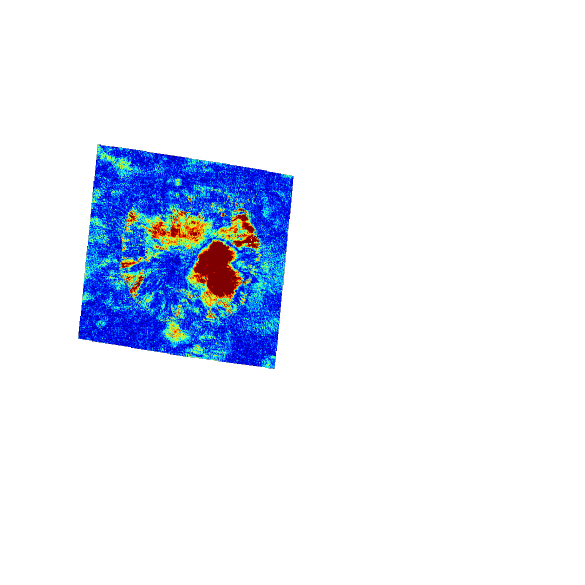} &&
    \includegraphics[width=\linewidth]{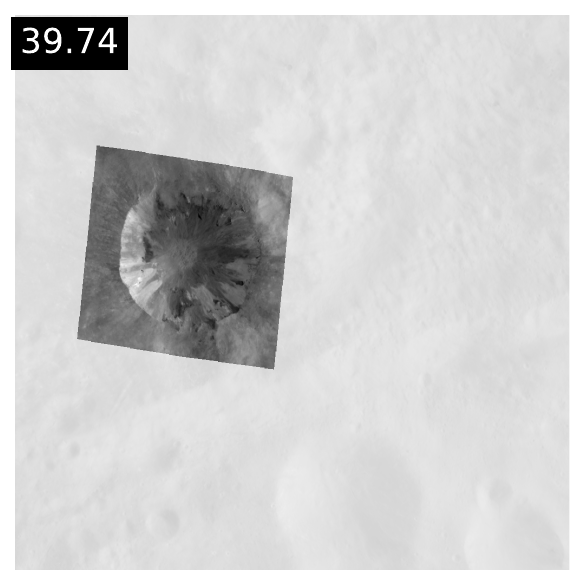} &
    \includegraphics[width=\linewidth]{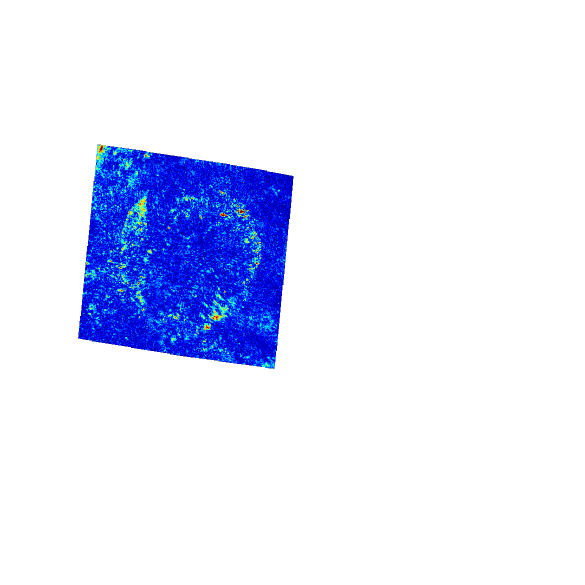} &
    \includegraphics[width=\linewidth]{figures/phomo_results/ahunamons/lunar_lambert/images_rendered/err-cb.pdf} \\
    & & & & \\ 
    & & & & \\ 
    &
    \includegraphics[width=\linewidth]{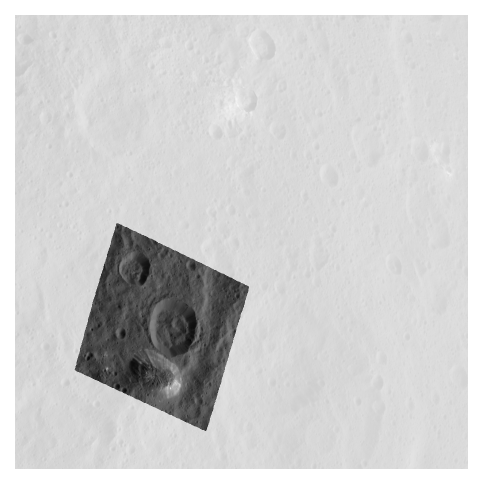} &&
    \includegraphics[width=\linewidth]{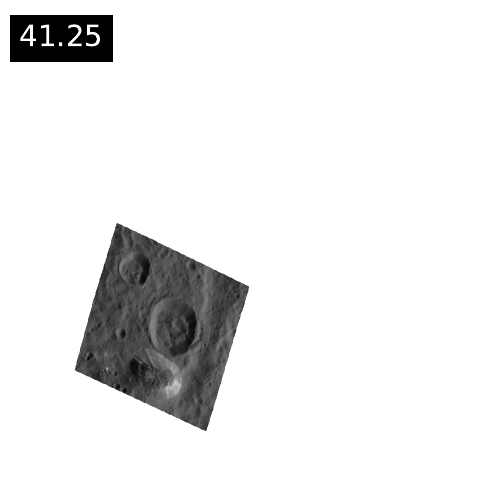} &
    \includegraphics[width=\linewidth]{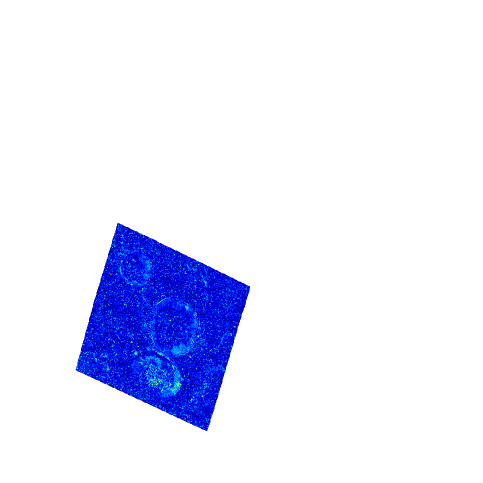} &&
    \includegraphics[width=\linewidth]{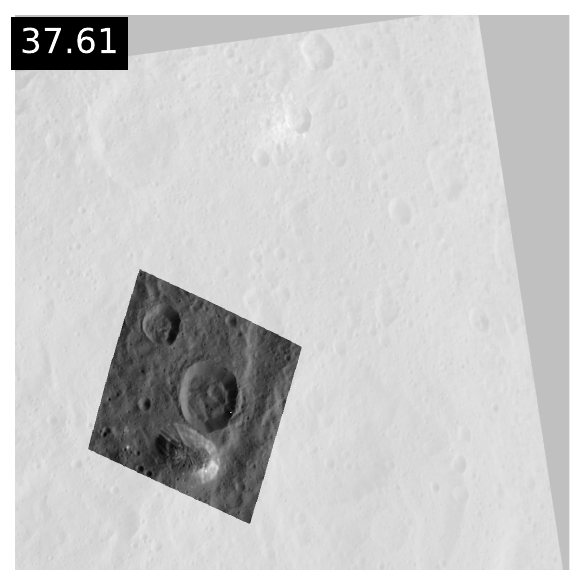} &
    \includegraphics[width=\linewidth]{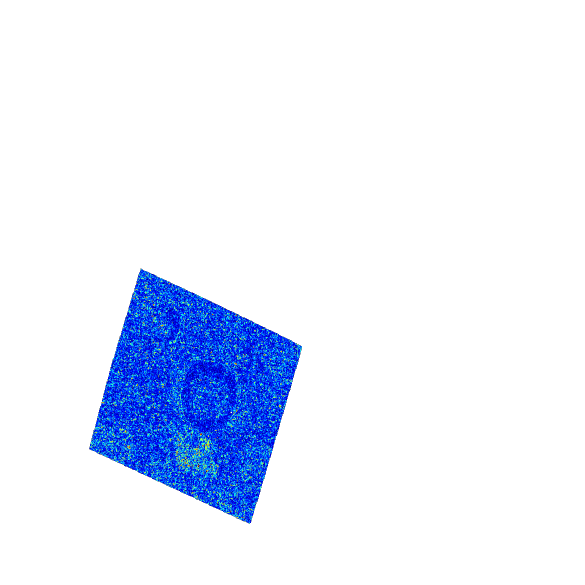} &&
    \includegraphics[width=\linewidth]{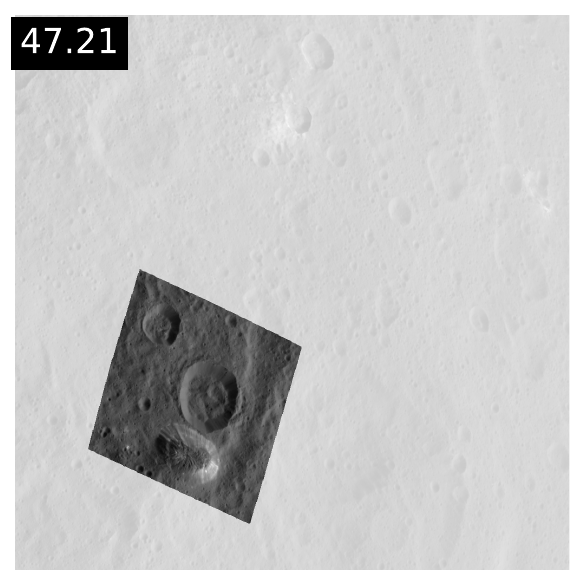} &
    \includegraphics[width=\linewidth]{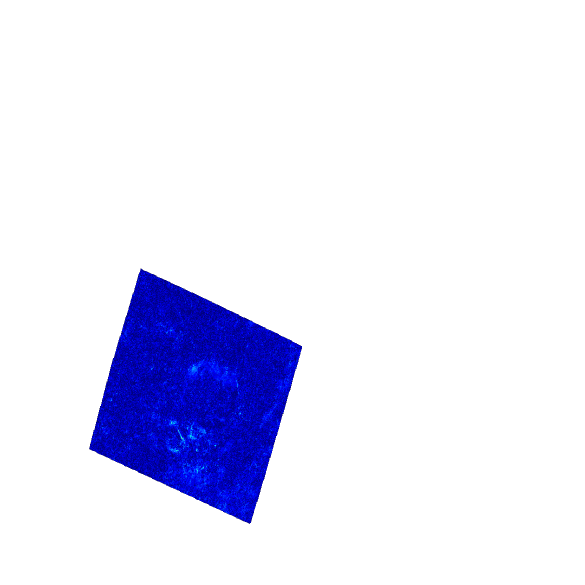} &
    \includegraphics[width=\linewidth]{figures/phomo_results/ahunamons/lunar_lambert/images_rendered/err-cb.pdf} \\
    \parbox[t]{2mm}{\rotatebox[origin=l]{90}{\textcolor{gray}{\normalsize{Ahuna Mons}}}} &
    \includegraphics[width=\linewidth]{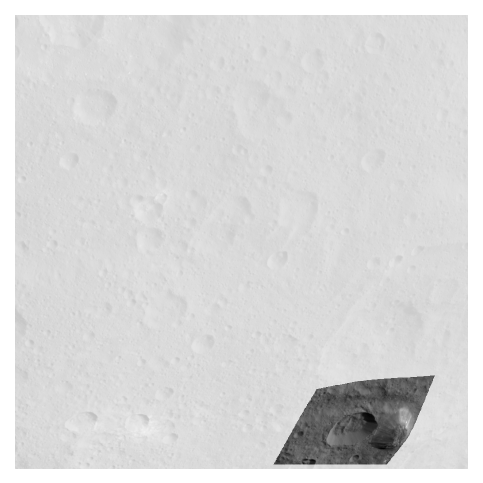} &&
    \includegraphics[width=\linewidth]{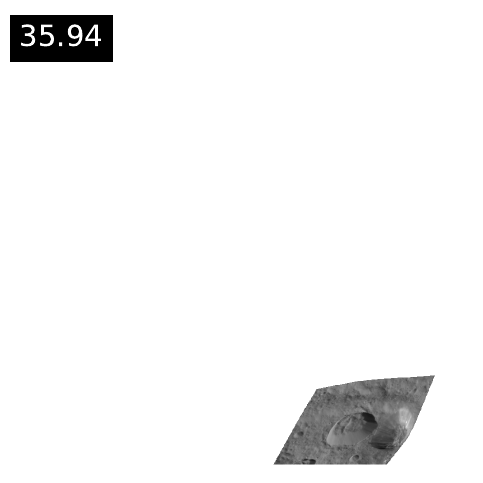} &
    \includegraphics[width=\linewidth]{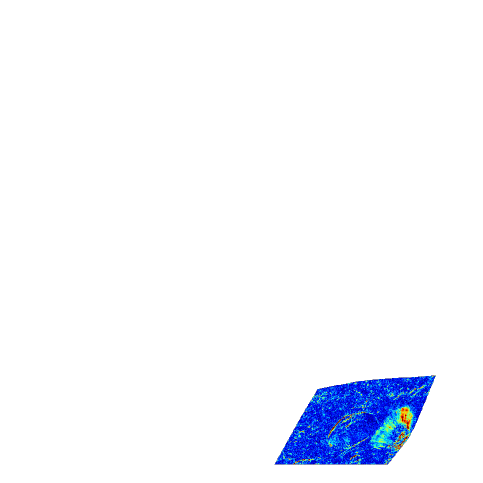} &&
    \includegraphics[width=\linewidth]{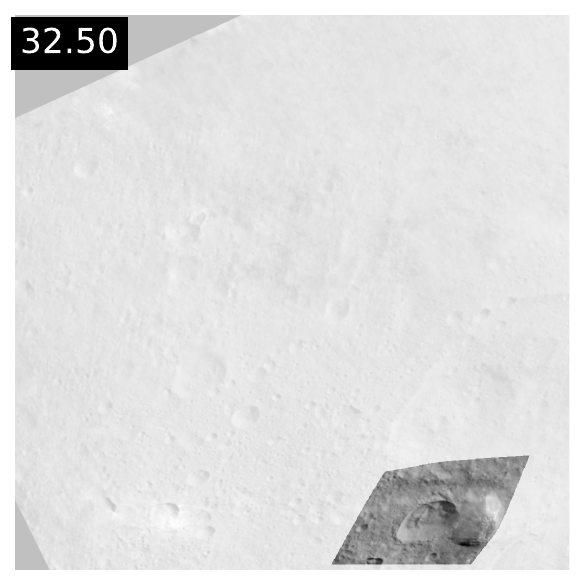} &
    \includegraphics[width=\linewidth]{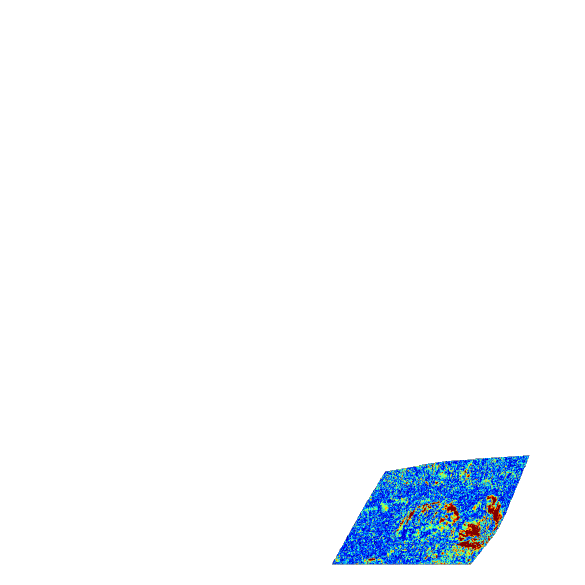} &&
    \includegraphics[width=\linewidth]{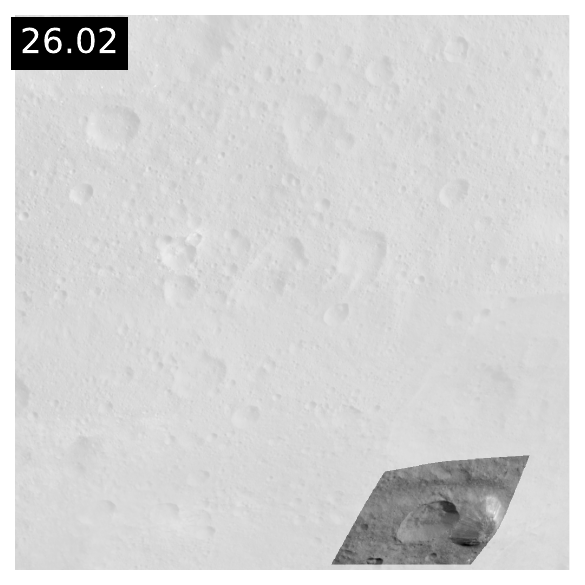} &
    \includegraphics[width=\linewidth]{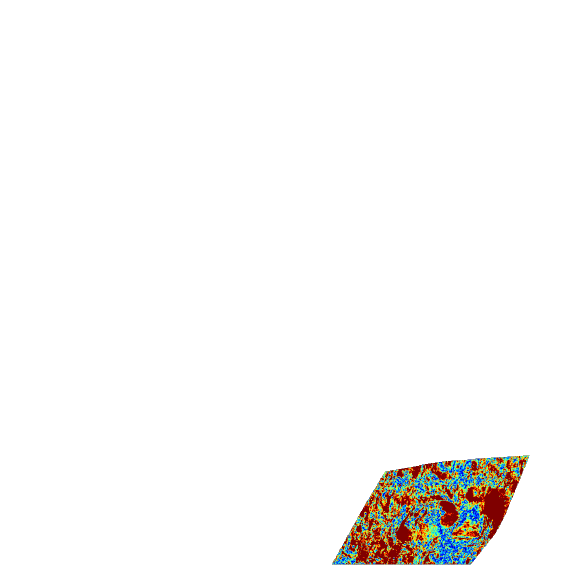} &
    \includegraphics[width=\linewidth]{figures/phomo_results/ahunamons/lunar_lambert/images_rendered/err-cb.pdf} \\
    & & & & \\ 
    & & & & \\ 
    &
    \includegraphics[width=\linewidth]{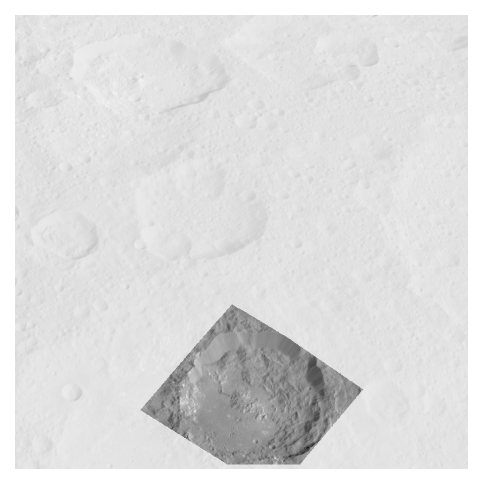} &&
    \includegraphics[width=\linewidth]{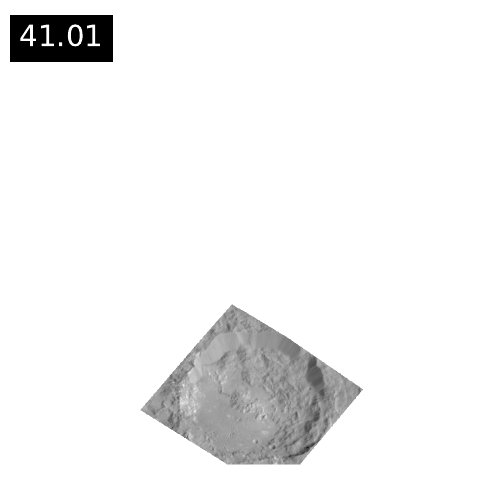} &
    \includegraphics[width=\linewidth]{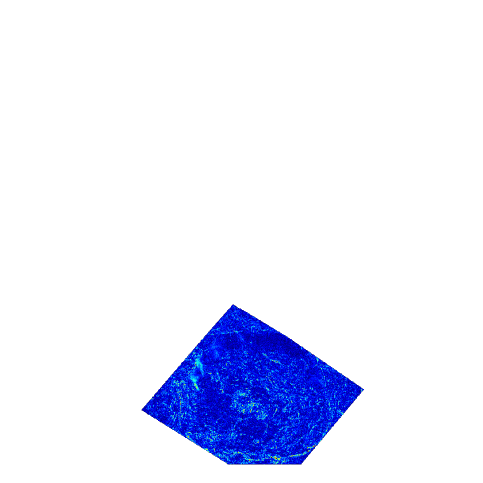} &&
    \includegraphics[width=\linewidth]{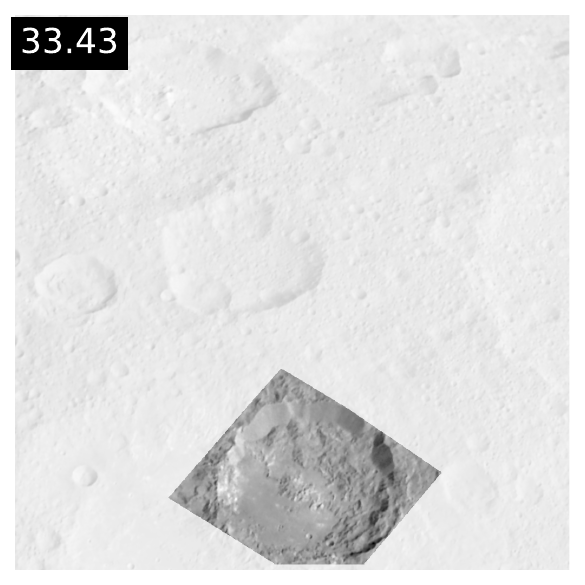} &
    \includegraphics[width=\linewidth]{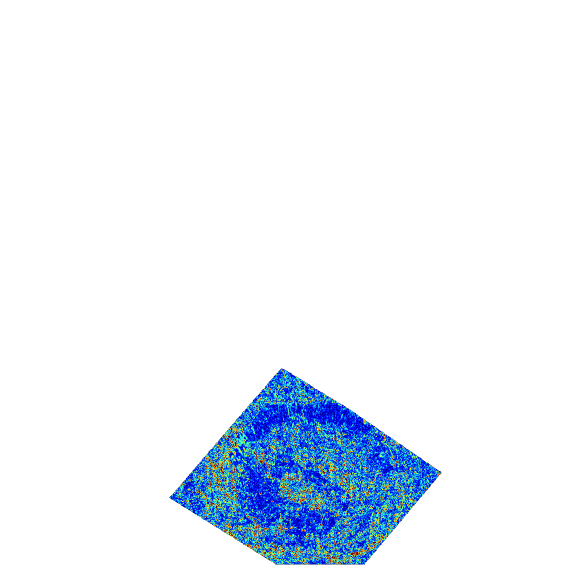} &&
    \includegraphics[width=\linewidth]{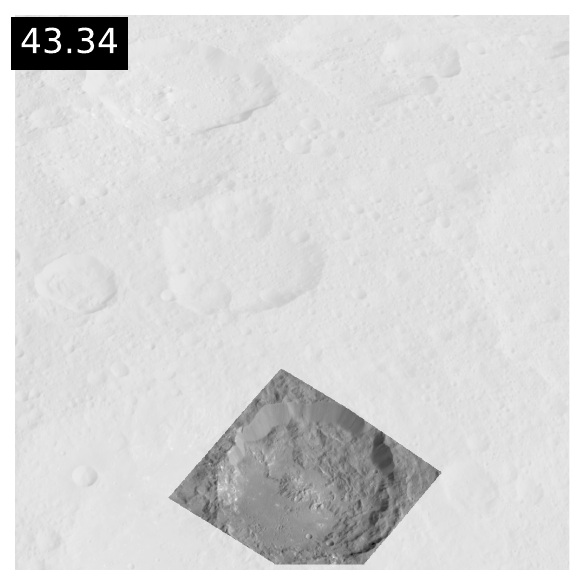} &
    \includegraphics[width=\linewidth]{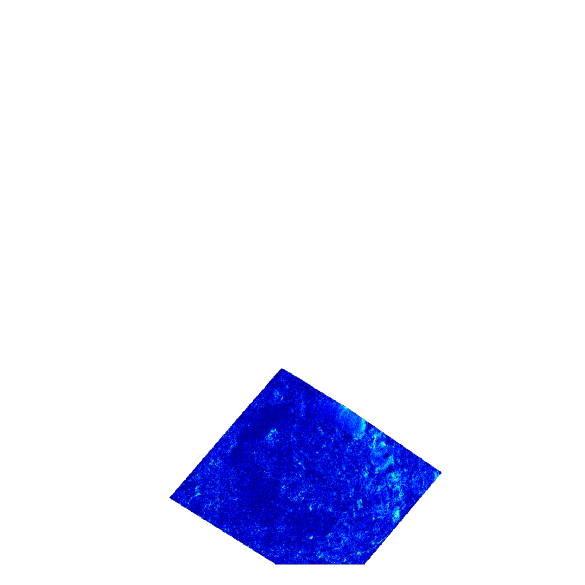} &
    \includegraphics[width=\linewidth]{figures/phomo_results/ahunamons/lunar_lambert/images_rendered/err-cb.pdf} \\
    \parbox[t]{2mm}{\rotatebox[origin=l]{90}{\textcolor{gray}{\normalsize{Ikapati}}}} &
    \includegraphics[width=\linewidth]{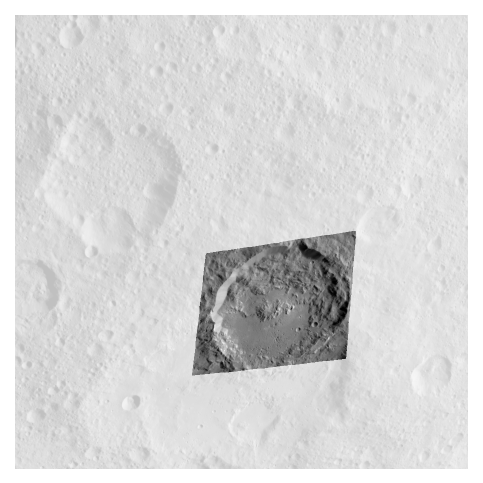} &&
    \includegraphics[width=\linewidth]{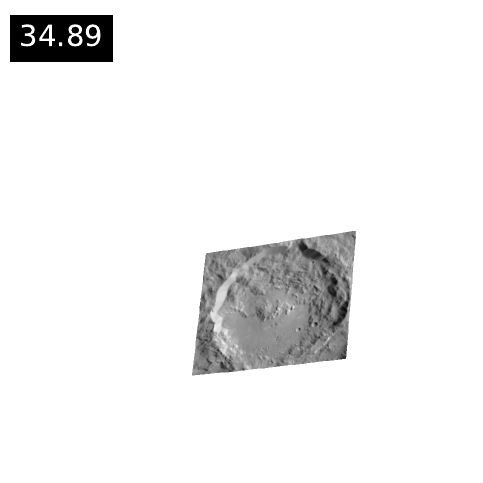} &
    \includegraphics[width=\linewidth]{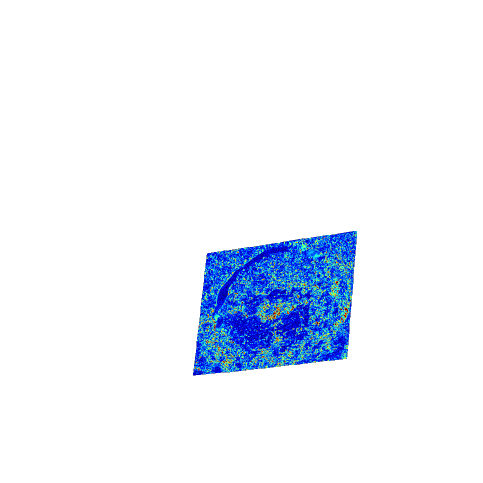} &&
    \includegraphics[width=\linewidth]{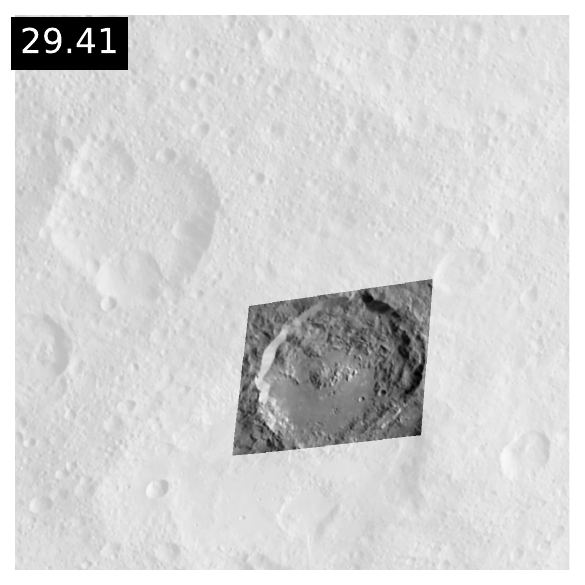} &
    \includegraphics[width=\linewidth]{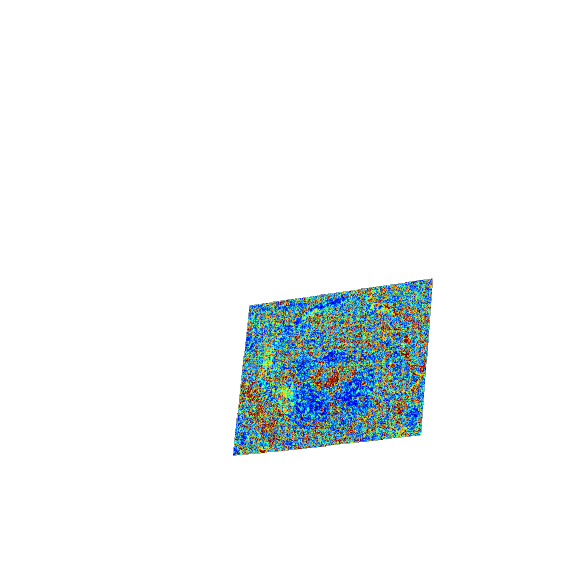} &&
    \includegraphics[width=\linewidth]{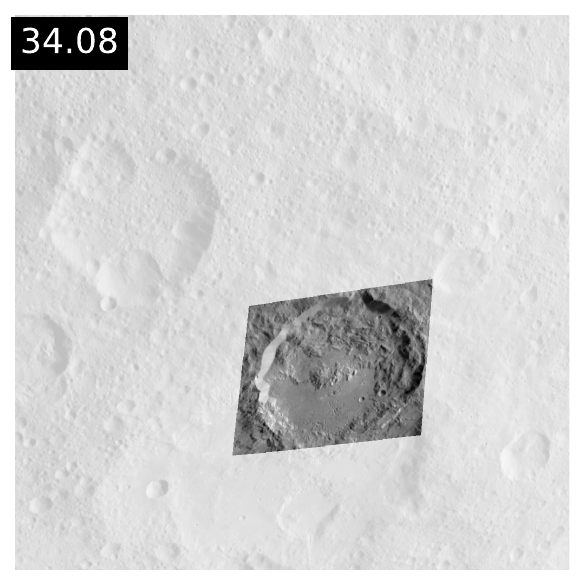} &
    \includegraphics[width=\linewidth]{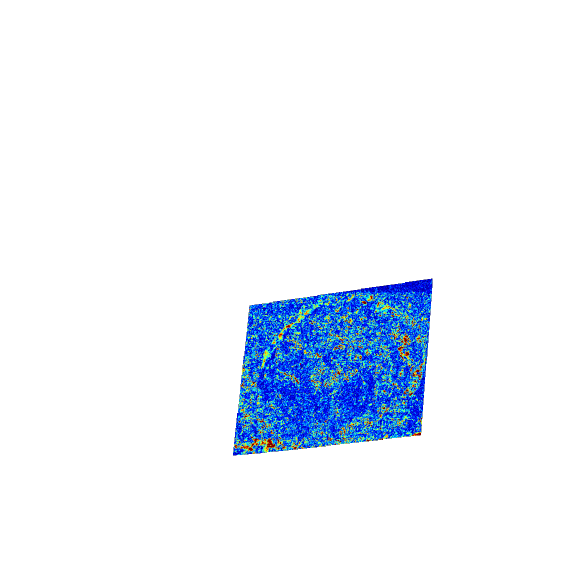} &
    \includegraphics[width=\linewidth]{figures/phomo_results/ahunamons/lunar_lambert/images_rendered/err-cb.pdf} \\
    & \centering\normalsize{Actual} && \centering\normalsize{Render} & \centering\normalsize{Abs. Error} && \centering\normalsize{Render} & \centering\normalsize{Abs. Error} && \centering\normalsize{Render} & \centering\normalsize{Abs. Error} \\
\end{tabular}

%% file: text/appendix.tex
\section{Measurement Function Partial Derivatives} \label{sec:partials}

Below we derive the measurement function partial derivatives for each of the factors proposed in this work.


\subsection{Photoclinometry Factors}

In this section, we explicitly derive the partial derivatives for the measurement function $I(T, \mathbf{s}, \vvec{\ell}, \vvec{n}, a)$, with the Lunar-Lambert model (\ref{eq:ref-mcewen}), for the factor $f_\mathrm{Ph}$ (\ref{eq:fSPC}).
Recall that $\vvec{e} = \vvec{r}_{\oC\oB} - \vvec{\ell}$. 
Then,
\begin{equation}
    \frac{\partial \vvec{e}}{\partial \vvec{\zeta}} = \begin{bmatrix}
        \mathrm{I}_{3\times 3} & R
    \end{bmatrix}, \qquad
    \frac{\partial \vvec{e}}{\partial \vvec{\ell}} = -\mathrm{I}_{3\times 3},
\end{equation}
where we have applied the retraction $r_{\oC\oB} \leftarrow r_{\oC\oB} + R\vvec{\tau}$. 
Let $\vvec{d} = \vvec{e}/\|\vvec{e}\|$, $f = \vvec{s}^\top\vvec{n}$, $w = \vvec{d}^\top\vvec{n}$, and $h = \vvec{s}^\top\vvec{d}$. Then 
\begin{equation}
    \begin{split}
        \frac{\partial f}{\partial \vvec{\xi}_{\vvec{s}}} &= \vvec{n}^\top B_{\vvec{s}},  \\
        \frac{\partial w}{\partial \vvec{\xi}_{\vvec{d}}} &= \vvec{n}^\top B_{\vvec{d}},  \\
        \frac{\partial h}{\partial \vvec{\xi}_{\vvec{s}}} &= \vvec{d}^\top B_{\vvec{s}},  \\
        \frac{\partial \vvec{\xi}_{\vvec{d}}}{\partial \vvec{e}}& = B_{\vvec{d}}^\top D_\mathrm{norm}(\vvec{e}),
    \end{split}
    \qquad
    \begin{split}
        \frac{\partial f}{\partial \vvec{\xi}_{\vvec{n}}} &= \vvec{s}^\top B_{\vvec{n}}, \\
        \frac{\partial w}{\partial \vvec{\xi}_{\vvec{n}}} &= \vvec{d}^\top B_{\vvec{n}}, \\
        \frac{\partial h}{\partial \vvec{\xi}_{\vvec{d}}} &= \vvec{s}^\top B_{\vvec{d}},
    \end{split}
\end{equation}
where we have applied the retraction $\vvec{x} \leftarrow \vvec{x} + B_{\vvec{x}}\vvec{\xi}_{\vvec{x}}$ with $B_{\vvec{x}}$ and $\vvec{\xi}_{\vvec{x}}$ representing the basis for and local coordinates in the tangent space at a unit vector $\vvec{x} \in \mathbb{S}^2$, respectively, and
\begin{equation}
    D_\mathrm{norm}(\vvec{v}) =
    \begin{bmatrix}
        v_y^2 + v_z^2 & -v_xv_y & -v_xv_z \\
        -v_xv_y & v_x^2 + v_z^2 & -v_yv_z \\
        -v_xv_z & -v_yv_z & v_x^2 + v_y^2
    \end{bmatrix} (v_x^2 + v_y^2 + v_z^2)^{-3/2}.
\end{equation}
Next, let $b = -\cos^{-1}(h) / 60$ and $g = \exp(b)$. Then,
\begin{equation}
    \begin{split}
        \frac{\partial b}{\partial h} &= \left(60\sqrt{1 - h^2}\right)^{-1}, \\
        \frac{\partial g}{\partial b} &= \exp(b), \\
    \end{split}
    \qquad
    \begin{split}
        \frac{\partial g}{\partial \vvec{\xi}_{\vvec{s}}} &= \frac{\partial g}{\partial b}\frac{\partial b}{\partial h}\frac{\partial h}{\partial \vvec{\xi}_{\vvec{s}}}, \\
        \frac{\partial g}{\partial \vvec{\xi}_{\vvec{d}}} &= \frac{\partial g}{\partial b}\frac{\partial b}{\partial h}\frac{\partial h}{\partial \vvec{\xi}_{\vvec{d}}}.
    \end{split}
\end{equation}
This gives
\begin{equation}
    \frac{\partial I}{\partial \vvec{e}} = a \left(   -f\frac{\partial g}{\partial \vvec{\xi}_{\vvec{d}}} + 2f \left( \frac{1}{f + w} \frac{\partial g}{\partial \vvec{\xi}_{\vvec{d}}} - \frac{g}{(f + w)^2} \frac{\partial w}{\partial \vvec{\xi}_{\vvec{d}}} \right)   \right)\frac{\partial \vvec{\xi}_{\vvec{d}}}{\partial \vvec{e}}.
\end{equation}
Finally,
\begin{align}
    \frac{\partial I}{\partial \zeta} &= \frac{\partial I}{\partial\vvec{e}} \frac{\partial\vvec{e}}{\partial \zeta}, \\
    \frac{\partial I}{\partial \vvec{\xi}_{\vvec{s}}} &= a \left(   \frac{\partial f}{\partial \vvec{\xi}_{\vvec{s}}} - \left( \frac{\partial g}{\partial \vvec{\xi}_{\vvec{s}}}f + g \frac{\partial f}{\partial \vvec{\xi}_{\vvec{s}}} \right) + \frac{2}{f + w}\left(\frac{\partial g}{\partial \vvec{\xi}_{\vvec{s}}}f + g \frac{\partial f}{\partial \vvec{\xi}_{\vvec{s}}}\right) - \frac{2gf}{(f + w)^2}\frac{\partial f}{\partial \vvec{\xi}_{\vvec{s}}}  \right), \\
    \frac{\partial I}{\partial \vvec{\ell}} &= \frac{\partial I}{\partial\vvec{e}} \frac{\partial \vvec{e}}{\partial\vvec{\ell}}, \\
    \frac{\partial I}{\partial \vvec{\xi}_{\vvec{n}}} &= a \left((1 - g)\frac{\partial f}{\partial \vvec{\xi}_{\vvec{n}}} + 2g \left(\frac{1}{f + w}\frac{\partial f}{\partial \vvec{\xi}_{\vvec{n}}} - \frac{f}{(f + w)^2}\left(\frac{\partial f}{\partial \vvec{\xi}_{\vvec{n}}} + \frac{\partial w}{\partial \vvec{\xi}_{\vvec{n}}}\right)\right)   \right), \\
    \frac{\partial I}{\partial a} &= (1 - g) f + g \frac{2f}{f + w}.
\end{align}


\subsection{Sun Vector Factors}

In this section, we explicitly derive the partial derivatives for the measurement function $\vvec{s}^\fC(T, \mathbf{s})$ for the factor $f_\mathrm{SS}$ (\ref{eq:fSS}). 
Recall that $T$ denotes $T_{\fB\fC}$ and $\vvec{s}$ denotes $\vvec{s}^\fB$. 
Then
\begin{align}
    \vvec{s}^\fC + B_{\vvec{s}^\fC}\vvec{\xi}_{\vvec{s}^\fC} &= R_{\fB\fC}^\top(\vvec{s} + B_{\vvec{s}}\vvec{\xi}_{\vvec{s}}) \\
    &= R_{\fB\fC}^\top\vvec{s} + R_{\fB\fC}^\top B_{\vvec{s}}\vvec{\xi}_{\vvec{s}},
\end{align}
where we've use the retraction $\vvec{s} \leftarrow \vvec{s} + B_{\vvec{s}}\vvec{\xi}_{\vvec{s}}$.
Thus,
\begin{equation}
    \frac{\partial\vvec{\xi}_{\vvec{s}^\mathcal{C}}}{\partial\vvec{\xi}_{\vvec{s}}} = B_{\vvec{s}^\mathcal{C}}^\top R_{\fB\fC}^\top B_{\vvec{s}}
\end{equation}
Moreover, 
\begin{align}
    \vvec{s}^\fC + B_{\vvec{s}^\fC}\vvec{\xi}_{\vvec{s}^\fC} &= (R_{\fB\fC}\Exp{\vvec{\gamma}})^\top\vvec{s} \\
    &\approx (R_{\fB\fC}(I_3 + \sk{\vvec{\gamma}}))^\top\vvec{s} \\
    &= R_{\fB\fC}^\top\vvec{s} - \sk{\vvec{\gamma}}R_{\fB\fC}^\top\vvec{s} \\
    &= R_{\fB\fC}^\top\vvec{s} + \sk{R_{\fB\fC}^\top\vvec{s}}\vvec{\gamma} \\
    &= R_{\fB\fC}^\top\vvec{s} + R_{\fB\fC}^\top\sk{\vvec{s}}R_{\fB\fC}\vvec{\gamma},
\end{align}
where we've used the retraction $R_{\fB\fC} \leftarrow R_{\fB\fC}\Exp{\vvec{\gamma}}$ and the rotational invariance of the cross-product.
Thus, 
\begin{equation}
    \frac{\partial\vvec{\xi}_{\vvec{s}^\mathcal{C}}}{\partial\vvec{\gamma}} = B_{\vvec{s}^\mathcal{C}}^\top R_{\fB\fC}^\top\sk{\vvec{s}}R_{\fB\fC}.
\end{equation}

\section*{B. Analysis of Reflectance Models} \label{sec:reflect-analysis}

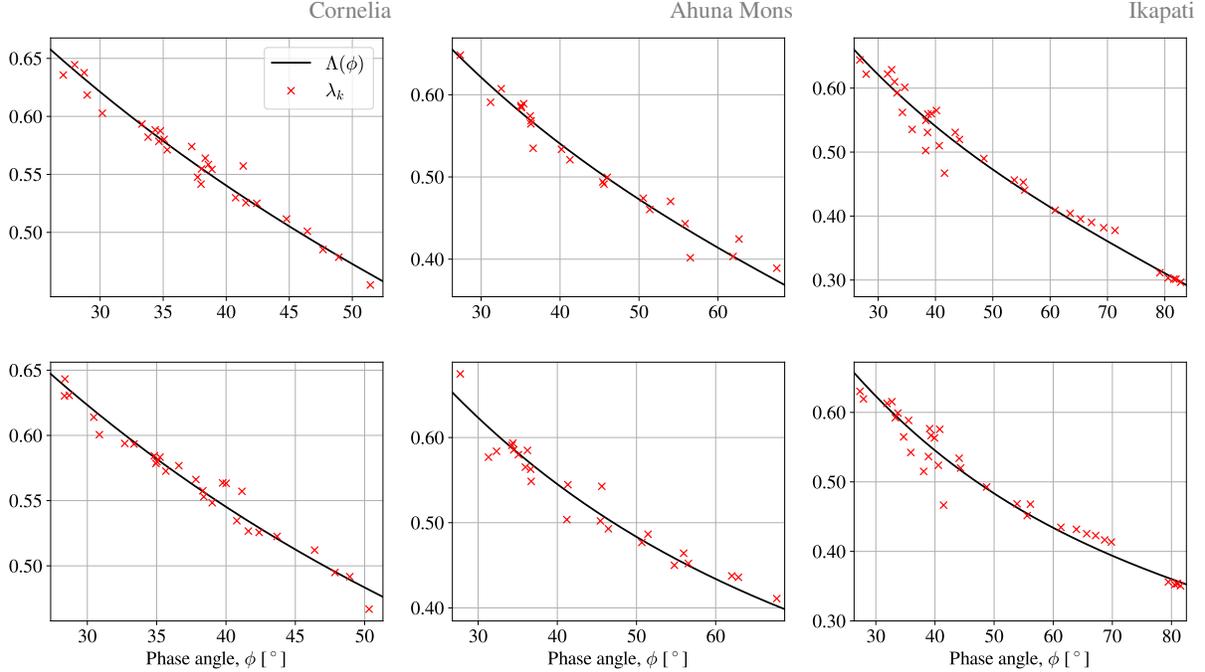
\begin{figure}[tb!]
    \centering
    \input{figures/scale_vs_aeq}
    \caption{Comparison between the (uncalibrated) scale factors $\lambda_k$ for the Akimov (top) and McEwen (bottom) models and the (calibrated) phase function $\Lambda(\phi)$ for the Akimov+ (top) and Lunar-Lambert (bottom).}
    \label{fig:scale_vs_aeq}
\end{figure}

One of the primary differences between the uncalibrated and calibrated models used in this work is the use of an explicit phase function in the calibrated case. 
It turns out that the per-image scale factors $\lambda_k$ estimated by the reflectance models used for the uncalibrated case (Equation \eqref{eq:Ik_angles_uncal}), McEwen and Akimov, after normalizing by the solar irradiance for each image (Equation \eqref{eq:brf}), approximately align with the phase correction term $\Lambda(\phi)$ in their calibrated counterparts, Lunar-Lambert and Akimov+, respectively. 
This is illustrated in Figure \ref{fig:scale_vs_aeq}, where it can be seen that the scale factors $\lambda_k$ can reliably approximate the phase-dependent brightness modeled by the phase correction term. 


\section*{C. 3D Gaussian Splatting Landmark Maps} \label{sec:landmark-err-3dgs}

\begin{figure}
    \centering
    \includegraphics[width=\linewidth]{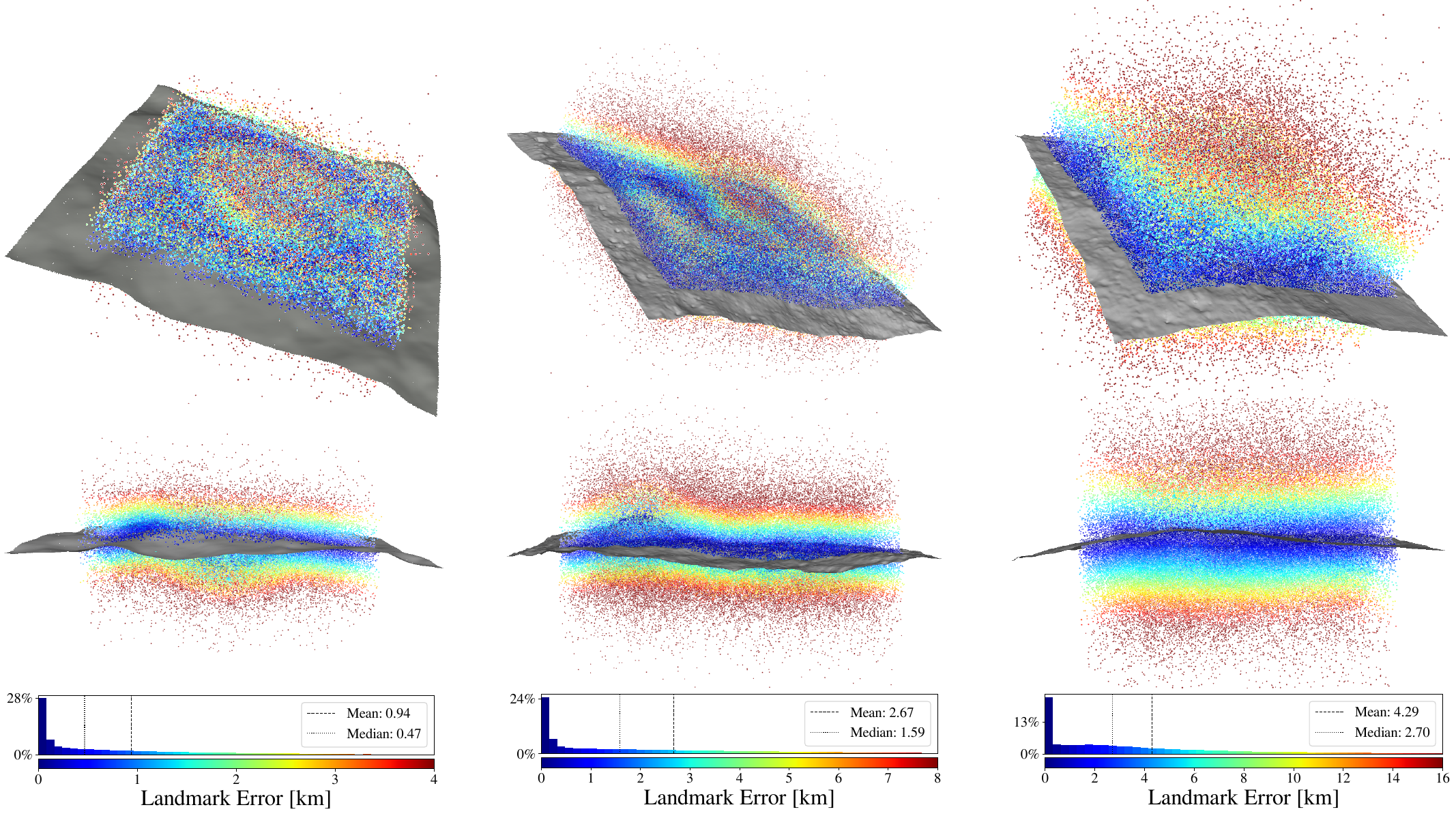}
    \caption{3DGS landmark map errors relative to the SPG baseline.}
    \label{fig:landmark-err-3dgs}
\end{figure}

The errors for the 3DGS landmarks, that is, the centers of the 3D Gaussians, relative to the SPG baselines are provided in Fig. \ref{fig:landmark-err-3dgs}. 
The 3DGS landmarks exhibit significantly more average error than the landmarks estimated by PhoMo, as shown in Fig. \ref{fig:spc-spg-sfm-comparison}. 
Specifically, the 3DGS landmarks have an average error of .94 km, 2.67 km, and 4.29 km for Cornelia, Ahuna Mons, and Ikapati, respectively, while the PhoMo landmarks (with the Lunar-Lambert model) have an average error of 15.99 m, 26.75 m, and 21.54 m for Cornelia, Ahuna Mons, and Ikapati, respectively. 
Thus, 3DGS features landmark errors that are nearly $100\times$ larger than PhoMo. 



%% file: figures/scale_vs_aeq.tex
\centering
\setlength{\extrarowheight}{-10pt}
\setlength{\tabcolsep}{2pt} 
\begin{tabular}{R{5.2cm}R{5.2cm}R{5.2cm}}
    \textcolor{gray}{\small{Cornelia}} & \textcolor{gray}{\small{Ahuna Mons}} & \textcolor{gray}{\small{Ikapati}} \\
    \includegraphics[width=\linewidth]{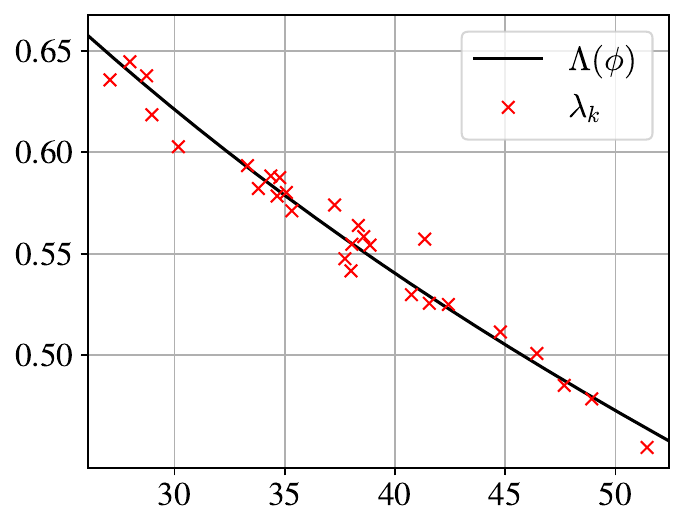} &
    \includegraphics[width=\linewidth]{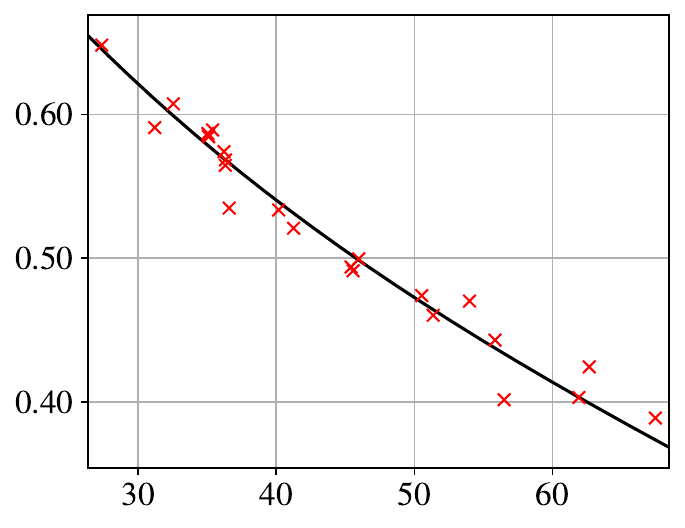} &
    \includegraphics[width=\linewidth]{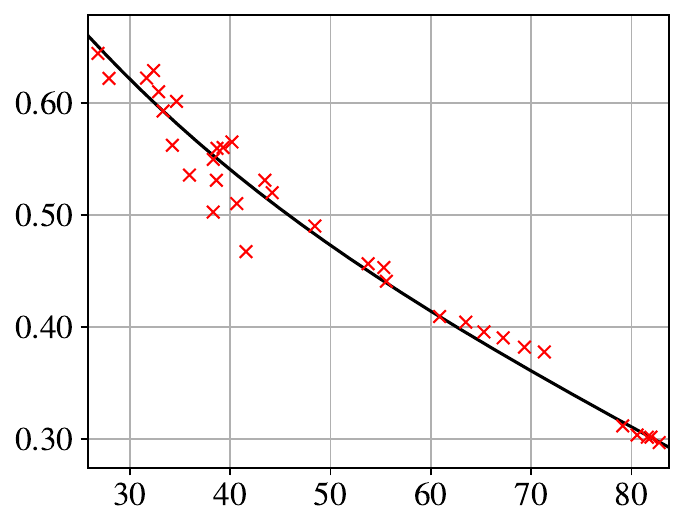} \\ \\
    \includegraphics[width=\linewidth]{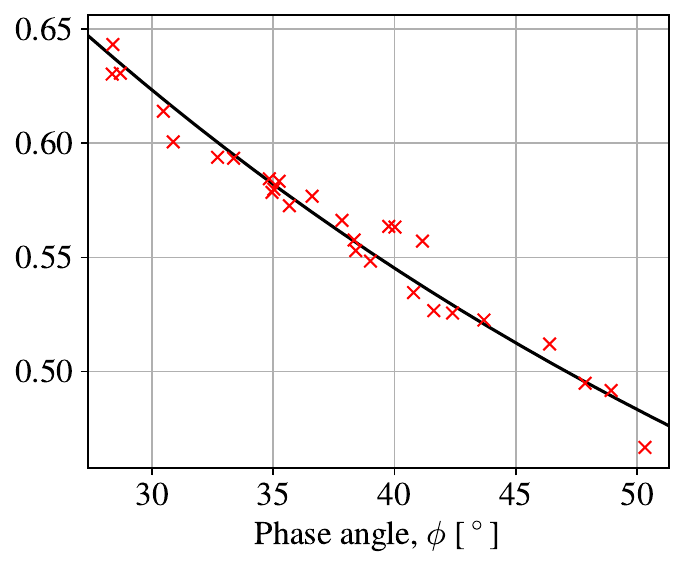} &
    \includegraphics[width=\linewidth]{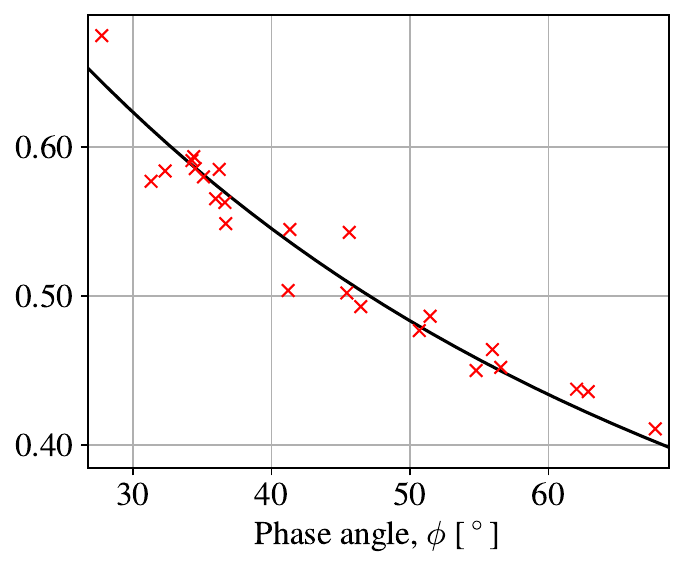} &
    \includegraphics[width=\linewidth]{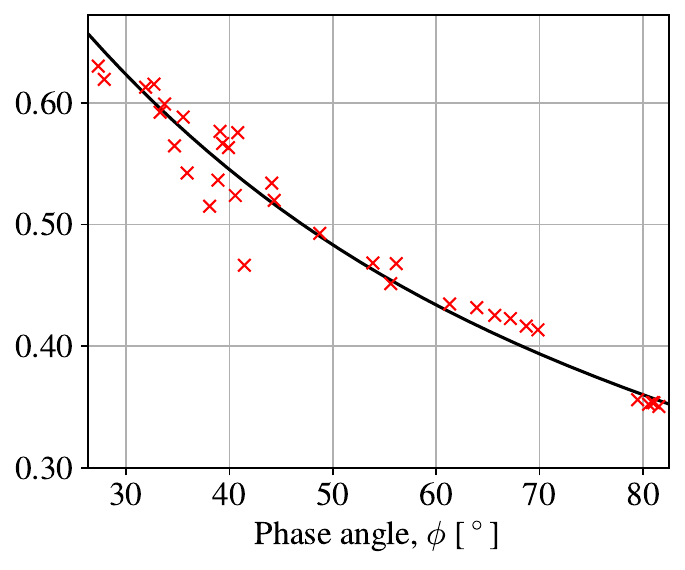} \\ 
\end{tabular}